\documentclass[conference, a4paper]{IEEEtran_ID}

\renewcommand{\contentsname}{Contents}
\renewcommand{\listfigurename}{List of Figures}
\renewcommand{\listtablename}{List of Tables}
\renewcommand{\refname}{References}
\renewcommand{\figurename}{Figure}
\renewcommand{\tablename}{Table}
\renewcommand{\appendixname}{Appendix}
\renewcommand{\abstractname}{Abstract}

\renewcommand{\IEEEkeywordsname}{Keywords}
\renewcommand{\IEEEproofname}{Proof}
\IEEEoverridecommandlockouts
\usepackage{cite}
\usepackage{fancyhdr}
\usepackage{lastpage}
\usepackage{amsmath,amssymb,amsfonts}
\usepackage{algorithmic}
\usepackage{graphicx}
\usepackage{textcomp}
\usepackage{xcolor}
\usepackage{arydshln}
\usepackage{subcaption}
\usepackage{pifont}
\def\BibTeX{{\rm B\kern-.05em{\sc i\kern-.025em b}\kern-.08em
    T\kern-.1667em\lower.7ex\hbox{E}\kern-.125emX}}

\usepackage{booktabs}

\begin{document}

\title{SimD3: A Synthetic drone Dataset with Payload and Bird Distractor Modeling for Robust Detection  
}
\author{
Ami Pandat\textsuperscript{1},
Kanyala Muvva\textsuperscript{2},
Punna Rajasekhar\textsuperscript{2},
Gopika Vinod\textsuperscript{1,2},
and Rohit Shukla\textsuperscript{1,2}\\
\textsuperscript{1}Homi Bhabha National Institute, Mumbai, India\\
\textsuperscript{2}Bhabha Atomic Research Centre, Mumbai, India
}



\maketitle
\thispagestyle{fancy}

\begin{abstract}
	Reliable drone detection is challenging due to limited annotated real-world data, large appearance variability, and the presence of visually similar distractors such as birds. To address these challenges, this paper introduces \textbf{SimD3}, a large-scale high-fidelity synthetic dataset designed for robust drone detection in complex aerial environments. Unlike existing synthetic drone datasets, SimD3 explicitly models drones with heterogeneous payloads, incorporates multiple bird species as realistic distractors, and leverages diverse Unreal Engine~5 environments with controlled weather, lighting, and flight trajectories captured using a 360$^\circ$ six-camera rig. Using SimD3, we conduct an extensive experimental evaluation within the YOLOv5 detection framework, including an attention-enhanced variant termed \textbf{Yolov5m+C3b}, where standard bottleneck-based C3 blocks are replaced with C3b modules. Models are evaluated on synthetic data, combined synthetic and real data, and multiple unseen real-world benchmarks to assess robustness and generalization. Experimental results show that SimD3 provides effective supervision for small-object drone detection and that Yolov5m+C3b consistently outperforms the baseline across in-domain and cross-dataset evaluations. These findings highlight the utility of SimD3 for training and benchmarking robust drone detection models under diverse and challenging conditions.
\end{abstract}

\begin{IEEEkeywords}
	Aerial Surveillance,
Attention Mechanisms,
Drone Detection,
Small Object Detection,
Synthetic Dataset,
Unmanned Aerial Vehicles,
YOLOv5
\end{IEEEkeywords}


\section*{Data Availability}
The dataset will be made publicly available upon acceptance of the paper.

\section*{Ethics Statement}
This study did not involve human participants, animal subjects, or sensitive data requiring ethics approval.
\section{INTRODUCTION}

Unmanned aerial vehicles (drones), commonly referred to as drones, have seen rapid adoption across a wide range of applications, including agriculture, logistics, surveillance, and recreational use. Alongside this growth, concerns related to security, privacy, and safety have increased, particularly due to unauthorized or malicious drone activity \cite{Barisic2021}. As a result, reliable drone detection has become an important research problem. Among various sensing modalities, vision-based approaches using cameras combined with deep learning (DL) techniques provide an effective and cost-efficient solution for drone detection \cite{Zhang2020}. However, the success of such models strongly depends on the availability of large and diverse annotated datasets \cite{Li2023}.

In practice, collecting large-scale real-world datasets for drone detection is challenging. Data acquisition is often time-consuming and expensive, and it may be constrained by factors such as weather conditions, restricted airspace, safety regulations, and limited access to controlled flight scenarios \cite{Chien2019}. These challenges have motivated the increasing use of synthetic data as it allows scalable and cost-effective dataset generation \cite{Dey2020}. Recent advances in game engine-based simulation have further improved the realism of synthetic data, enabling the creation of complex environments suitable for vision-based learning tasks \cite{Cattani2022}. Such platforms allow precise control over scene parameters, including lighting, weather, and time of day, which are difficult to systematically capture in real-world settings.

Despite these advantages, models trained solely on synthetic data often experience performance degradation when applied to real-world scenarios. This issue arises from the inherent gap between simulated environments and actual operational conditions, particularly in cluttered or visually ambiguous scenes. A common strategy to alleviate this problem is the use of hybrid datasets that combine synthetic and real data, where even a limited amount of real-world data can improve robustness \cite{Li2023}. However, dependence on real data reduces scalability and limits applicability in scenarios where data collection is costly, restricted, or unsafe.

More importantly, existing synthetic datasets for drone detection remain limited in their ability to capture key sources of visual ambiguity encountered in practice. In particular, many datasets do not explicitly model challenging distractors such as birds, nor do they adequately represent dynamic environmental factors such as varying weather and lighting conditions. These omissions limit the effectiveness of synthetic data for training robust drone detection models in realistic deployment scenarios.

To address these limitations, we introduce SimD3, a fully synthetic dataset designed specifically for vision-based drone detection. SimD3 incorporates diverse environments, drone models, camera viewpoints, and lighting conditions, while explicitly including multiple bird models as distractor objects. In addition, the dataset models drones with attached payloads, such as bags, boxes, and other carried objects, which are critical for distinguishing benign drones from potentially malicious ones in security sensitive scenarios. In contrast to existing synthetic datasets such as SynDroneVision \cite{Lenhard2025SynDroneVision} and Sim2Air \cite{Barisic2021}, SimD3 also simulates challenging weather conditions, including fog, rain, and snow, as well as complex drone–bird interactions commonly encountered in real-world airspace.

This paper provides a detailed description of the SimD3 dataset, including its generation pipeline and design choices. We further evaluate its effectiveness through extensive experiments using state-of-the-art drone detection models, both in isolation and in combination with real-world data. The results demonstrate that SimD3 improves detection robustness and generalization, particularly in out-of-distribution and challenging real-world scenarios.

\section{Related Work}
\label{sec:related}

This section summarizes related work on vision-based drone detection and the datasets used for training and evaluation. We review representative detection methods and then discuss key real and synthetic datasets relevant to camera-based surveillance.

\subsection{Vision-Based Drone Detection}
\label{subsec:vision_based_detection}

Early vision-based drone monitoring systems relied on hand-crafted features and traditional tracking pipelines, which were often ineffective in complex backgrounds and for small, fast-moving targets. With the adoption of deep learning, convolutional neural networks (CNNs) and modern object detectors have become the dominant approach for vision-based drone detection. Chen et al. \cite{Chen2017DroneMonitoring} proposed one of the earliest CNN-based drone monitoring systems using visible-light imagery, combining detection and tracking to handle small drones in cluttered scenes, and introduced the widely used USC Drone dataset . Subsequent work extended these ideas by integrating visible and thermal modalities to improve robustness under low illumination and adverse conditions \cite{Wang2019VisibleThermal,Elsayed2021Review}.

More recent approaches build upon general-purpose object detection architectures such as Faster R-CNN, SSD, and particularly the YOLO family. Isaac-Medina et al.\cite{IsaacMedina2021Benchmark}  provided a comprehensive benchmark of deep detectors across multiple drone datasets, highlighting both performance gains and significant sensitivity to domain shifts. To better address challenges specific to drone detection, several drone-tailored architectures have been proposed, focusing on small-object representation, scale variation, and background clutter. Representative examples include high-resolution feature enhancement and background-difference strategies, such as SAG-YOLOv5s \cite{Lv2022SAGYOLO}, as well as lightweight models optimized for real-time aerial surveillance, including EDGS-YOLOv8 \cite{Huang2024EDGSYOLO}. Recent surveys consistently identify small target size, extreme viewpoints, and changing illumination as persistent challenges, and emphasize the strong dependence of detection performance on dataset diversity and coverage \cite{Elsayed2021Review,Tang2024Survey,Dong2025Survey}.

Recent comparative analyses also suggest that newer detector versions do not necessarily provide consistent gains for all drone detection settings, particularly when the target occupies only a small number of pixels. In our earlier study evaluating multiple YOLO variants (v5-v12) for small-drone detection, YOLOv5 \cite{yolov5} was found to offer a strong accuracy efficiency trade-off and to remain highly competitive relative to newer versions under small-object and long-range conditions \cite{pandat2025}. For instance, TOE‑YOLO \cite{toeyolo}, based on YOLOv11, achieves high precision on UAV imagery with a lightweight design optimized for tiny target detection, outperforming several baseline detectors in multi-scale scenarios. Additionally, CheapConv \cite{cheapconv} integrates optimized convolutional operators into a YOLOv8 framework to enhance dense object segmentation and detection under resource constraints. These approaches share common models Yolo but differ in version and architectureal imporvement based on applications.  Motivated by these observations, we adopt YOLOv5 as the baseline detection framework in this work and focus on improving feature representation for challenging aerial scenarios through architectural refinement and dataset design.

\subsection{Real and Synthetic Drone Datasets}
\label{subsec:datasets}

A variety of real-world RGB and RGB–thermal datasets have been proposed for drone detection and tracking. The USC Drone Detection and Tracking dataset contains visible-light videos captured on a university campus with diverse viewpoints and illumination conditions and has served as a benchmark for early CNN-based methods \cite{Chen2017DroneMonitoring}. The Drone Dataset by Aksoy et al.\ provides over 4{,}000 high-resolution images of amateur drones for single-class detection \cite{Aksoy2019DroneDataset}. MAV-VID offers tens of thousands of frames captured from multiple platforms, supporting research on multi-view detection and tracking \cite{RodriguezRamos2020MAVVID}. Det-Fly focuses on drone-to-drone scenarios, providing more than 13{,}000 images recorded from a flying platform, and is primarily used for air-to-air sensing research \cite{Wu2021DetFly}. UAV-Eagle and UAVData provide smaller but challenging datasets emphasizing illumination changes, clutter, and model diversity \cite{Barisic2019UAVEagle,Zeng2021UAVData}.

Beyond RGB-only datasets, several multi-modal benchmarks have been introduced. The Halmstad multi-sensor dataset includes synchronized RGB, infrared, audio, and radar data collected at airports, supporting multi-sensor fusion research \cite{Svanstrom2021MultiSensor}. DUT-AntiUAV provides 10{,}000 annotated RGB images and additional videos for detection and tracking in complex outdoor environments \cite{Zhao2022DUTAntiUAV}, while related benchmarks extend this to RGB–thermal pairs \cite{Jiang2023AntiUAV}. VisioDECT emphasizes scenario-based multi-drone detection with over 20{,}000 RGB images captured under different weather and time-of-day conditions \cite{Ajakwe2022VisioDECT}. The Malicious Drones dataset addresses payload-aware classification but is limited in scale, containing fewer than 1{,}000 images \cite{Jamil2022MaliciousDrones}. The Drone-vs-Bird Detection Challenge dataset includes large-scale video data with both drones and birds present; however, only drones are annotated, leaving birds as unlabeled distractors rather than structured negative examples \cite{Coluccia2024DroneVsBird}.

To alleviate the cost and operational constraints associated with collecting real-world drone imagery, synthetic data has increasingly been explored for UAV detection. Barisic et al.\cite{Barisic2021} introduced S-UAV-T, a synthetic drone-to-drone dataset generated in Blender with strong texture randomization, and demonstrated that models trained on synthetic data can transfer to real videos when combined with limited real-world fine-tuning. Lenhard et al.\cite{Lenhard2025SynDroneVision} proposed SynDroneVision, a large-scale synthetic dataset targeting ground-based surveillance viewpoints using Unreal Engine environments, and showed that synthetic pretraining can significantly improve robustness when followed by fine-tuning on real data . Complementary approaches further aim to reduce the simulation-to-reality gap through domain randomization, image-to-image translation, or weather-based augmentation strategies \cite{Marez2020DomainRand,Dieter2023Synthetic,Dieter2023SimRealGap}.

Despite these advances, existing synthetic datasets remain limited in their ability to capture realistic aerial clutter and target ambiguity. In particular, S-UAV-T and related datasets focus primarily on drone-to-drone monitoring scenarios and do not model non-drone aerial objects such as birds \cite{Barisic2022Sim2Air}. While SynDroneVision extends synthetic data generation to surveillance-style viewpoints and provides large-scale annotated imagery, it similarly focuses exclusively on drone targets without explicit modeling of common aerial distractors \cite{Lenhard2025SynDroneVision}. Although some multi-modal benchmarks include birds and other aerial objects, they are often not designed as image-only detection datasets or lack explicit negative-class annotations, limiting their suitability for studying drone–bird discrimination in vision-based settings \cite{Svanstrom2021MultiSensor,Jiang2023AntiUAV}.

Overall, existing real-world datasets provide valuable benchmarks but are often limited in scale, scene diversity, or annotation scope, particularly for small-object and long-range drone detection. At the same time, current synthetic datasets primarily focus on drone-only targets or air-to-air scenarios and rarely model realistic sources of aerial ambiguity such as birds, payload-equipped UAVs, or adverse weather conditions in a unified framework. These limitations highlight the need for a more comprehensive synthetic dataset. In the following section, we describe the data generation pipeline adopted in this work to address these gaps.

\begin{table*}[t]
\centering
\renewcommand{\arraystretch}{1.15}
\setlength{\tabcolsep}{5pt}
\caption{Comparison of representative real and synthetic datasets for vision-based drone detection using RGB imagery.}
\label{tab:datasets_comparison}
\begin{tabular}{l r c c c c c}
\toprule
\textbf{Dataset} &
\textbf{Total Images} &
\textbf{Video} &
\textbf{Multi-Drone} &
\textbf{Birds} &
\textbf{Payload} &
\textbf{Max. Resolution} \\
\midrule
Drone Dataset \cite{Aksoy2019DroneDataset}               & 4{,}012          & No   & No   & No        & No  & 3840$\times$2160 \\
UAV-Eagle \cite{Barisic2019UAVEagle}                     & 510$^\star$      & Yes  & No   & No        & No  & 1920$\times$1080 \\
DUT Anti-UAV \cite{Zhao2022DUTAntiUAV}                   & 10{,}000         & (Yes)& Yes  & No        & No  & 5616$\times$3744 \\
VisioDECT \cite{Ajakwe2022VisioDECT}                     & 20{,}924$^\star$ & Yes  & Yes  & No        & No  & 852$\times$480 \\
UAVData \cite{Zeng2021UAVData}                           & 13{,}803$^\star$ & Yes  & Yes  & No        & No  & 1280$\times$720 \\
USC Drone Det.\ \& Track.\ \cite{Chen2017DroneMonitoring}& 27{,}000$^\star$ & Yes  & No   & No        & No  & 1920$\times$1080 \\
MAV-VID \cite{RodriguezRamos2020MAVVID}                  & 40{,}232         & Yes  & Yes  & No        & No  & -- \\
Det-Fly \cite{Wu2021DetFly}                              & 13{,}271$^\star$ & (Yes)& No   & No        & No  & 3840$\times$2160 \\
Malicious Drones \cite{Jamil2022MaliciousDrones}         & 776              & No   & Yes  & No        & Yes & 224$\times$224 \\
Drone-vs-Bird Ch.\ \cite{Coluccia2024DroneVsBird}        & 85{,}904         & Yes  & Yes  & Unlabeled & No  & 3840$\times$2160 \\
\midrule
S-UAV-T \cite{Barisic2022Sim2Air} (\emph{synthetic})     & 52{,}500$^\star$ & No   & Yes  & No        & No  & 608$\times$608 \\
SynDroneVision \cite{Lenhard2025SynDroneVision} (\emph{synthetic}) & 140{,}038 & Yes & Yes & No & No & 2560$\times$1489 \\
\textbf{SimD3 (ours, synthetic)}                         & \textbf{178{,}639} & No & Yes & Labeled$^\dagger$ & Yes & 1920$\times$1080 \\
\bottomrule
\end{tabular}

\vspace{1mm}
\footnotesize{
$^\star$ no predefined train/validation/test split \quad
\textbf{Yes}: explicitly supported \quad
\textbf{(Yes)}: limited or partial support \quad
$^\dagger$ VFX subset contains unannotated bird-like clutter, while Non-VFX and Weather subsets include labeled birds
}
\end{table*}

\section{Data Generation Process}
\label{sec:data_generation}

This section describes the simulation framework, 3D assets, and scenario design used to construct the SimD3 dataset. The primary objective is to generate visually diverse and realistic scenes that capture a wide range of drone appearances, flight behaviors, and environmental conditions, while explicitly incorporating challenging distractors such as birds and payload-carrying drones. SimD3 is generated using the CoSys AirSim plugin \cite{cosysairsim} integrated with Unreal Engine~5 \cite{unreal}, which enables high-fidelity rendering and control over camera sensors, object motion, and environmental parameters.

To improve scene diversity and generalization, we leverage multiple environments sourced from the Unreal Engine Marketplace \cite{marketplace}. These environments represent a variety of real-world backgrounds, including hilly terrain, mountainous regions, urban areas with dense buildings and skyscrapers, water bodies such as lakes and coastal regions, and forest-like landscapes. Drone flights are simulated across these heterogeneous scenes to ensure that the dataset captures a wide range of background textures, depth cues, and structural complexity.

\subsection{Drone and Payload Modeling}

\begin{figure*}
\centering
\captionsetup[subfigure]{labelformat=empty} 

\begin{subfigure}{0.21\textwidth}
    \centering
    \includegraphics[width=\textwidth]{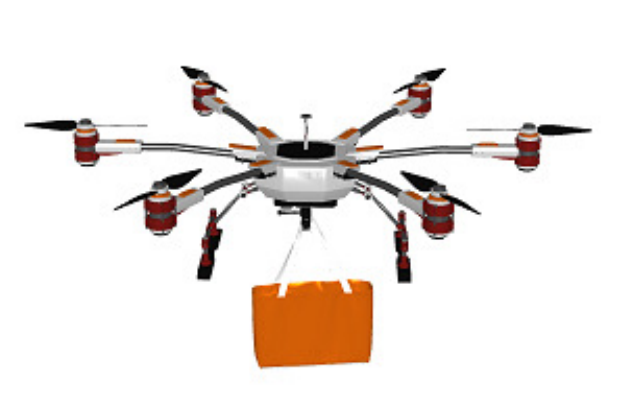}
    \caption{Model 1}
\end{subfigure}%
\begin{subfigure}{0.21\textwidth}
    \centering
    \includegraphics[width=\textwidth]{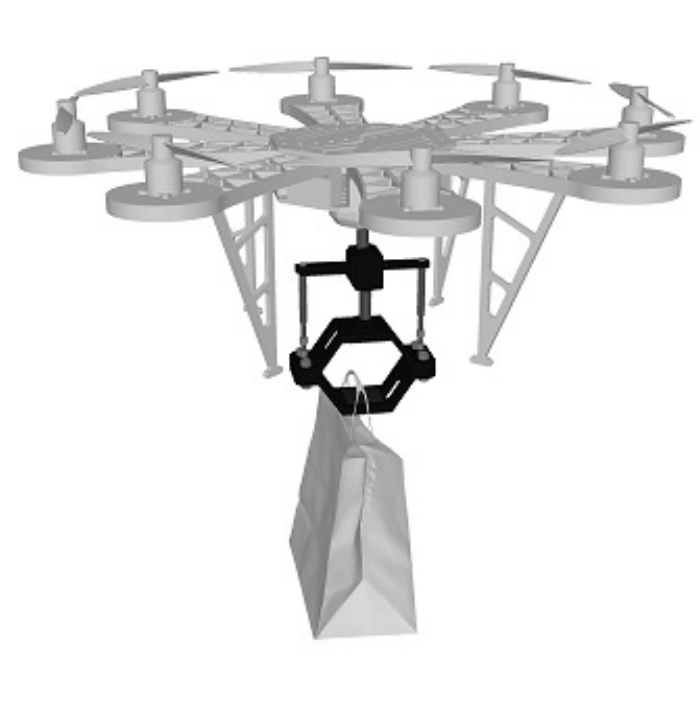}
    \caption{Model 2}
\end{subfigure}%
\begin{subfigure}{0.21\textwidth}
    \centering
    \includegraphics[width=\textwidth]{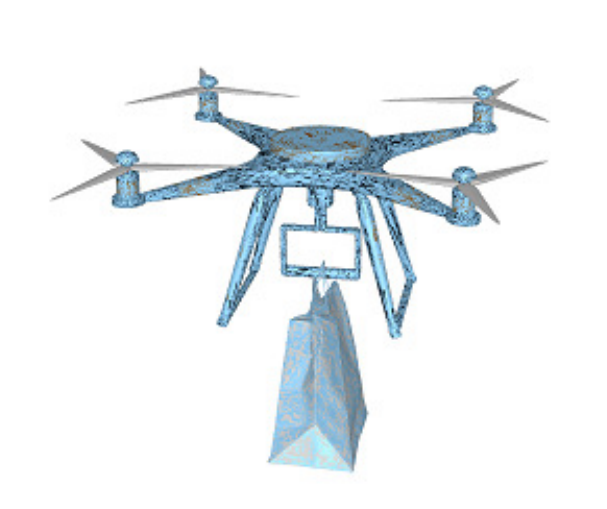}
    \caption{Model 3}
\end{subfigure}%
\begin{subfigure}{0.21\textwidth}
    \centering
    \includegraphics[width=\textwidth]{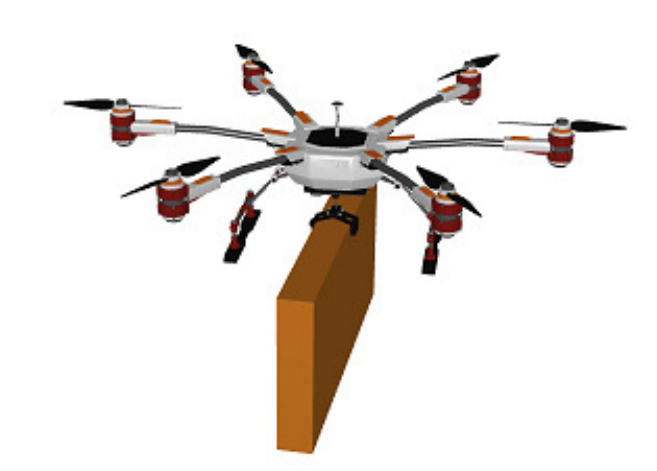}
    \caption{Model 4}
\end{subfigure}%
\begin{subfigure}{0.21\textwidth}
    \centering
    \includegraphics[width=\textwidth]{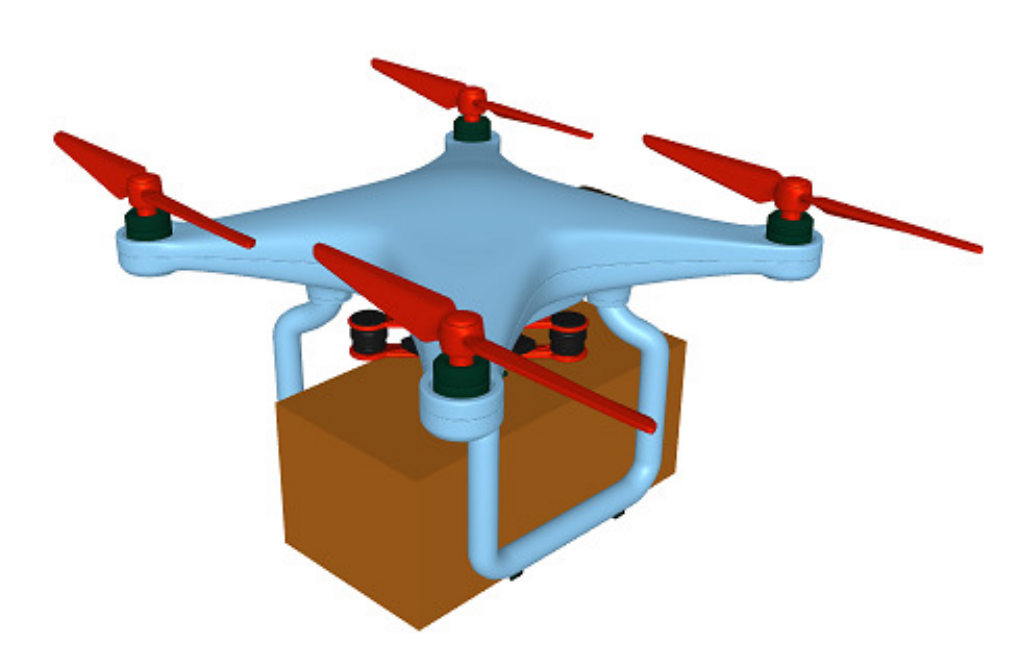}
    \caption{Model 5}
\end{subfigure}

\vspace{0.5cm}

\begin{subfigure}{0.21\textwidth}
    \centering
    \includegraphics[width=\textwidth]{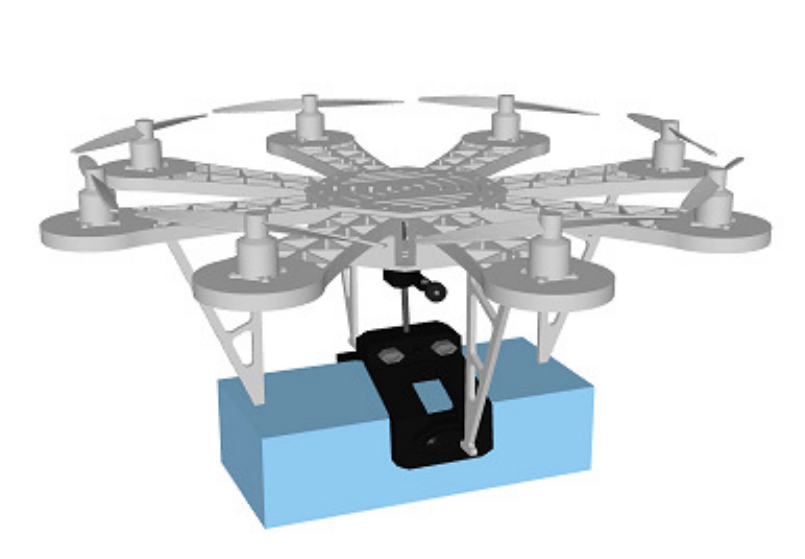}
    \caption{Model 6}
\end{subfigure}%
\begin{subfigure}{0.21\textwidth}
    \centering
    \includegraphics[width=\textwidth]{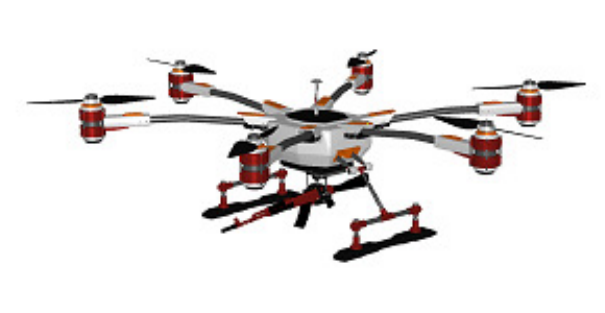}
    \caption{Model 7}
\end{subfigure}%
\begin{subfigure}{0.21\textwidth}
    \centering
    \includegraphics[width=\textwidth]{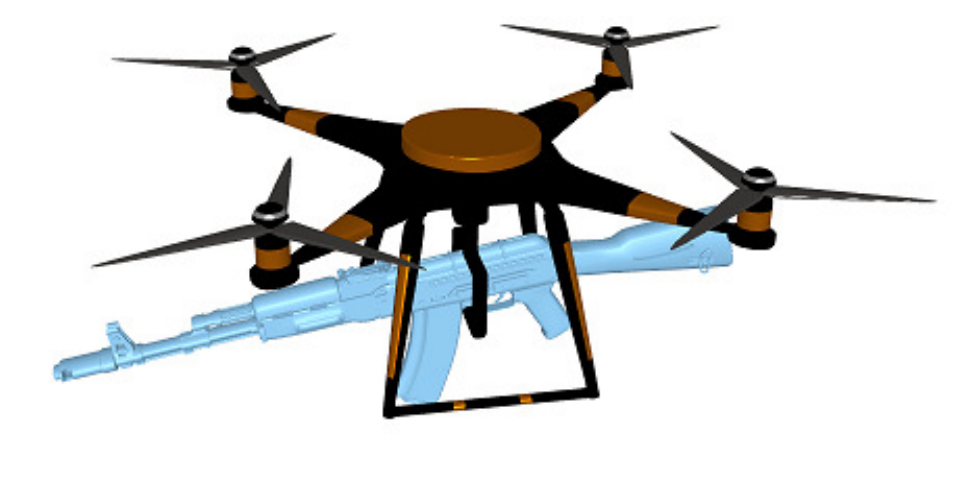}
    \caption{Model 8}
\end{subfigure}%
\begin{subfigure}{0.21\textwidth}
    \centering
    \includegraphics[width=\textwidth]{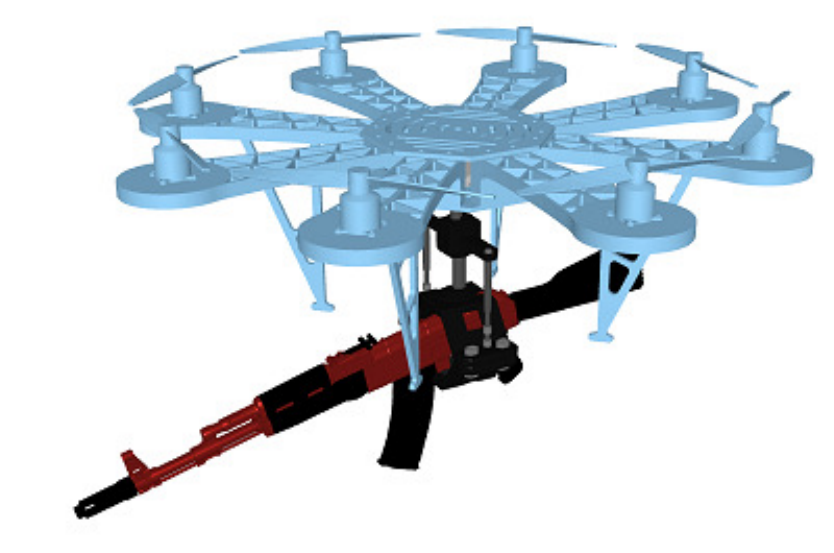}
    \caption{Model 9}
\end{subfigure}%
\begin{subfigure}{0.21\textwidth}
    \centering
    \includegraphics[width=\textwidth]{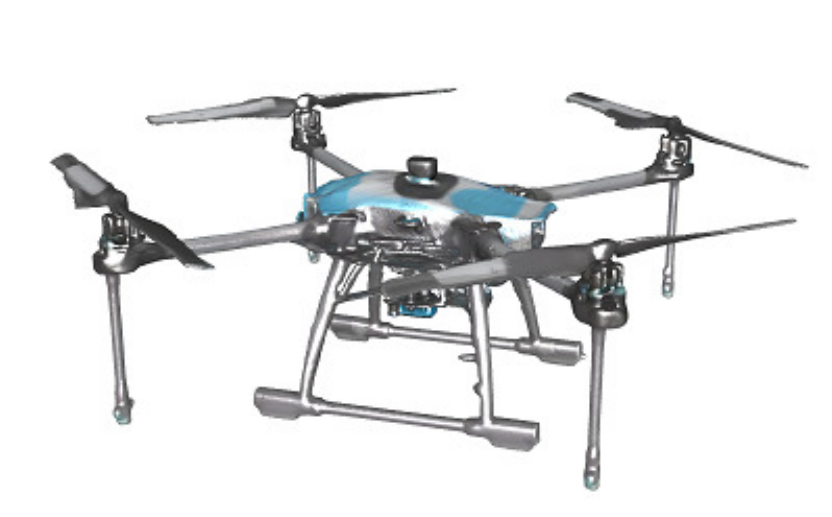}
    \caption{Model 10}
\end{subfigure}

\vspace{0.5cm}

\begin{subfigure}{0.21\textwidth}
    \centering
    \includegraphics[width=\textwidth]{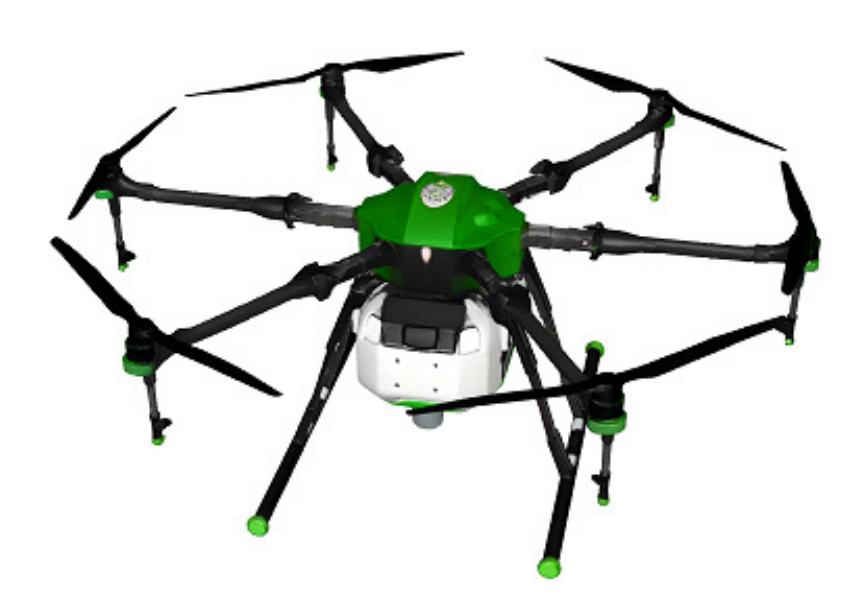}
    \caption{Model 11}
\end{subfigure}%
\begin{subfigure}{0.21\textwidth}
    \centering
    \includegraphics[width=\textwidth]{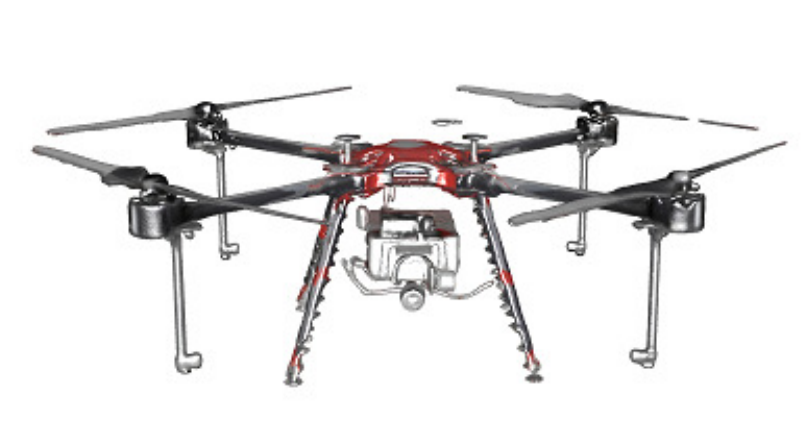}
    \caption{Model 12}
\end{subfigure}%
\begin{subfigure}{0.21\textwidth}
    \centering
    \includegraphics[width=\textwidth]{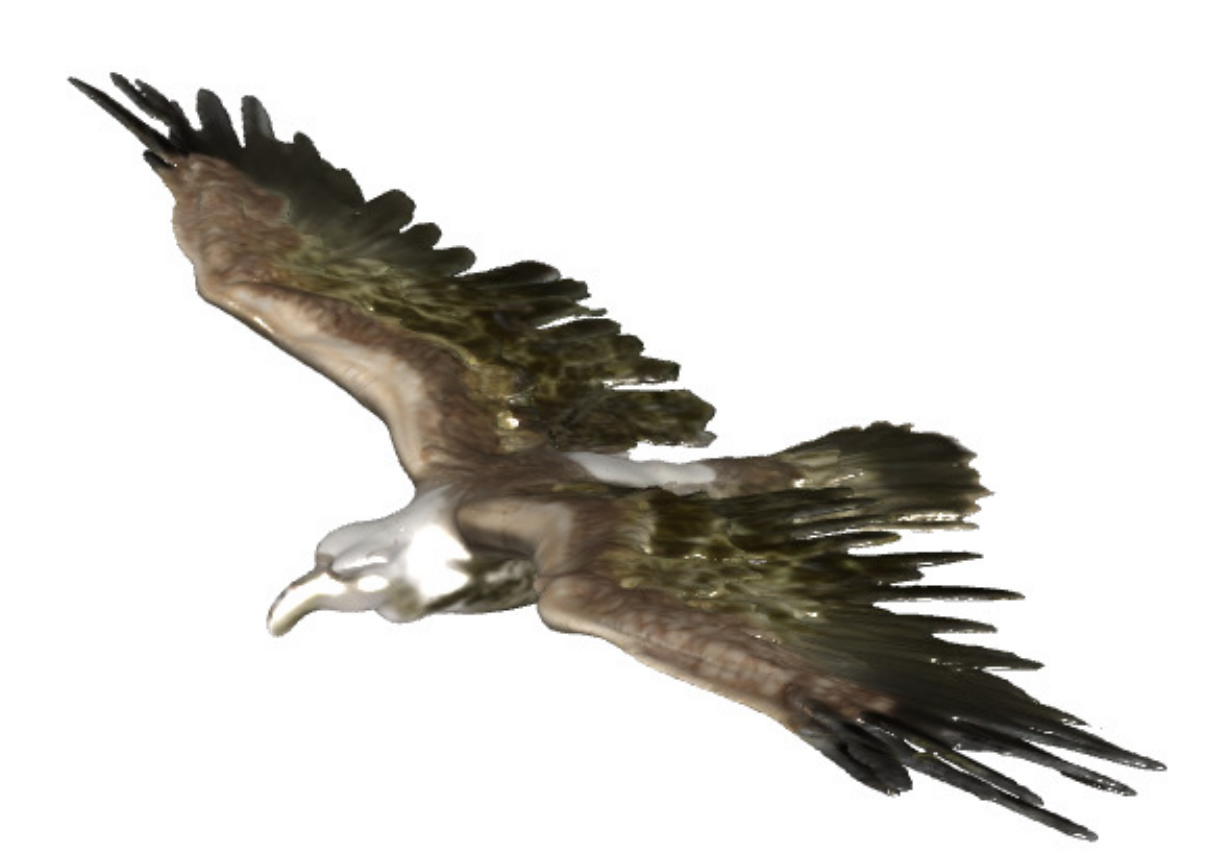}
    \caption{Model 13}
\end{subfigure}%
\begin{subfigure}{0.21\textwidth}
    \centering
    \includegraphics[width=\textwidth]{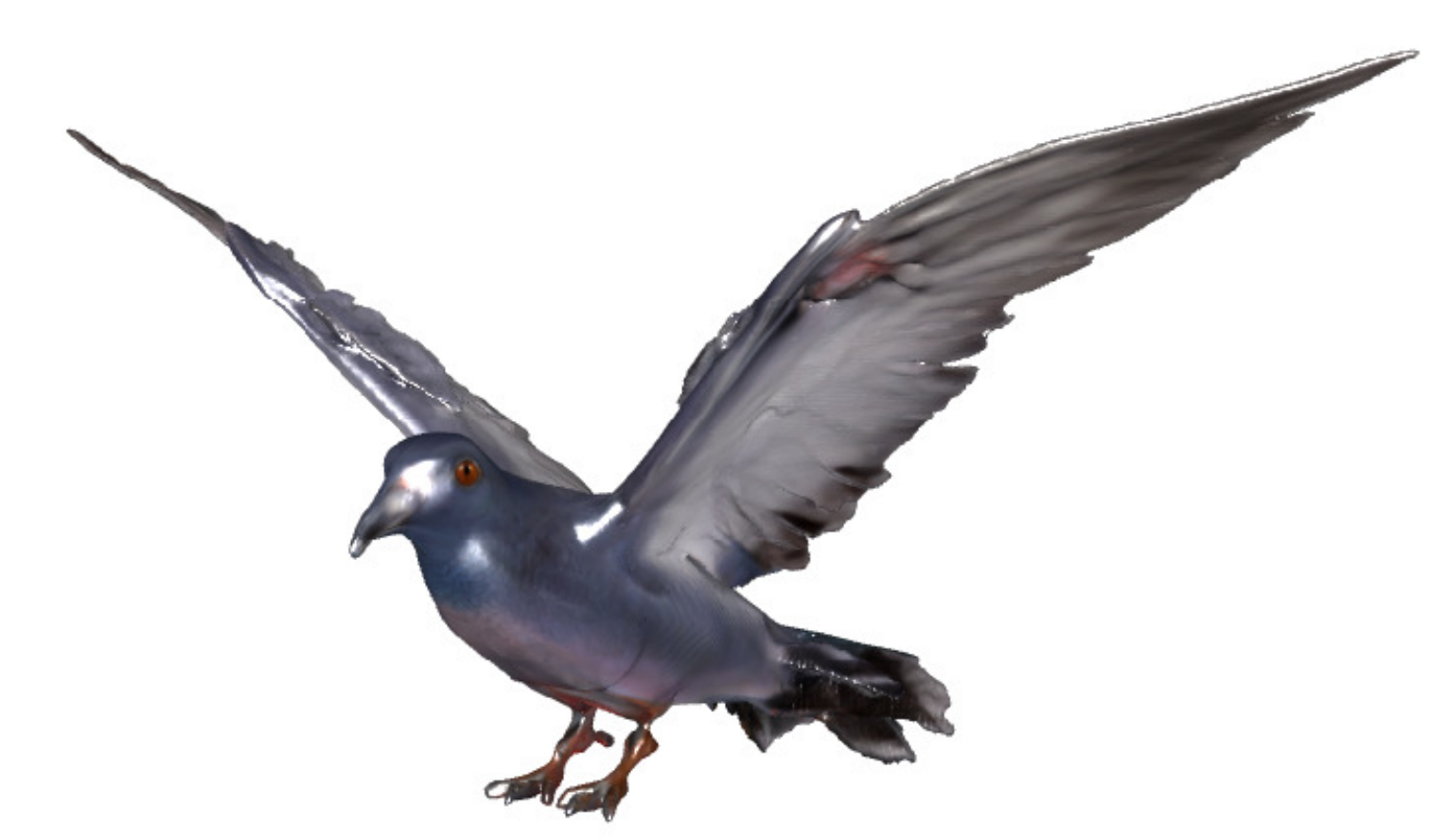}
    \caption{Model 14}
\end{subfigure}%
\begin{subfigure}{0.21\textwidth}
    \centering
    \includegraphics[width=\textwidth]{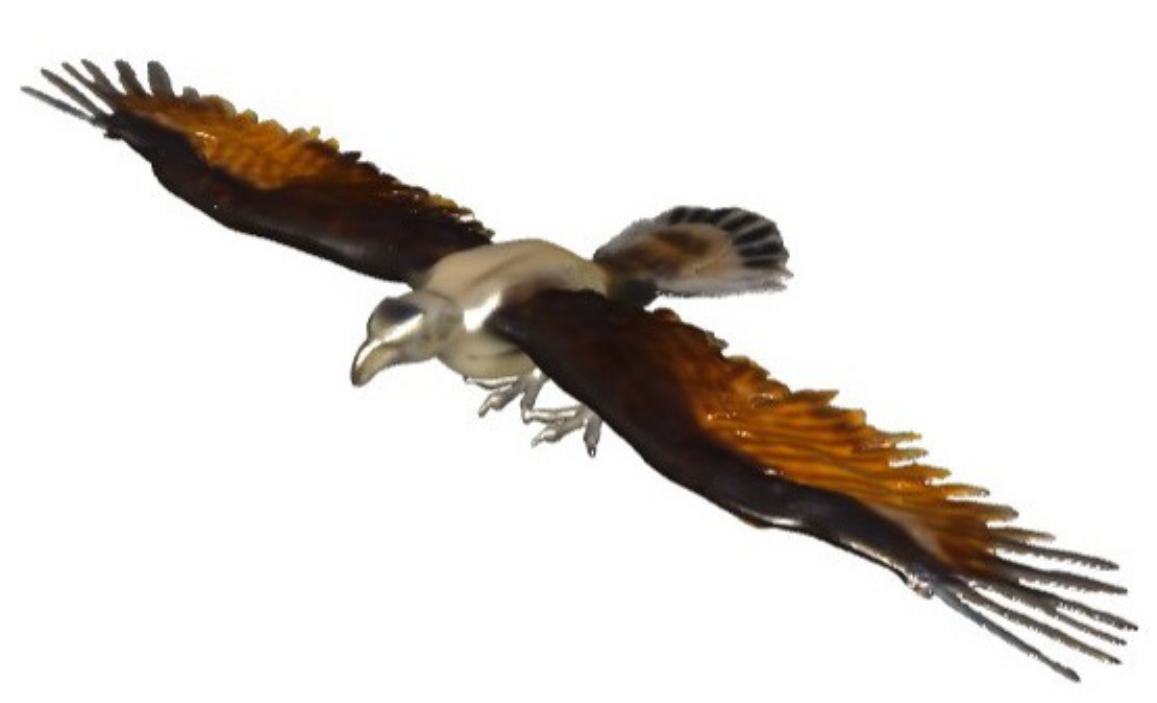}
    \caption{Model 15}
\end{subfigure}

\caption{3D models of drones and birds used for dataset generation. Models 1 to 9 feature drones equipped with different payloads, including bags, boxes, and guns. Model 11 includes an agricultural payload with a spraying kit attached, simulating real-world scenarios. These models, along with various bird species used as distractors, provide a comprehensive set of testing scenarios for drone detection.}
\label{fig:models_grid}
\end{figure*}

To construct a diverse set of aerial targets, we created a total of 15 distinct drone models using a combination of Blender \cite{blender} and Meshy AI \cite{meshy_ai}. The models span common drone categories, including quadcopters, hexacopters, and octocopters, and are designed to reflect both commercial and custom-built drones encountered in real-world scenarios. Figure \ref{fig:models_grid} highlights a few drone models used in SimD3 dataset.

Among these models, 8 drones are equipped with explicit payloads, while the remaining 7 represent  payload-free drones:
\begin{itemize}
    \item \emph{Payload equipped drones (8 models):} These drones carry visually distinctive payloads, including box shaped parcels, bag like objects, and weapon shaped payloads (e.g., gun). The presence of payloads significantly alters the silhouette and visual appearance of the drones, increasing detection difficulty and better reflecting security sensitive scenarios involving potentially malicious drones.
    \item \emph{Non payload drones (7 models):} These models represent conventional drones equipped only with onboard camera sensors and serve as baseline drone instances within the dataset.
\end{itemize}

Additional variability is introduced through randomization of materials, color schemes, and minor geometric attributes. During data generation, payload type, orientation, and attachment configuration are randomized across scenes to maximize intra-class diversity and to prevent overfitting to specific visual patterns.

\subsection{Bird Models as Realistic Distractors}

In addition to drones, SimD3 incorporates 8 distinct 3D bird models that act as realistic aerial distractors. Birds are placed at different distances, altitudes, and motion trajectories relative to drones to simulate realistic airspace activity and introduce hard negative examples.

To further enhance realism, we employ Niagara-based visual effects \cite{UE_niagara} to simulate bird flocks in selected scenes. These flocking behaviors introduce complex motion patterns and occlusions commonly observed in natural environments. These bird assets vary in size, wing geometry, and texture, and are animated using physically plausible flight behaviors such as flapping, gliding, and banking turns. Importantly, while individual bird models are explicitly annotated, bird flocks generated via Niagara effects \cite{UE_niagara} are intentionally left unannotated. As a result, the dataset contains both annotated bird instances and unannotated bird-like background motion, allowing users to treat birds either as explicit object classes or as background clutter, depending on the target application.

The inclusion of bird objects serves two key purposes:
\begin{enumerate}
    \item It reflects real-world surveillance conditions, where drones are frequently confused with birds, especially at long ranges or under low visibility.
    \item It encourages the development of more robust detection models by requiring discrimination between drones and visually similar non-drone aerial objects.
\end{enumerate}

To the best of our knowledge, no publicly available synthetic dataset for vision-based drone detection explicitly includes both labeled bird objects and unlabeled bird-like distractors within the same framework. This design choice represents a core contribution of the SimD3 dataset and significantly enhances its applicability for realistic and safety critical drone detection research.

\subsection{Virtual Environments, Weather Conditions, and Camera Configuration}

\begin{figure*}
\centering
    \begin{subfigure}[t]{0.32\textwidth}
        \includegraphics[width=\linewidth]{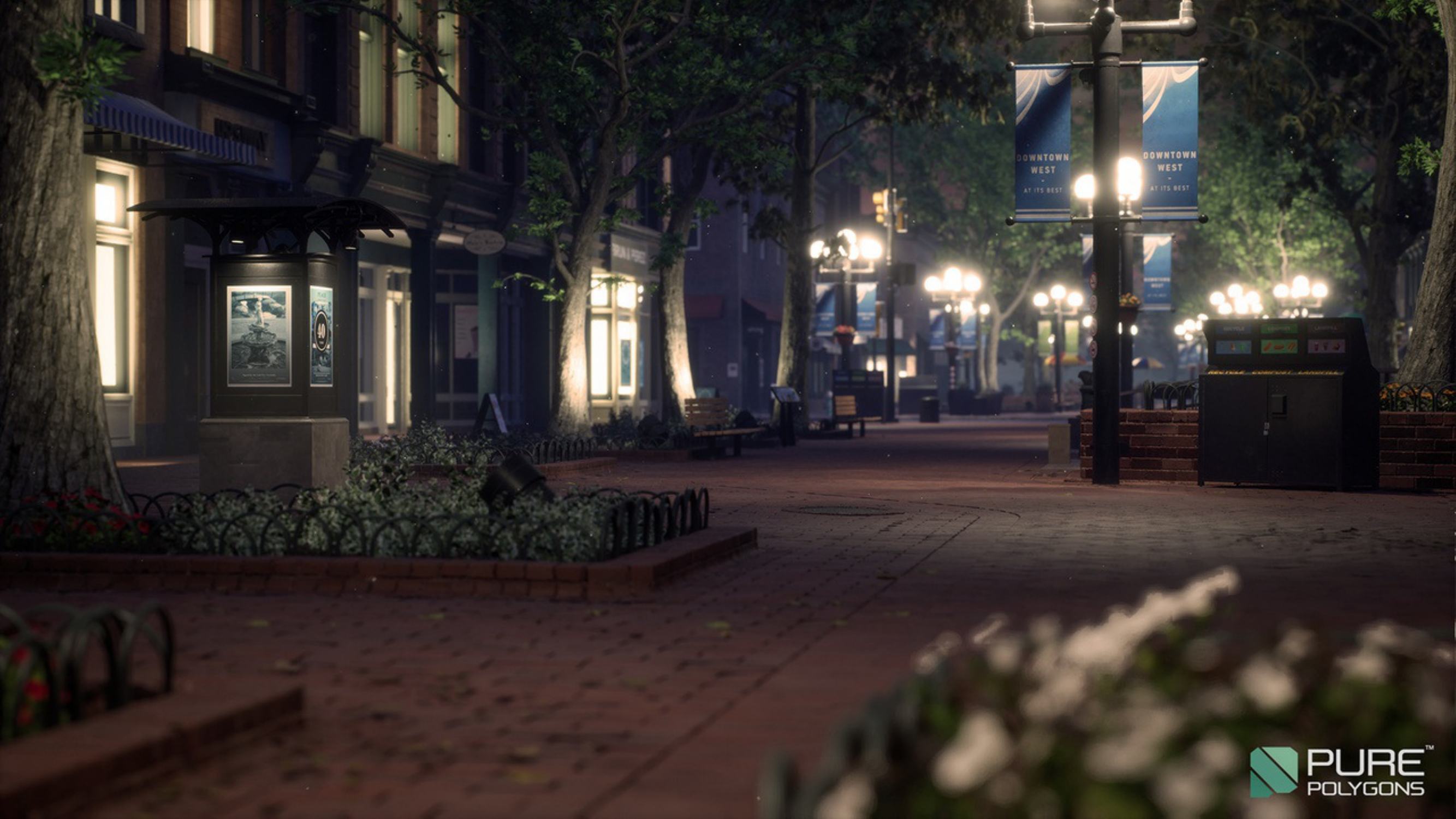}
    \end{subfigure}
    \begin{subfigure}[t]{0.32\textwidth}
        \includegraphics[width=\linewidth]{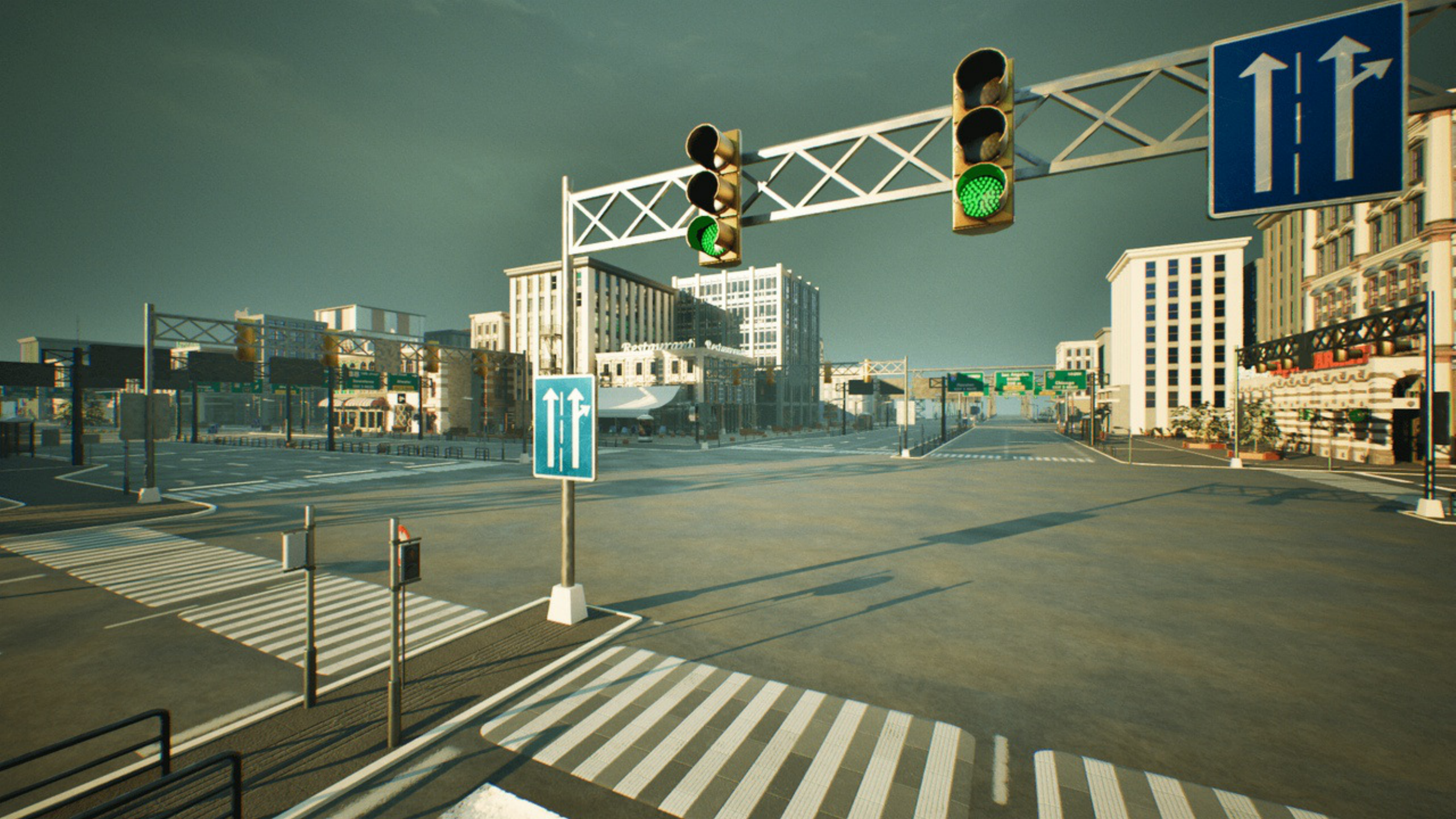}
    \end{subfigure}
    \begin{subfigure}[t]{0.32\textwidth}
          \includegraphics[width=\linewidth]{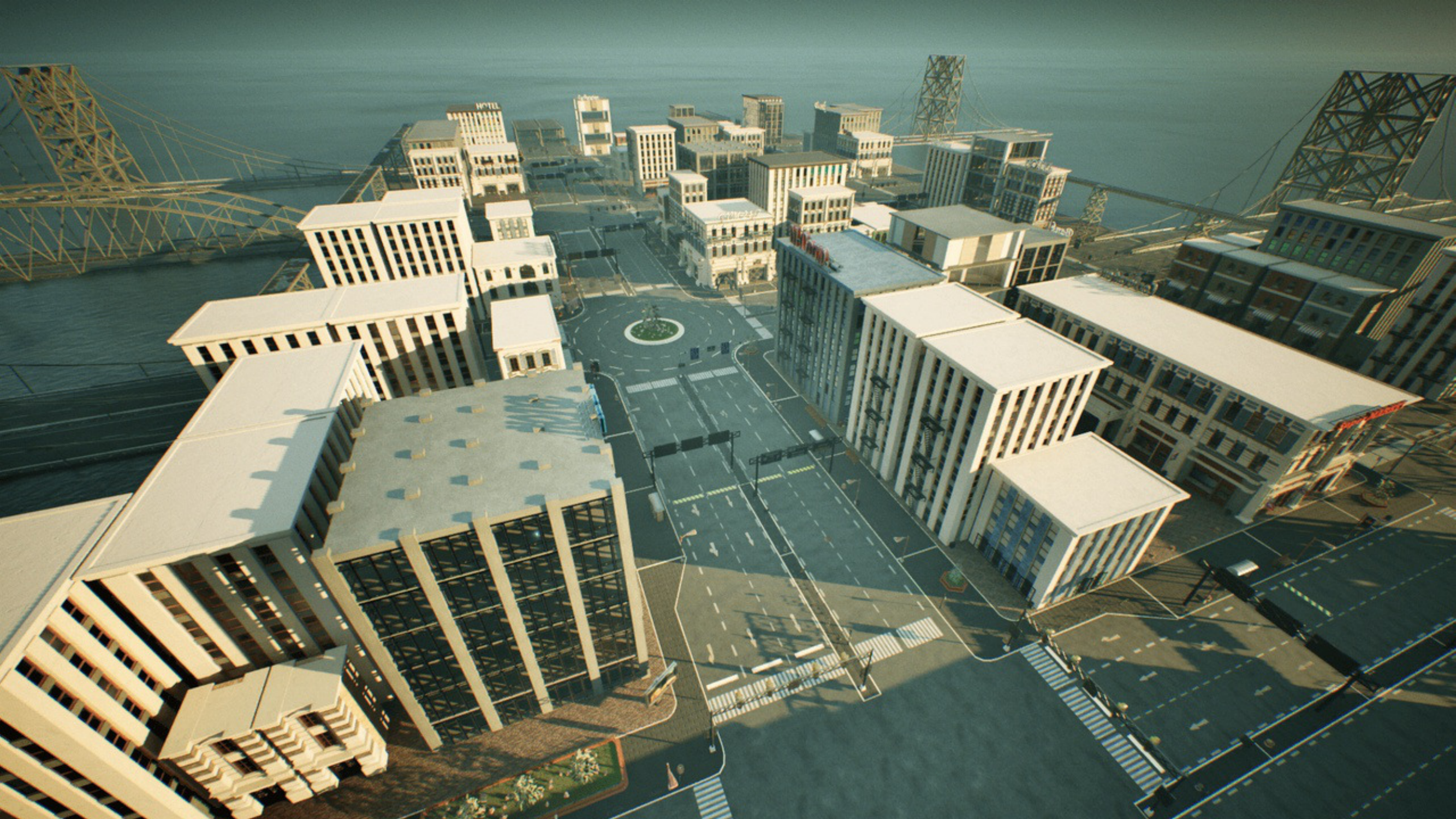}
    \end{subfigure}

    \begin{subfigure}[t]{0.32\textwidth}
        \includegraphics[width=\linewidth]{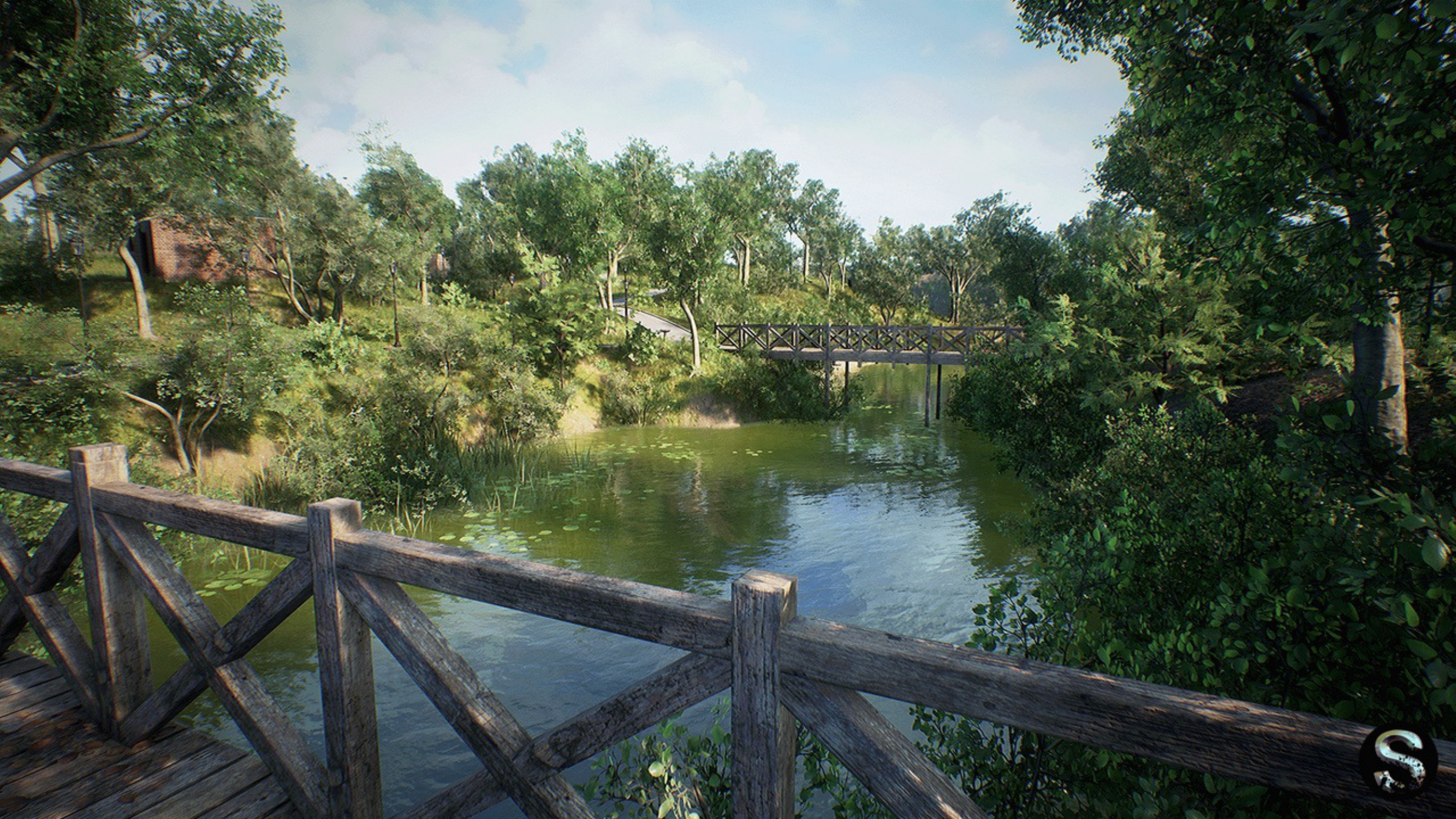}
    \end{subfigure}
    \begin{subfigure}[t]{0.32\textwidth}
        \includegraphics[width=\linewidth]{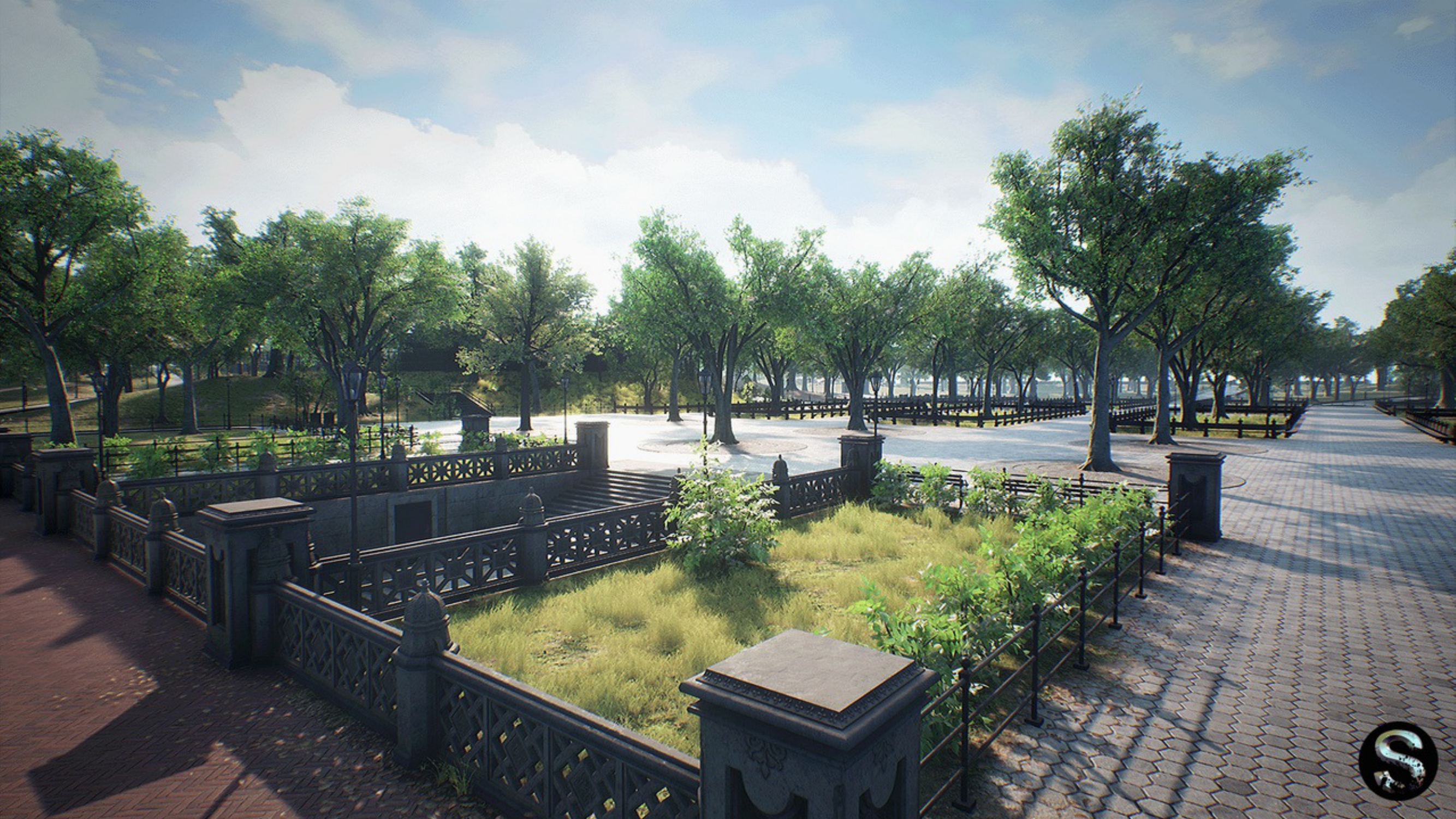}
    \end{subfigure}
    \begin{subfigure}[t]{0.32\textwidth}
         \includegraphics[width=\linewidth]{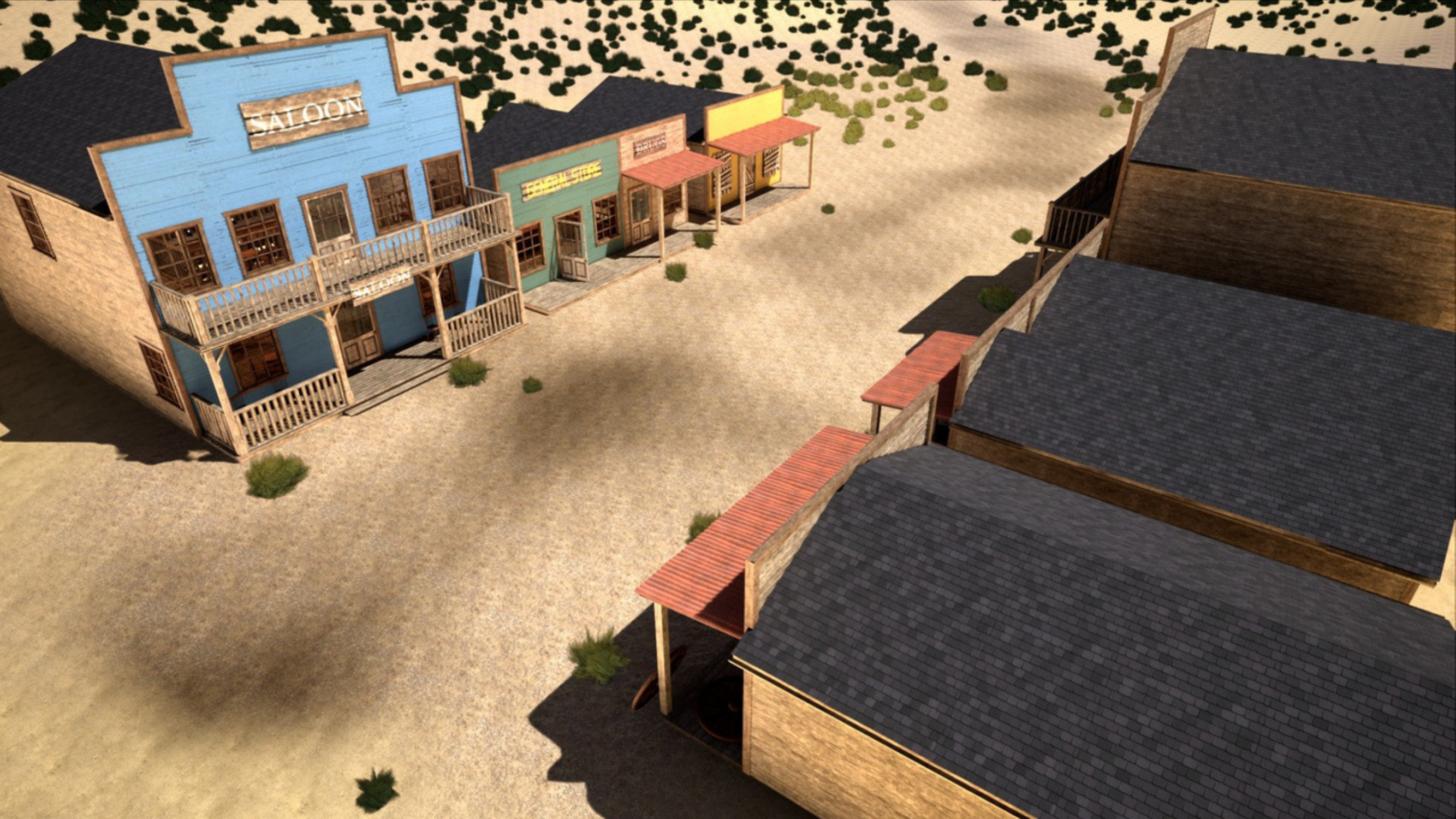}
    \end{subfigure}
 \vspace{0.2in}
    \begin{subfigure}[t]{0.32\textwidth}
        \includegraphics[width=\linewidth]{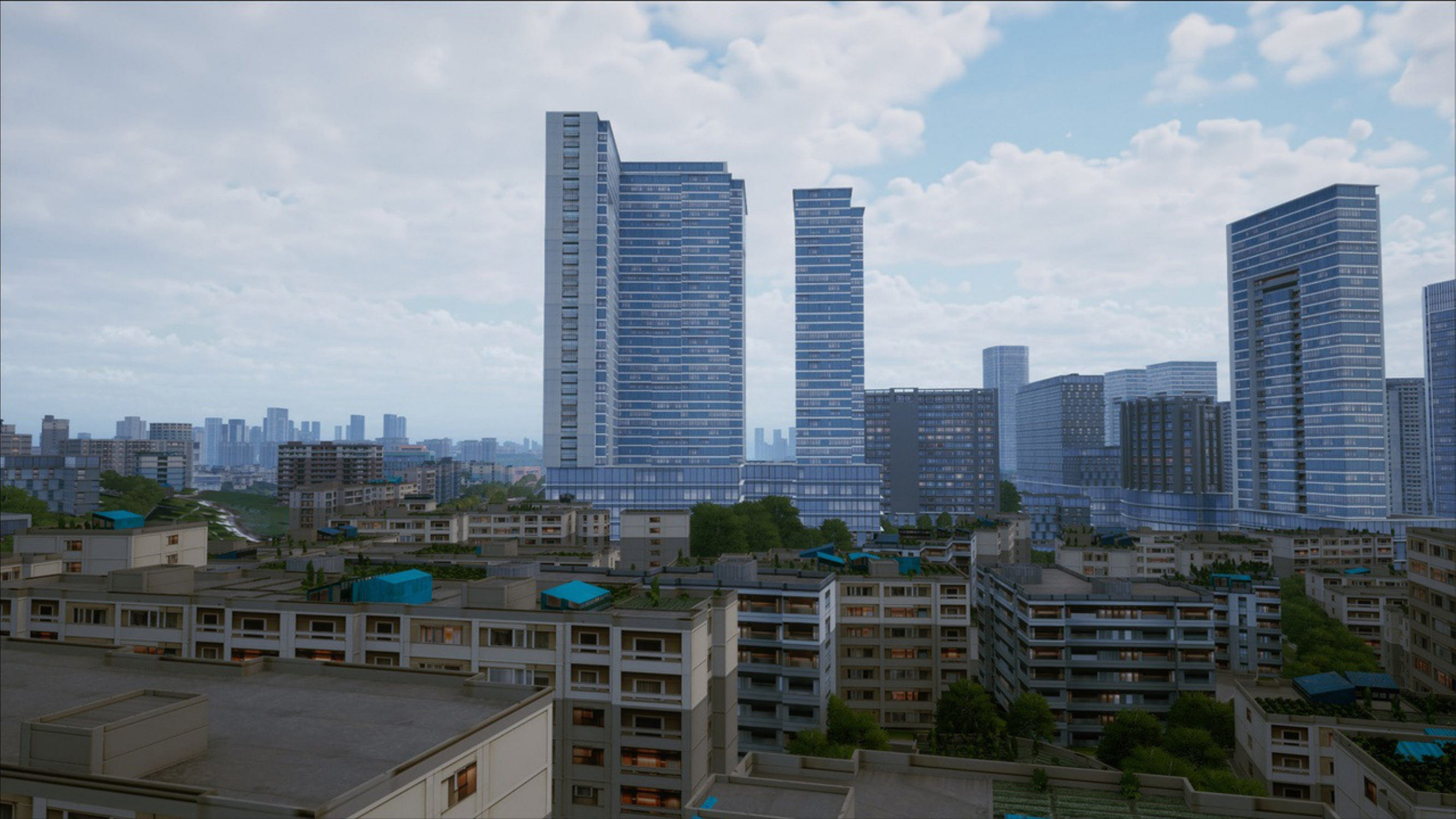}
    \end{subfigure}
    \begin{subfigure}[t]{0.32\textwidth}
        \includegraphics[width=\linewidth]{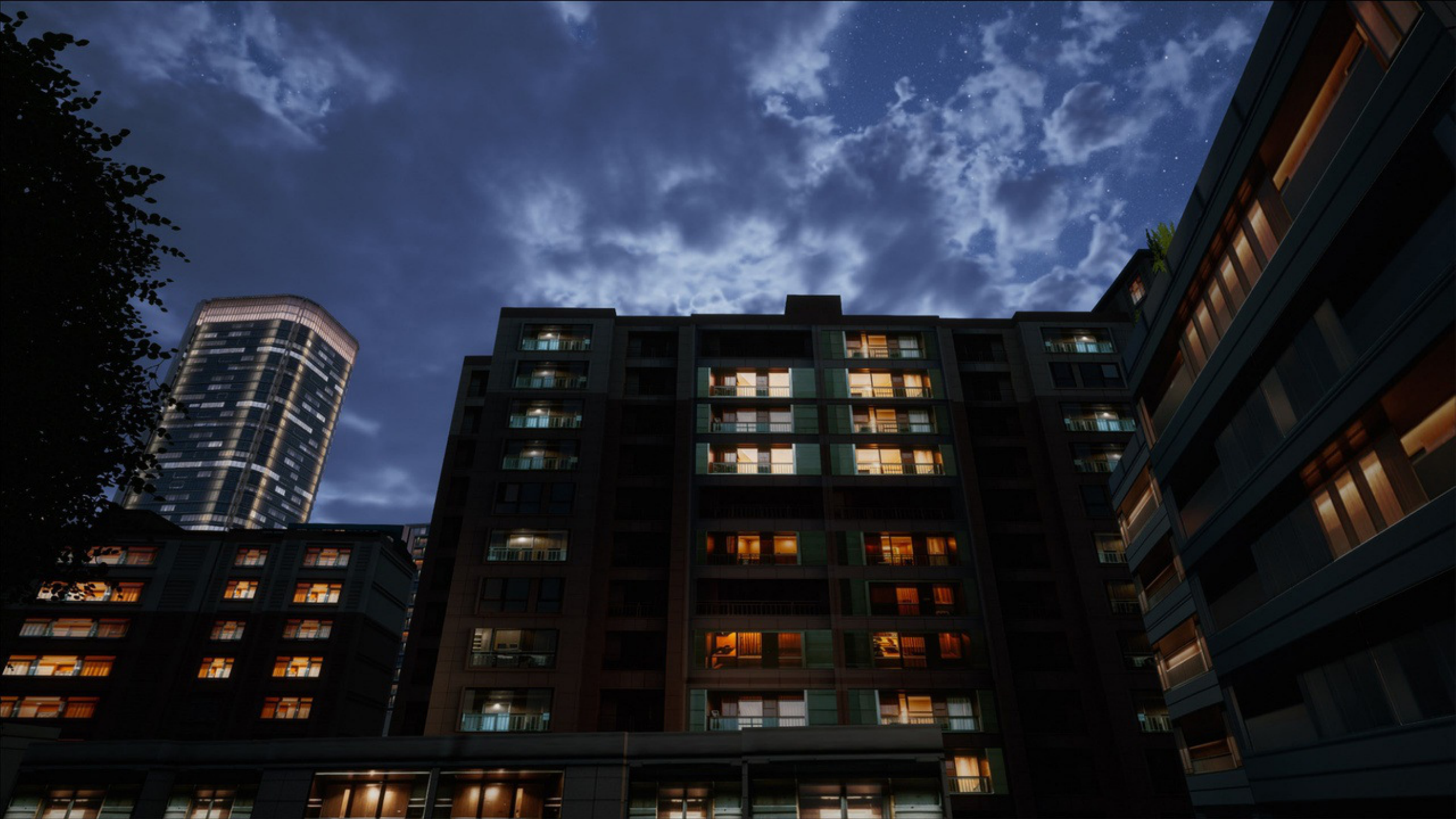}
    \end{subfigure}
    \begin{subfigure}[t]{0.32\textwidth}
        \includegraphics[width=\linewidth]{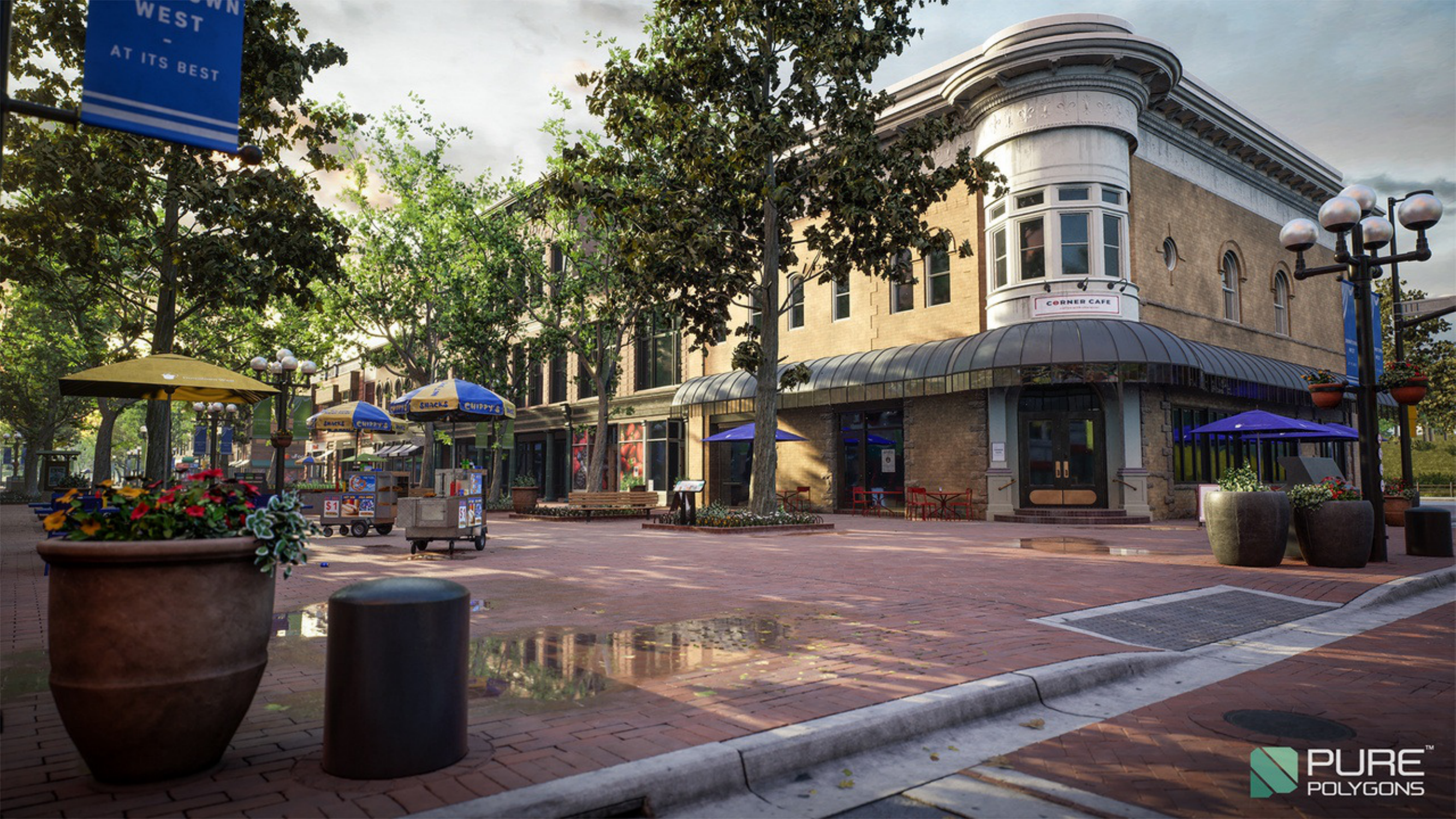}
    \end{subfigure}
    
    \caption{Virtual Unreal Environments integrated from Unreal Engine Marketplace}
    \label{fig:inf_local}
\end{figure*}

Data generation in SimD3 is conducted across multiple virtual environments created using Unreal Engine~5 and sourced from the Unreal Engine Marketplace. These environments are selected to represent a broad range of real-world surveillance contexts and include assets such as \emph{City Park} \cite{UE_CityPark}, \emph{City Creator} \cite{UE_CityCreator}, \emph{Downtown} \cite{UE_Downtown}, \emph{Dynamic City} \cite{UE_DynamicCity}, \emph{Rural Australia} \cite{UE_RuralAustralia}, \emph{Bridge} \cite{UE_Bridge}, and \emph{Wild West Town} \cite{UE_WildWestTown}. Collectively, these scenes cover diverse background structures, including dense urban areas with skyscrapers, suburban and park-like regions, rural and forested landscapes, water bodies, and complex terrain. A few exmples of these scenes is shown in Figure \ref{fig:inf_local}

\begin{figure*}[t]
\centering
\setlength{\tabcolsep}{2pt}
\renewcommand{\arraystretch}{1.0}

\begin{tabular}{lcccccc}
 & \scriptsize cam0 & \scriptsize cam1 & \scriptsize cam2 & \scriptsize cam3 & \scriptsize cam4 & \scriptsize cam5 \\ \hline

\scriptsize CityCreator \cite{UE_CityCreator} &
\includegraphics[width=0.12\textwidth]{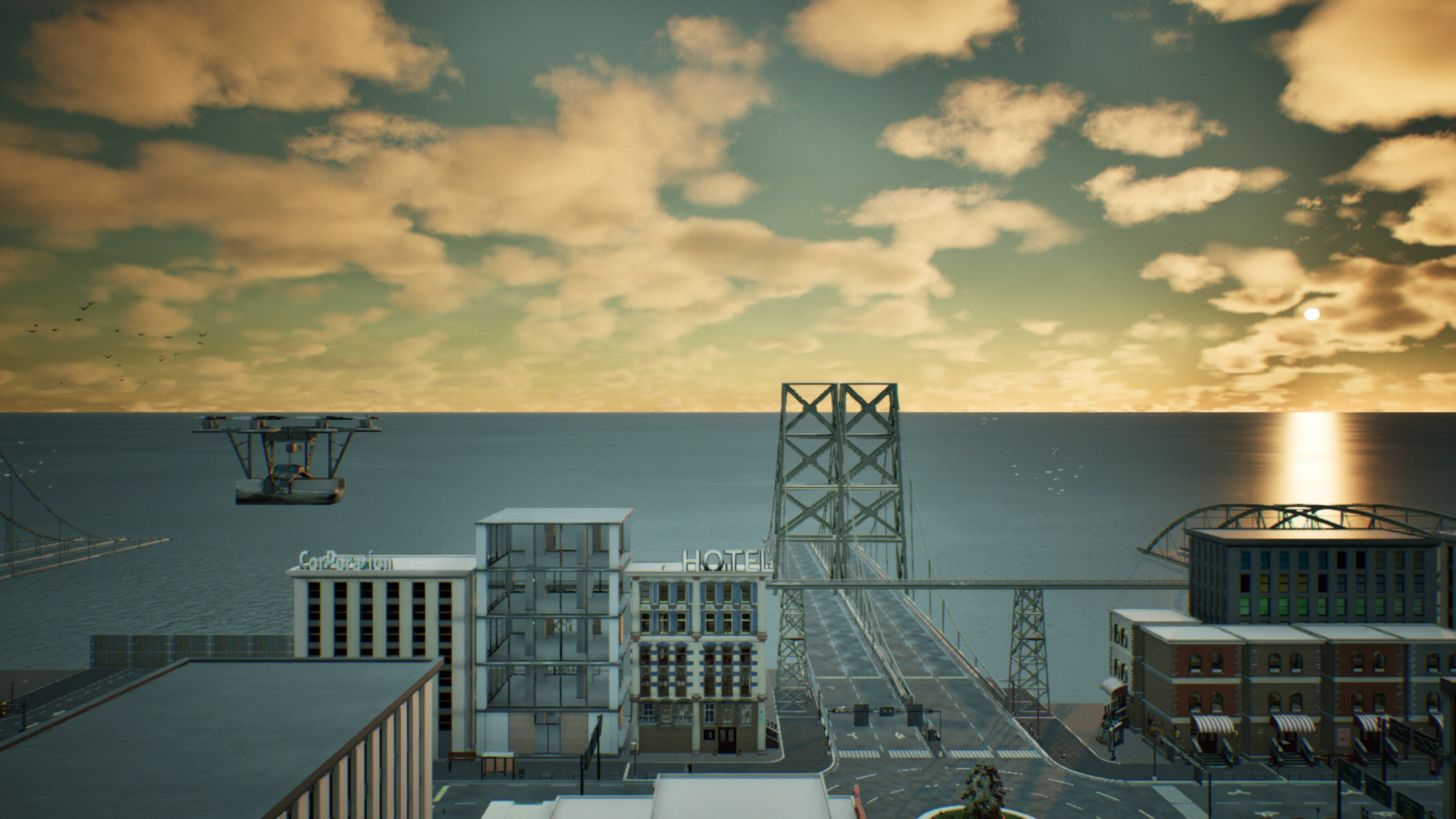} &
\includegraphics[width=0.12\textwidth]{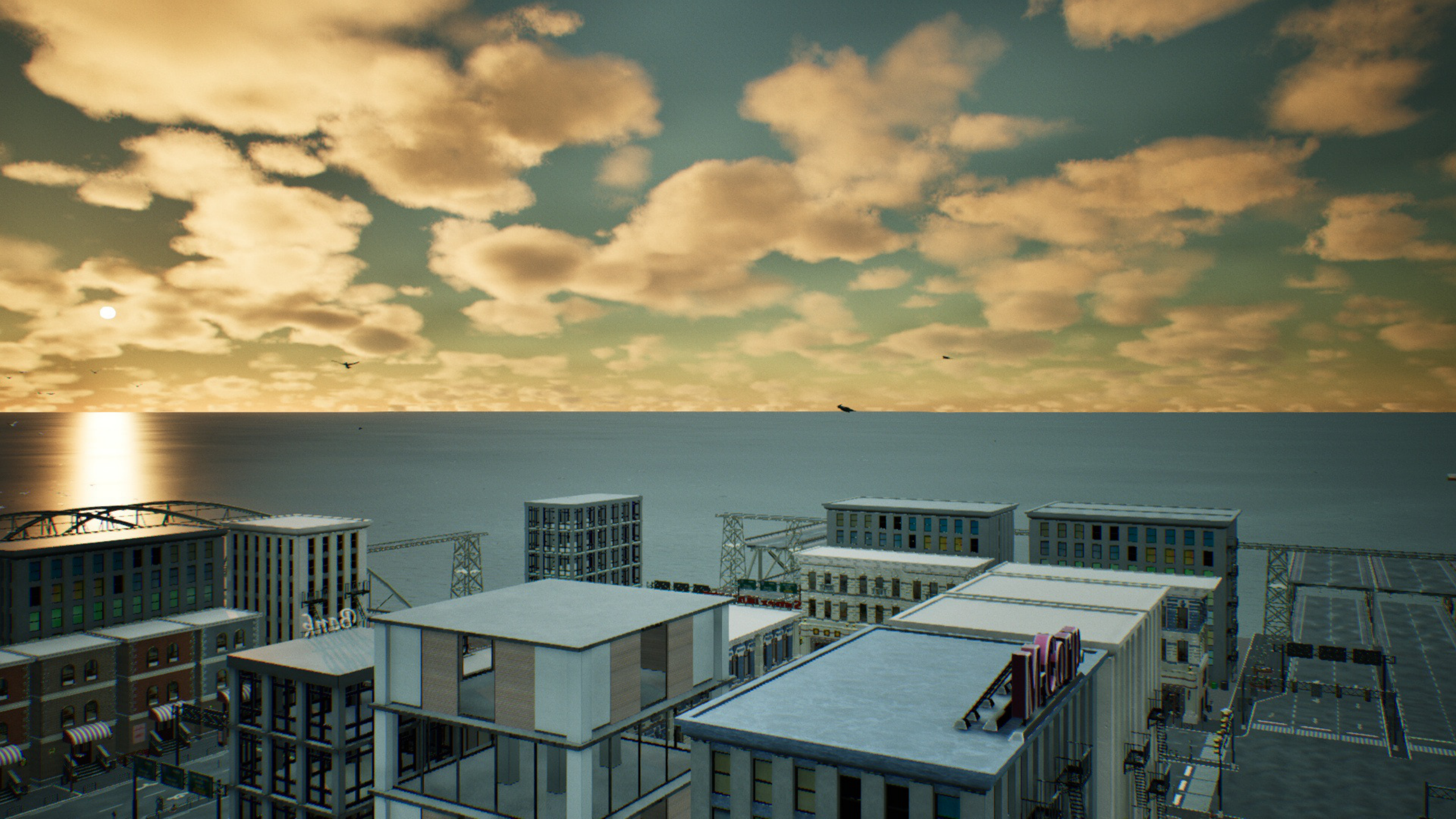} &
\includegraphics[width=0.12\textwidth]{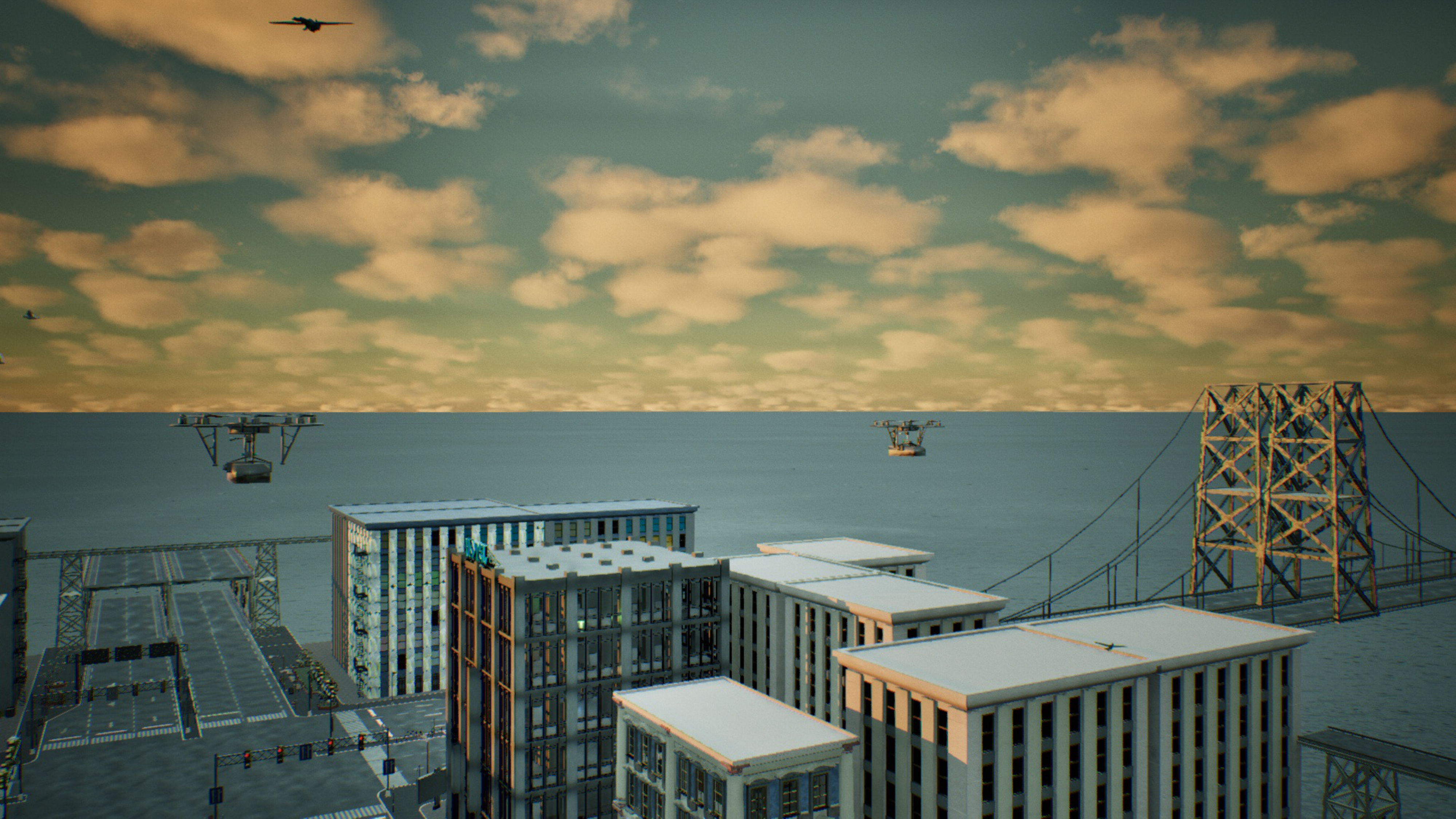} &
\includegraphics[width=0.12\textwidth]{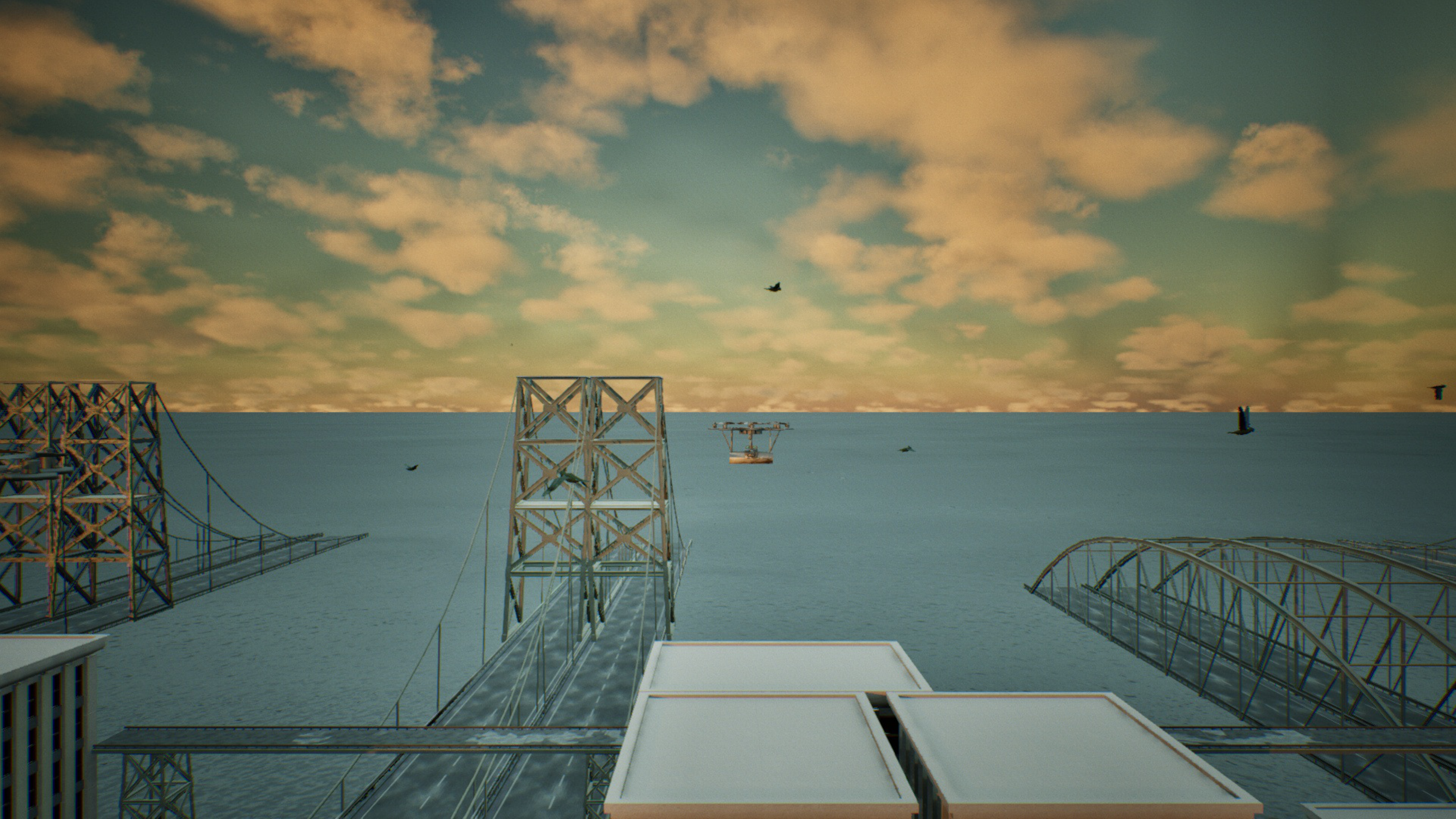} &
\includegraphics[width=0.12\textwidth]{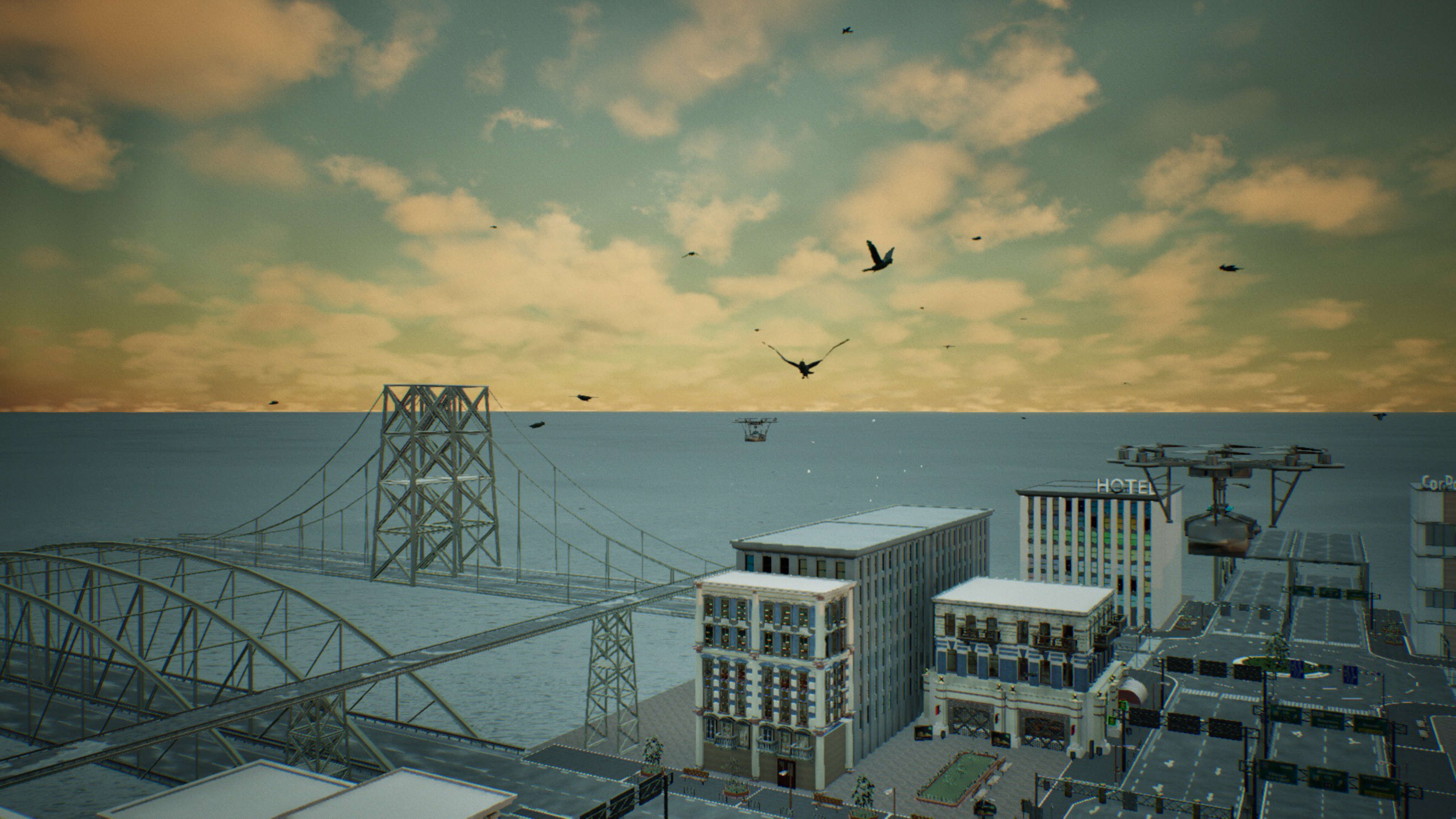} &
\includegraphics[width=0.12\textwidth]{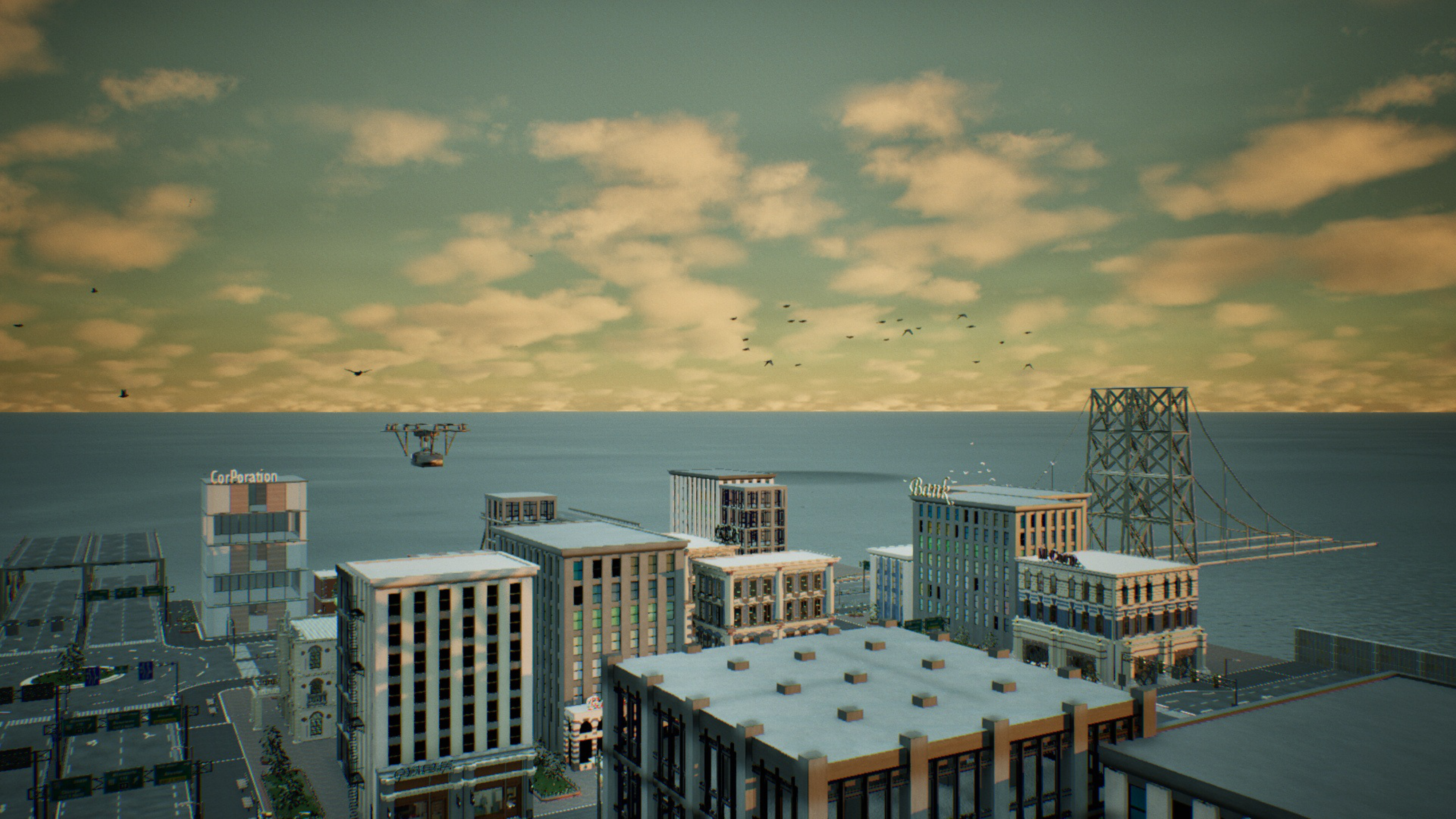} \\

\scriptsize CityPark \cite{UE_CityPark} &
\includegraphics[width=0.12\textwidth]{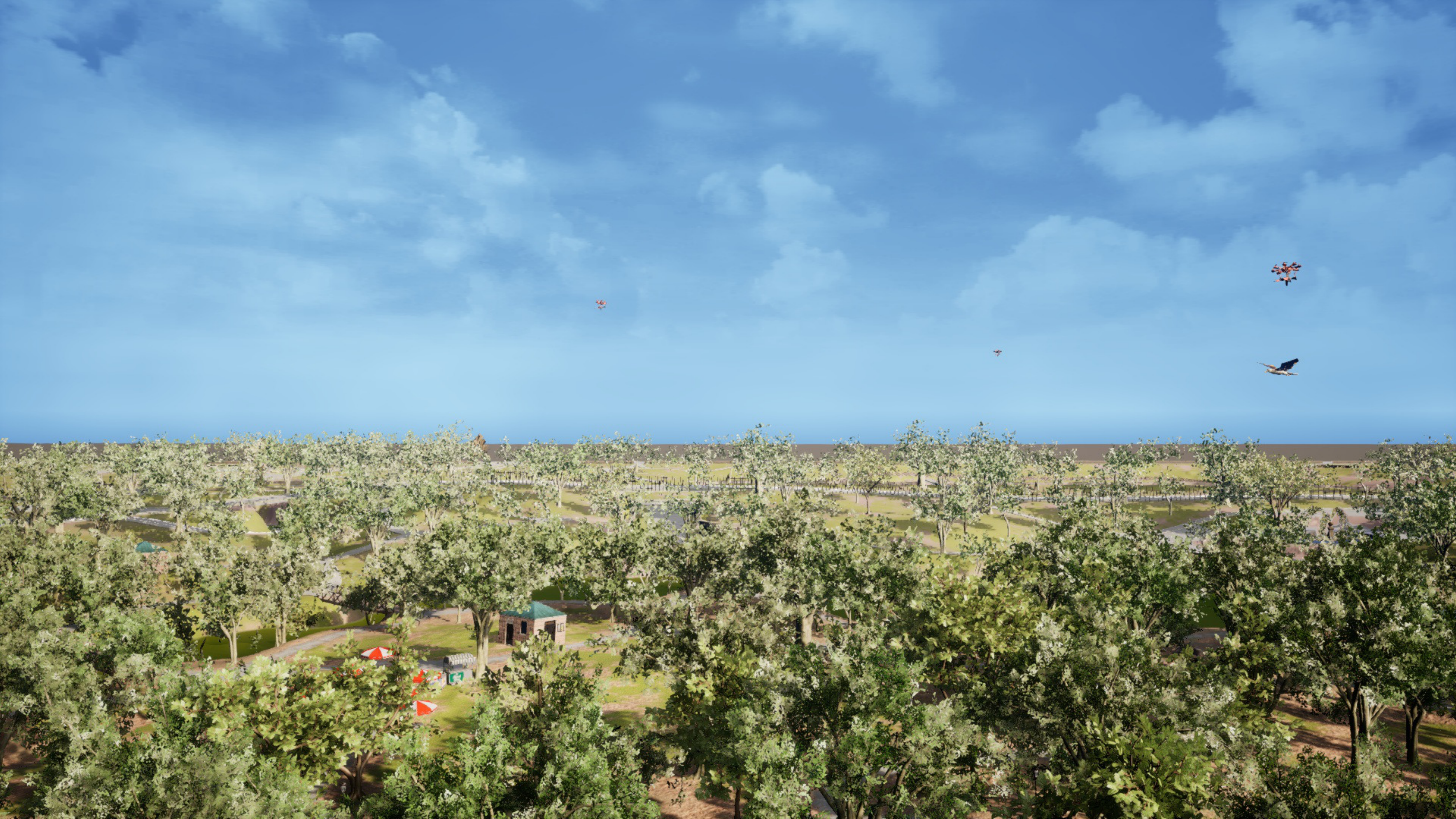} &
\includegraphics[width=0.12\textwidth]{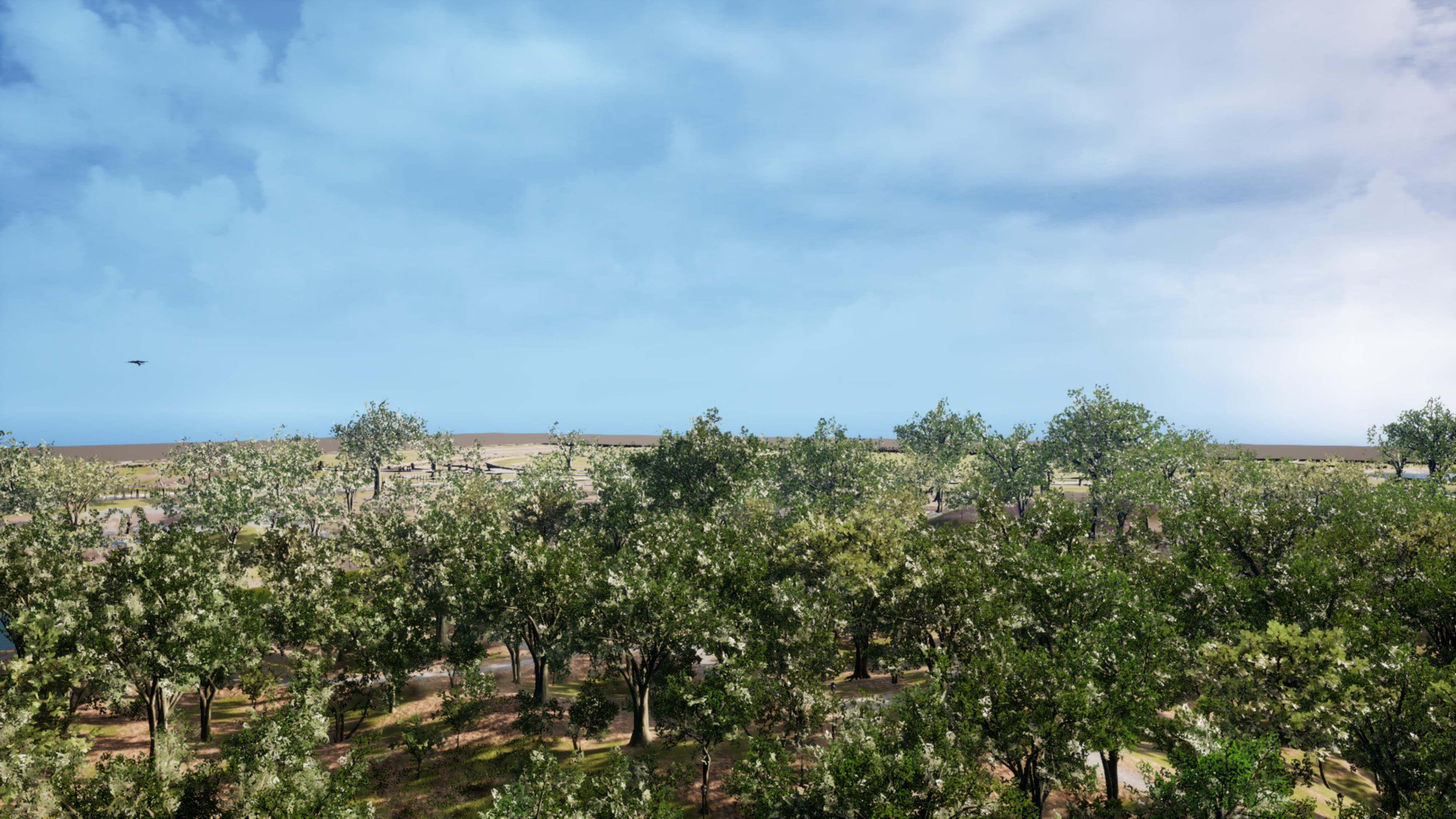} &
\includegraphics[width=0.12\textwidth]{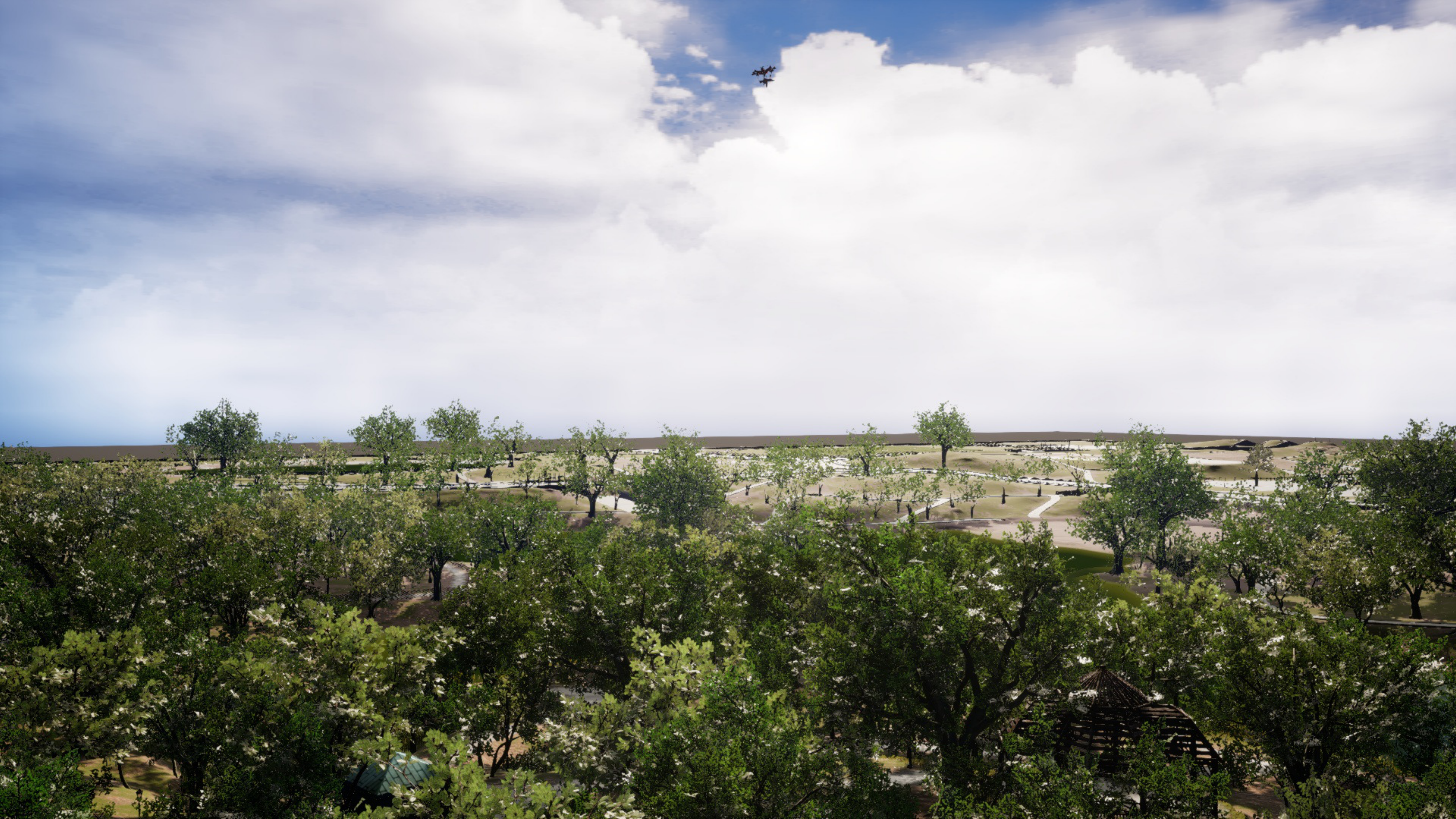} &
\includegraphics[width=0.12\textwidth]{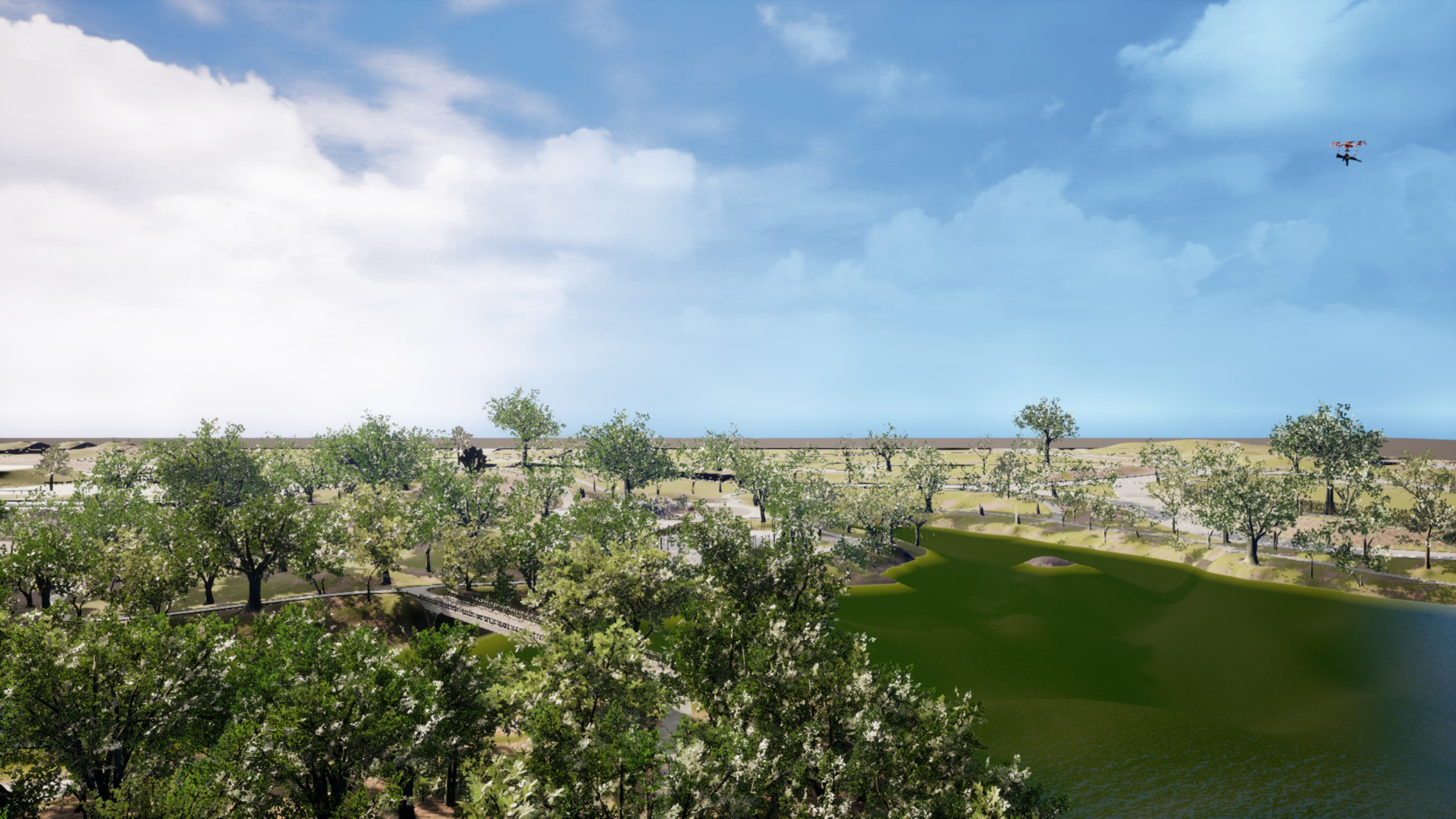} &
\includegraphics[width=0.12\textwidth]{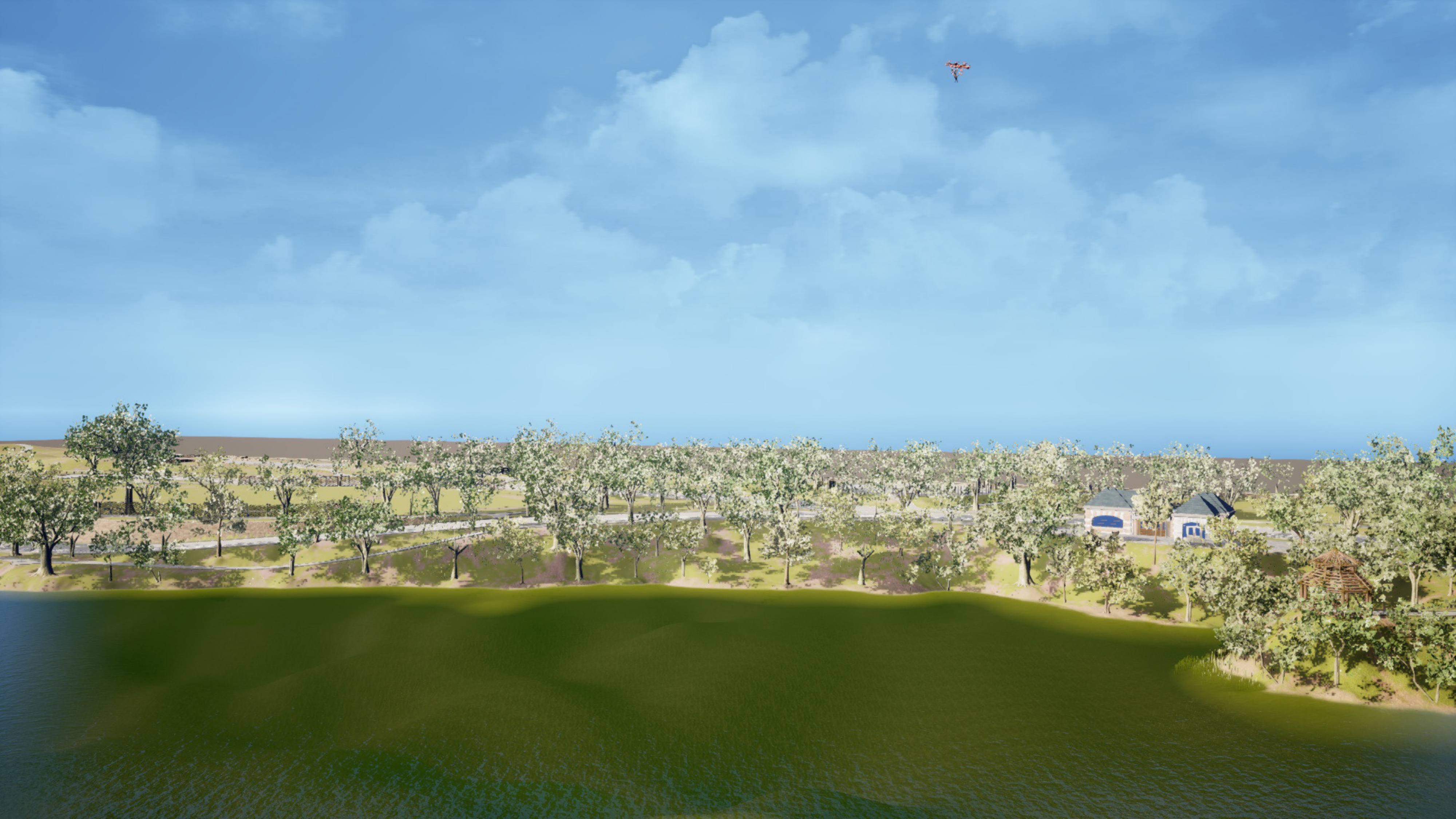} &
\includegraphics[width=0.12\textwidth]{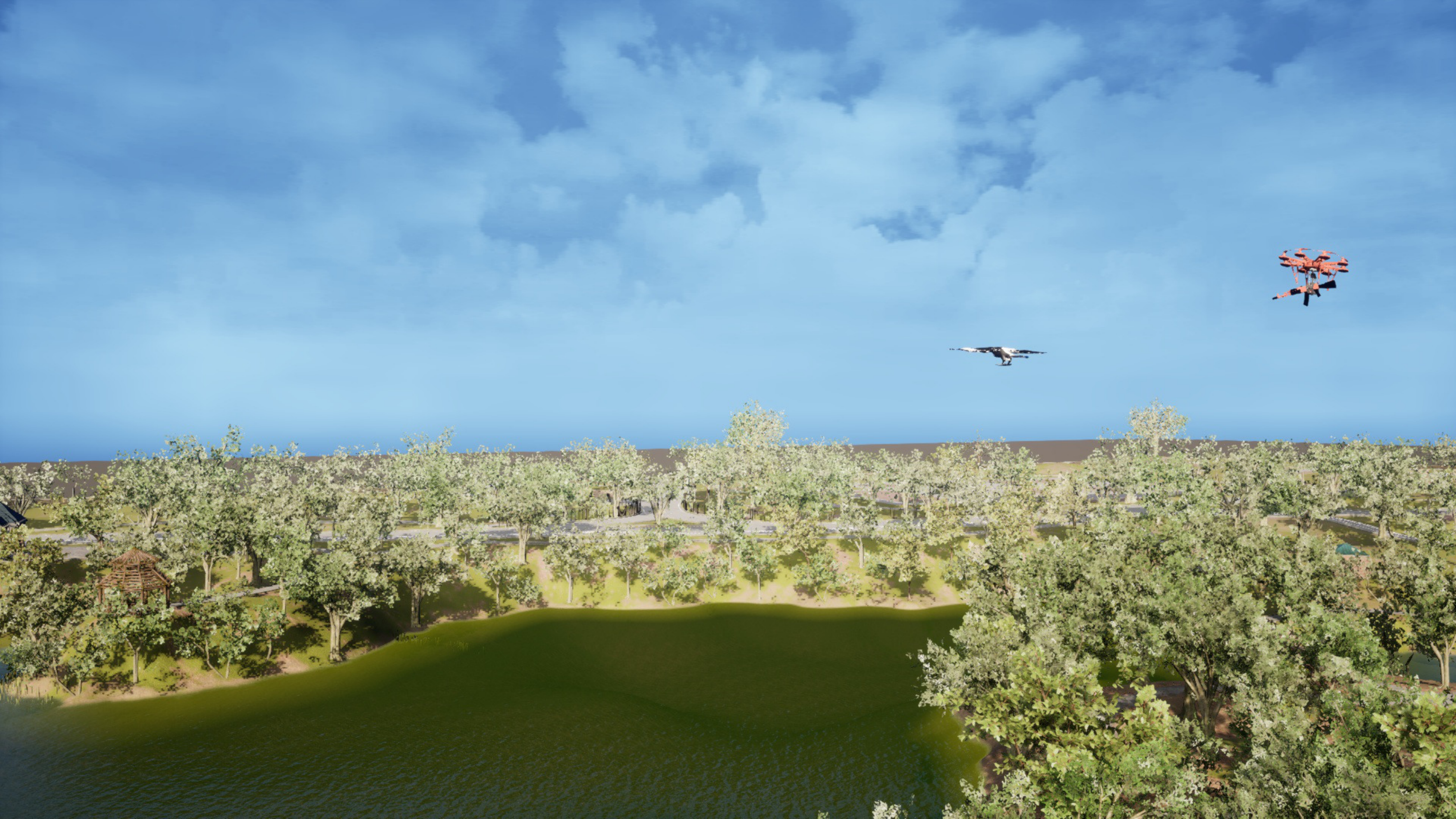} \\

\scriptsize Bridge\cite{UE_Bridge} &
\includegraphics[width=0.12\textwidth]{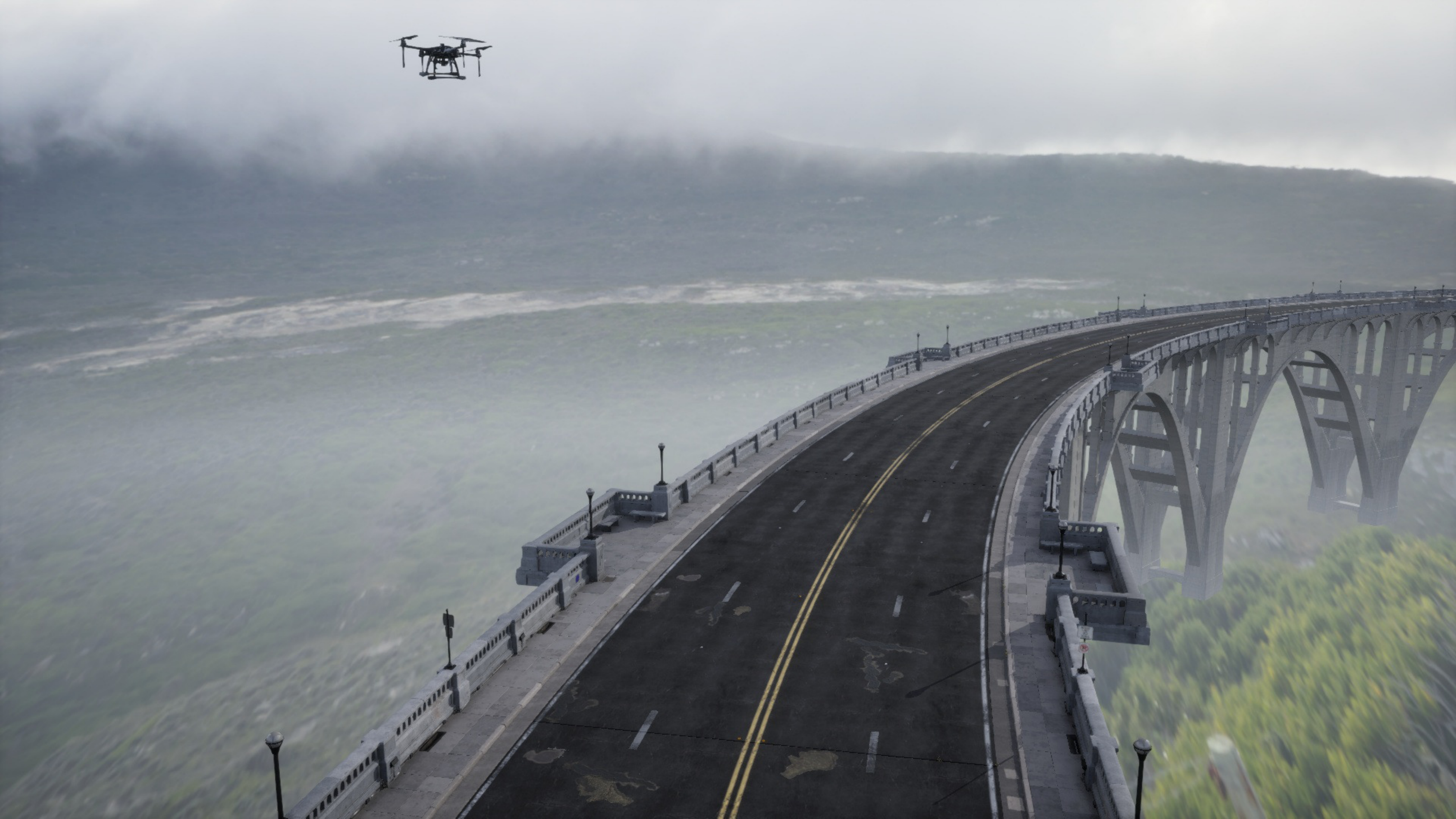} &
\includegraphics[width=0.12\textwidth]{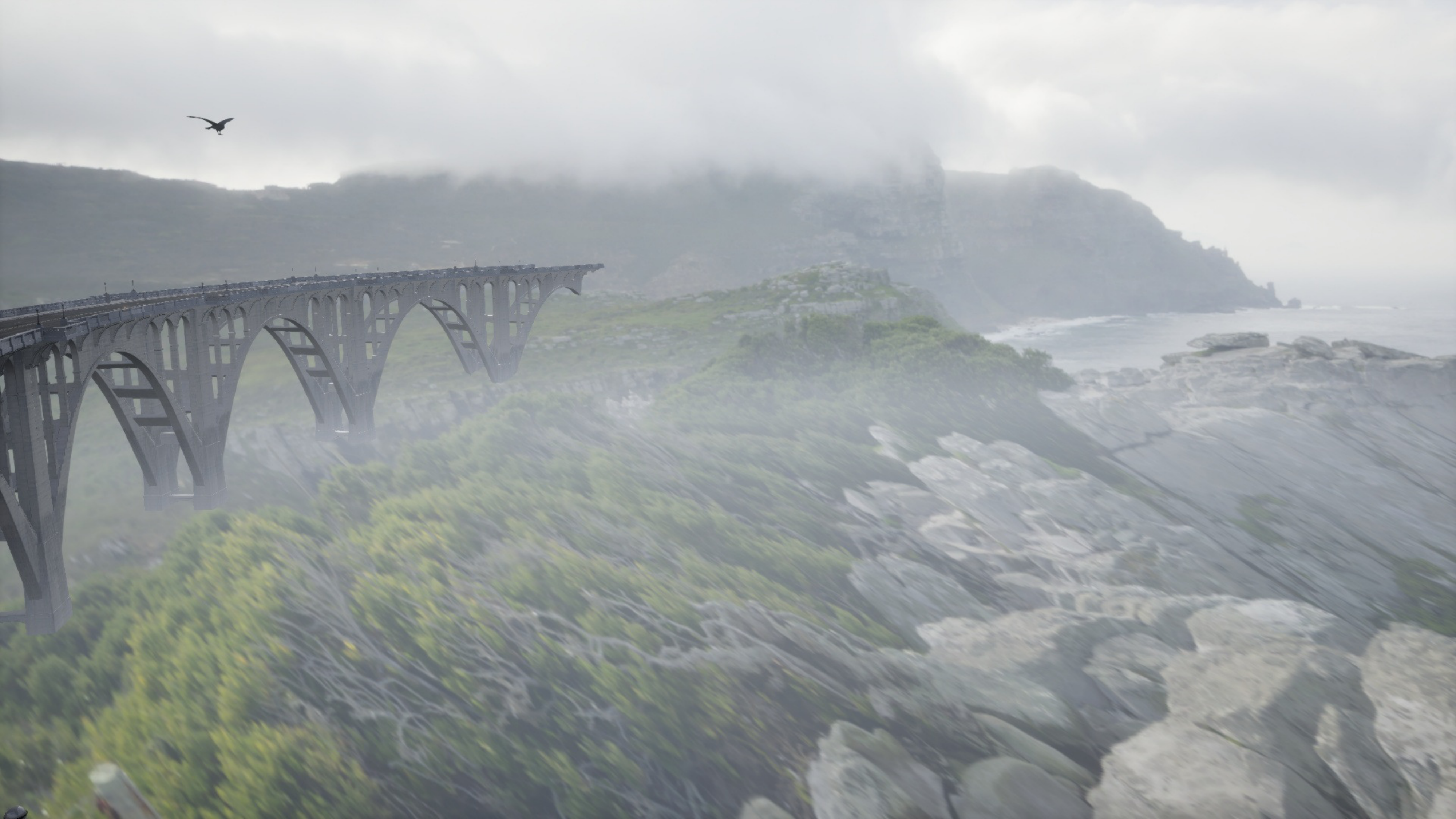} &
\includegraphics[width=0.12\textwidth]{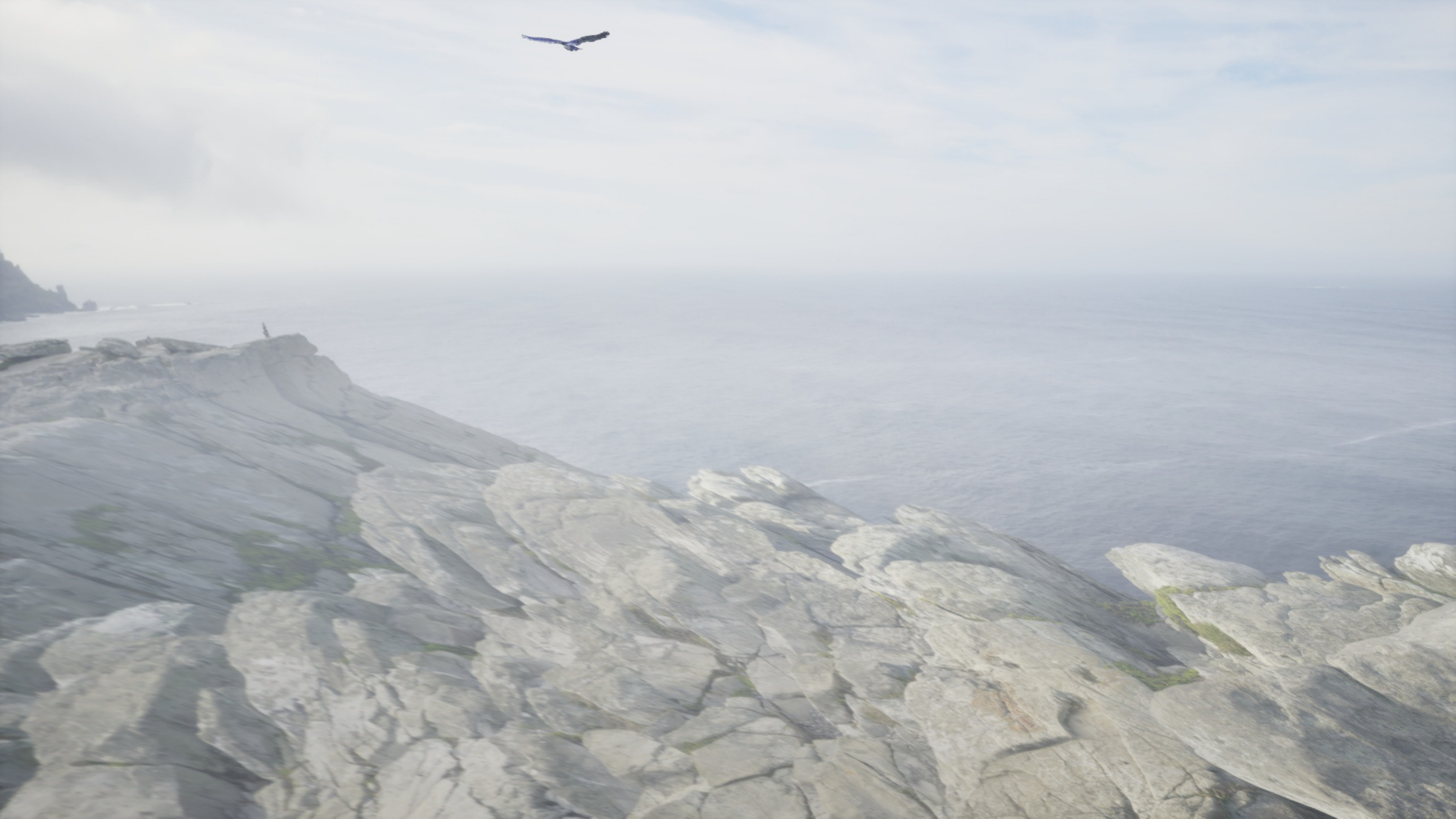} &
\includegraphics[width=0.12\textwidth]{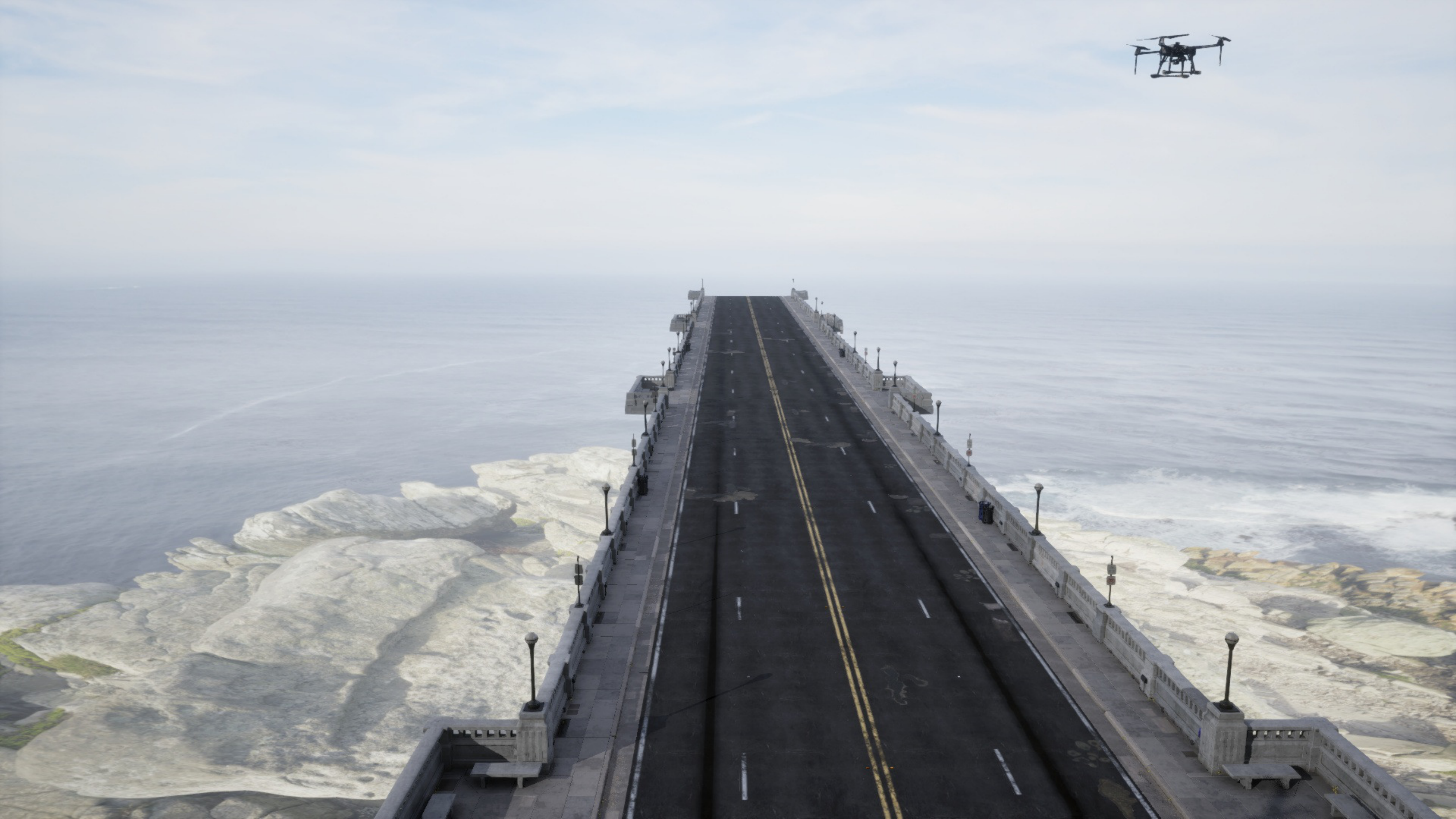} &
\includegraphics[width=0.12\textwidth]{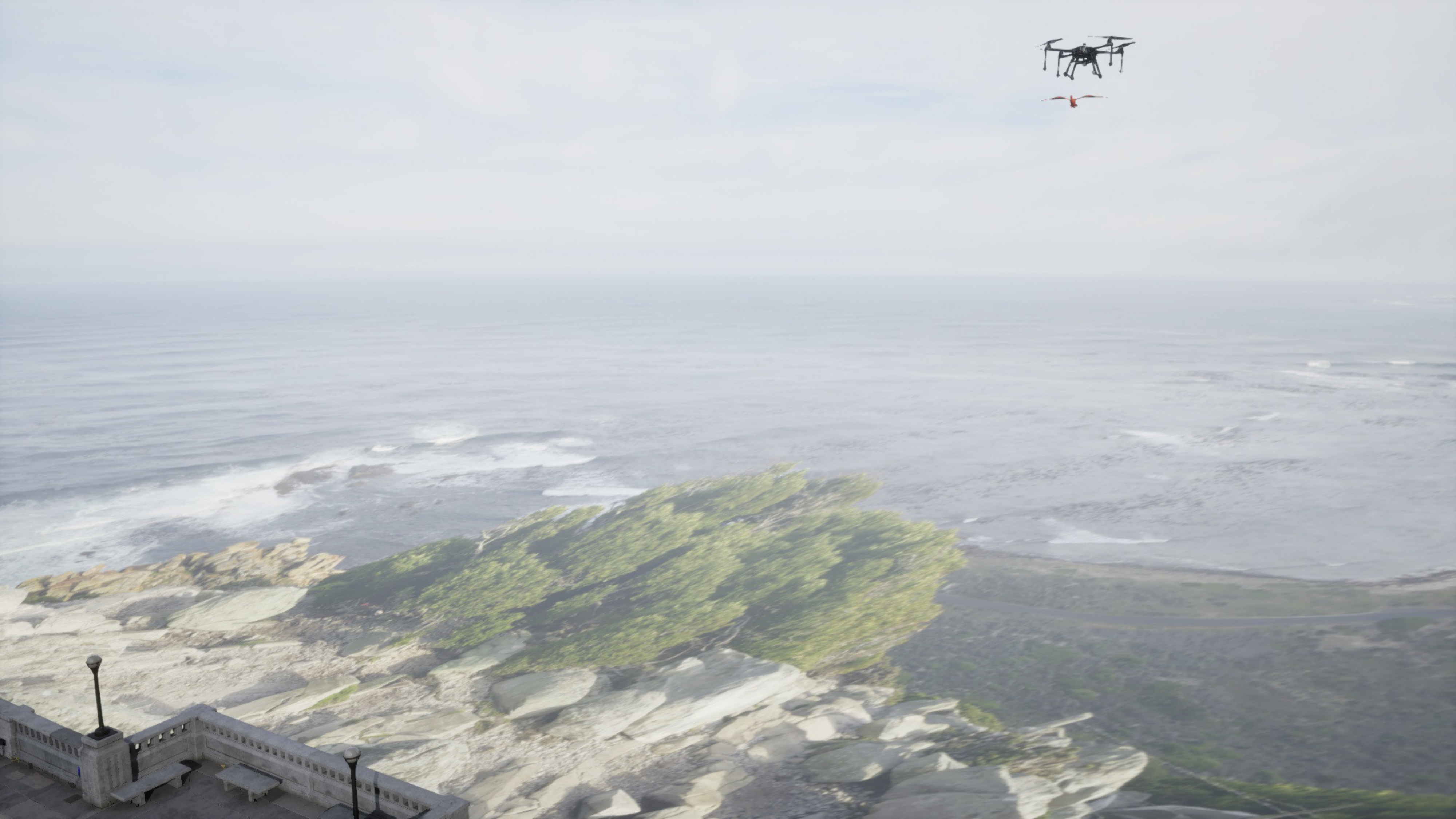} &
\includegraphics[width=0.12\textwidth]{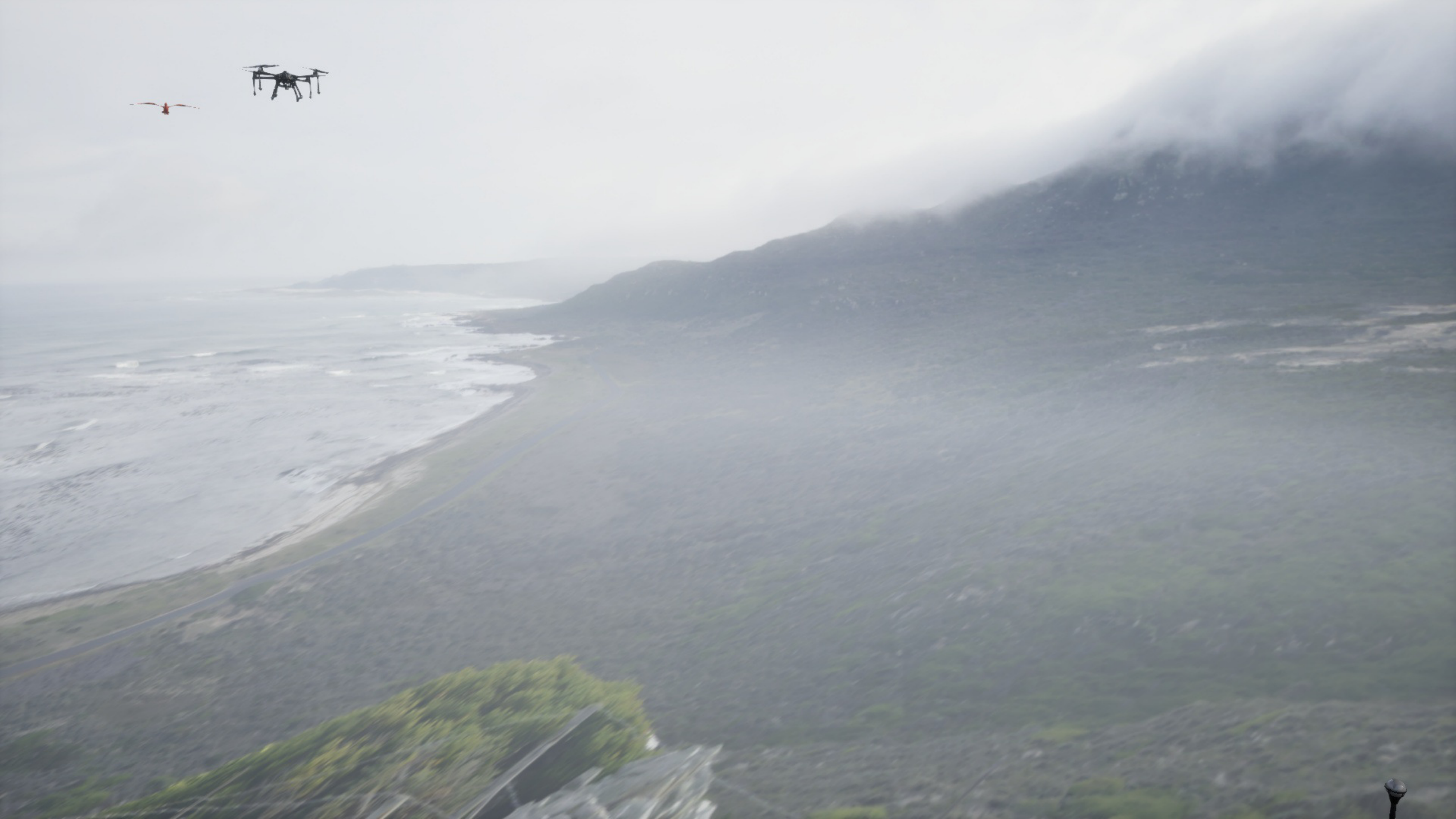} \\

\scriptsize WildWestTown \cite{UE_WildWestTown} &
\includegraphics[width=0.12\textwidth]{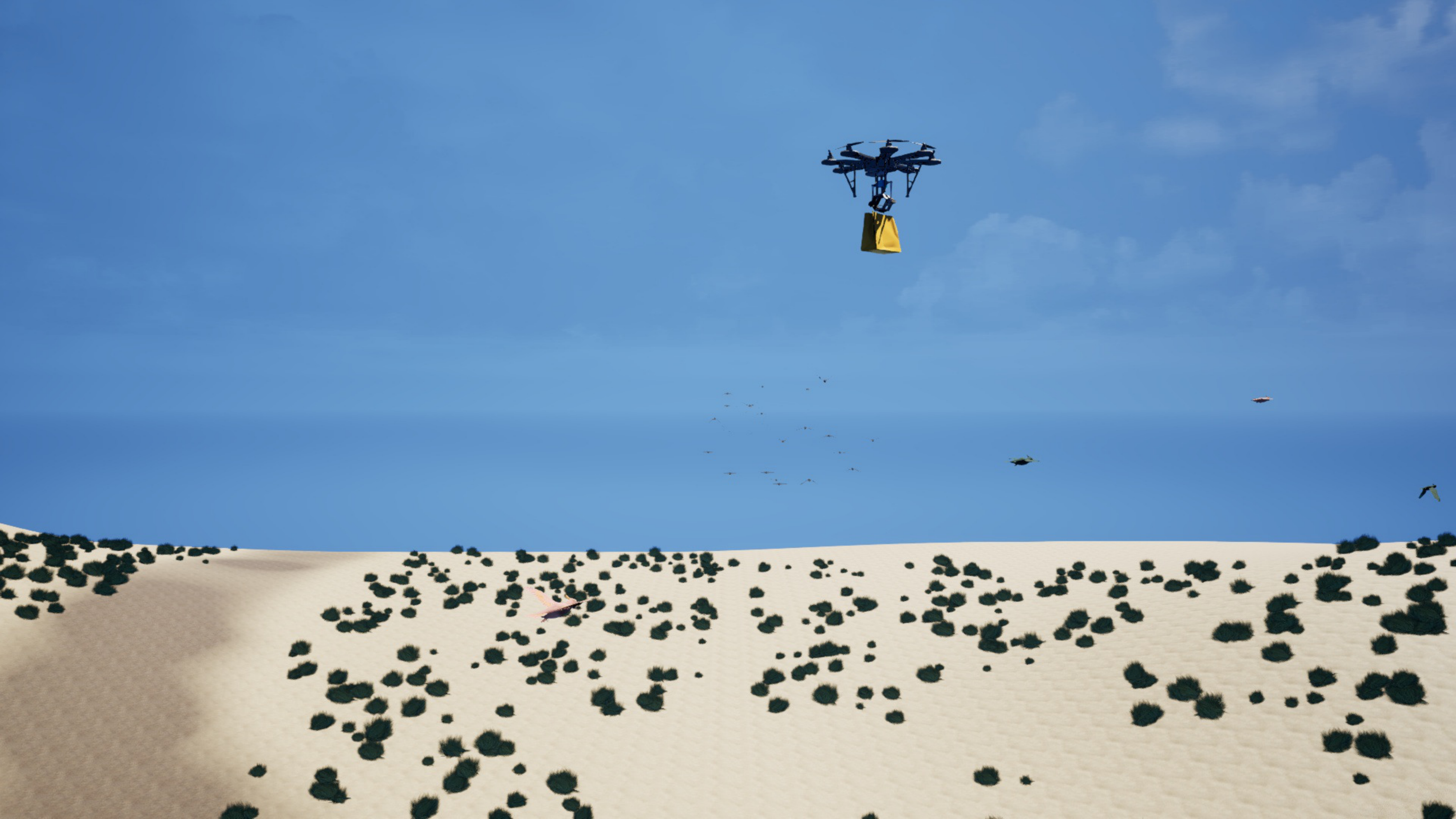} &
\includegraphics[width=0.12\textwidth]{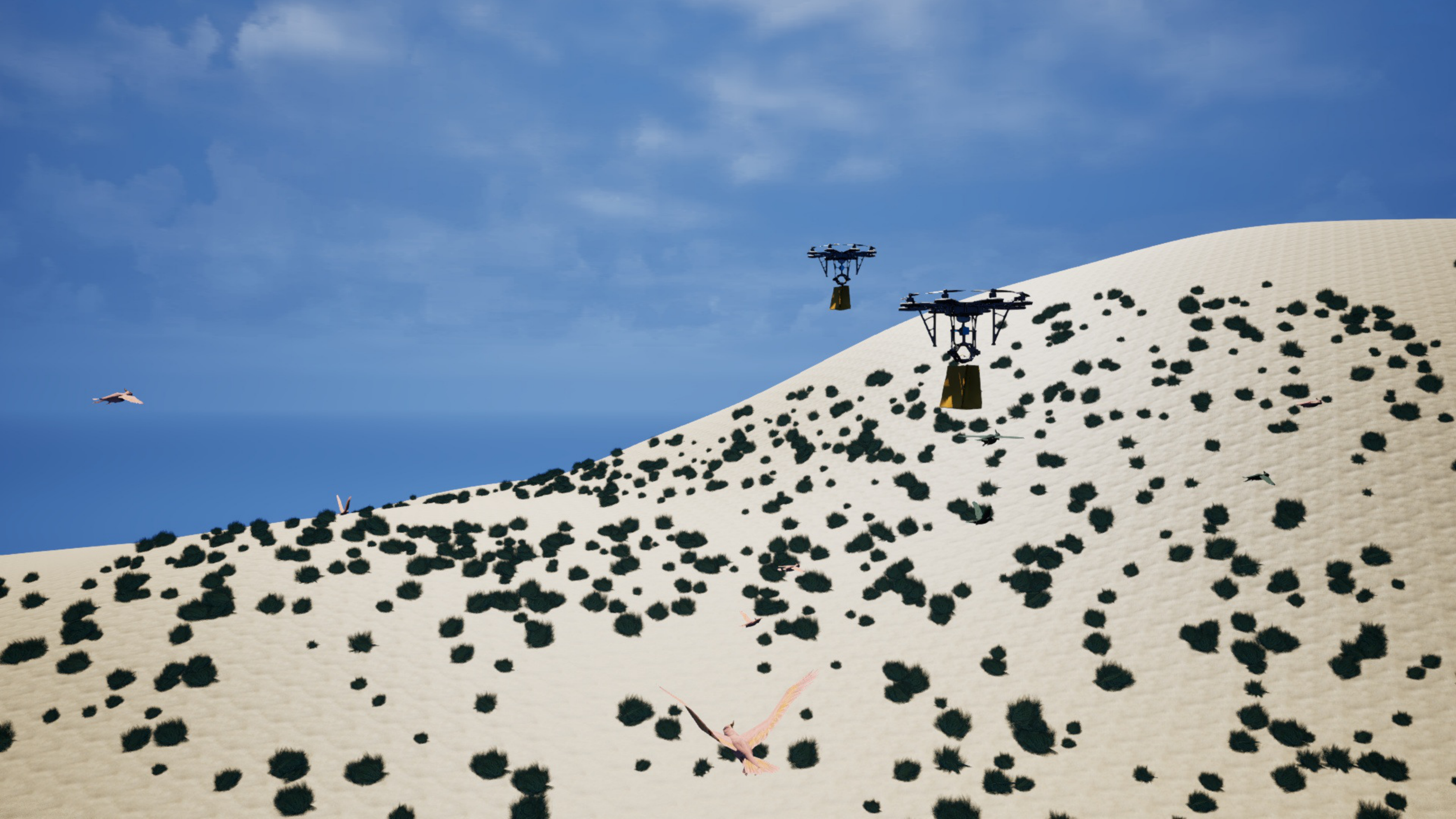} &
\includegraphics[width=0.12\textwidth]{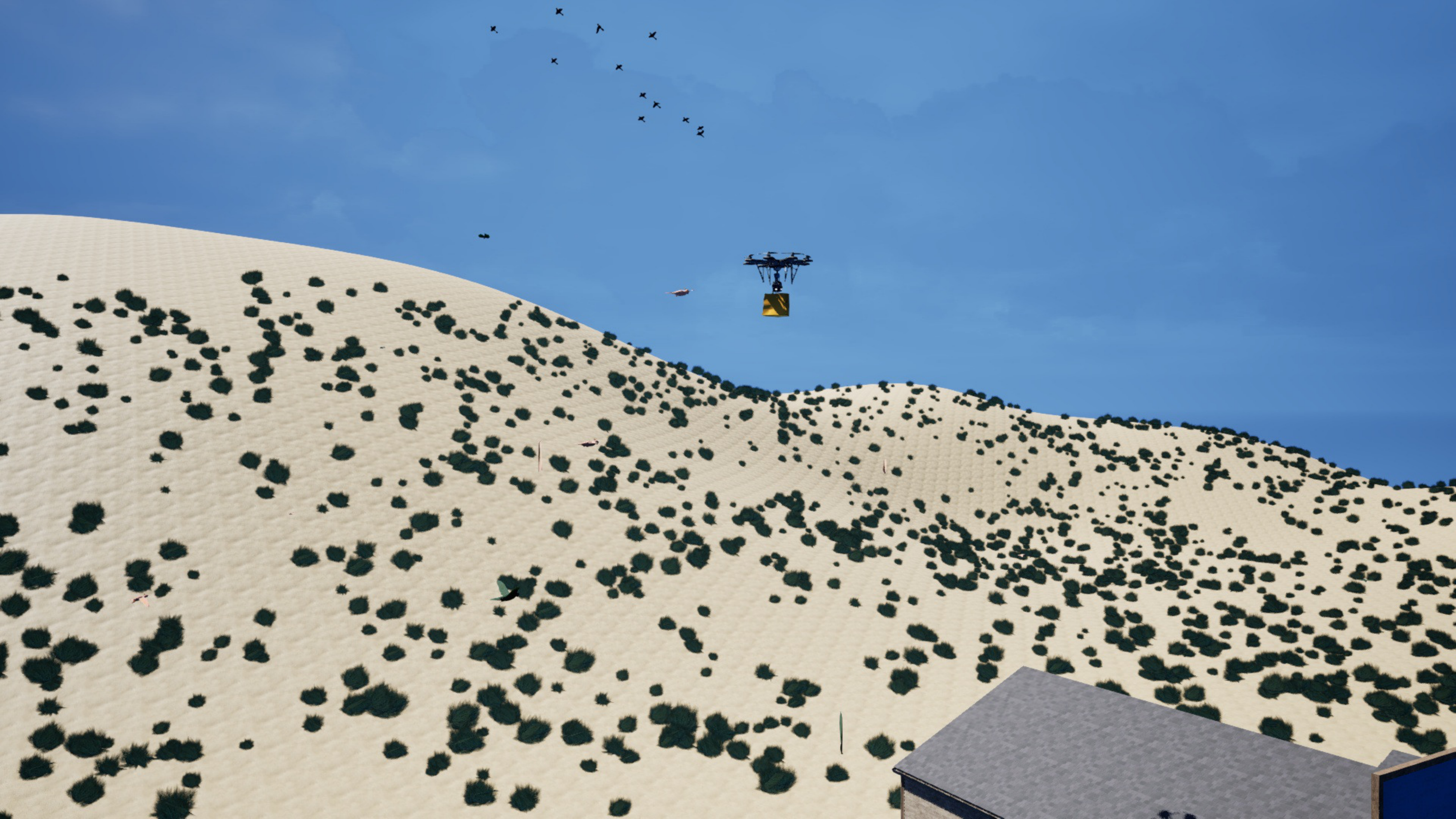} &
\includegraphics[width=0.12\textwidth]{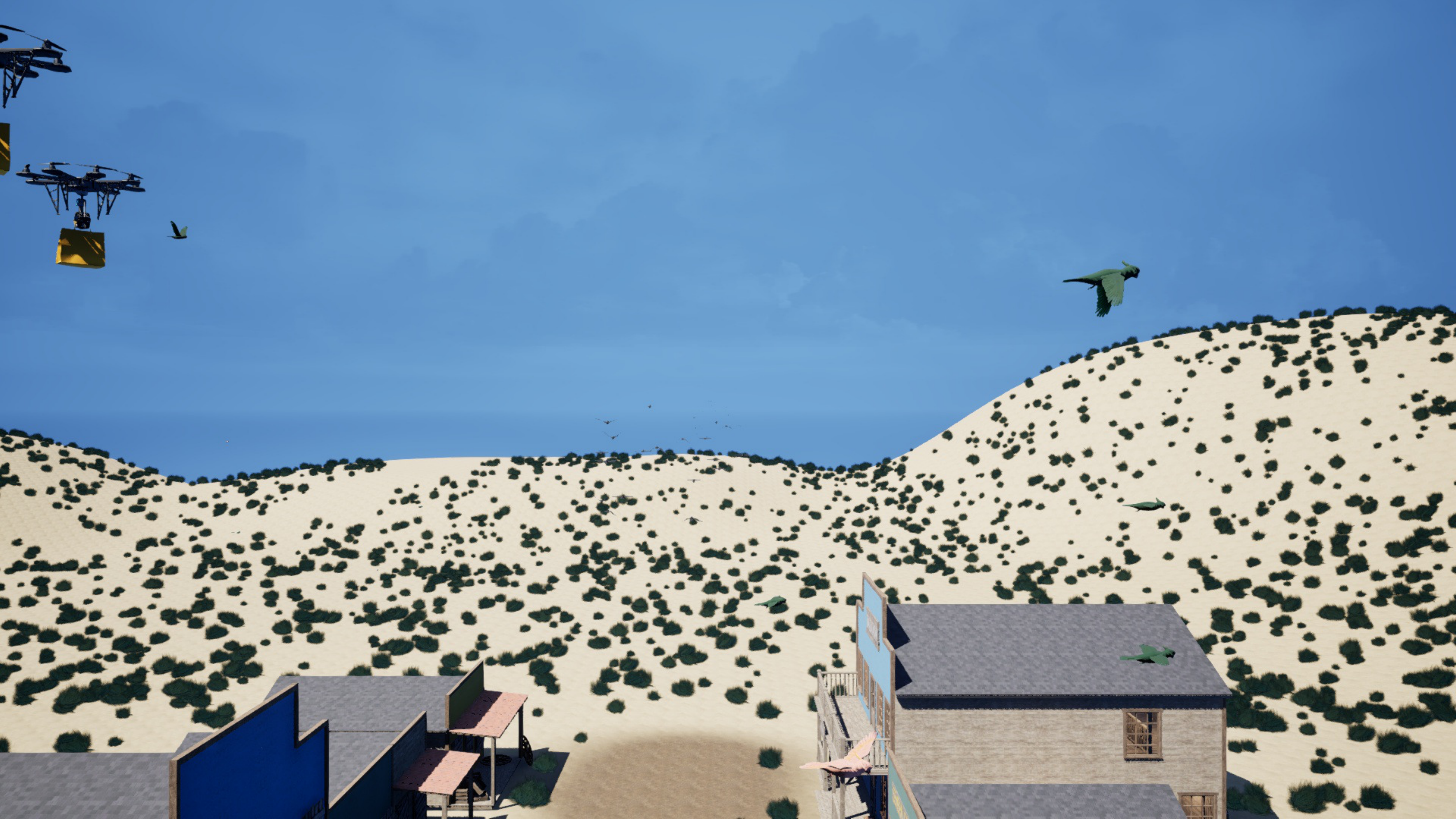} &
\includegraphics[width=0.12\textwidth]{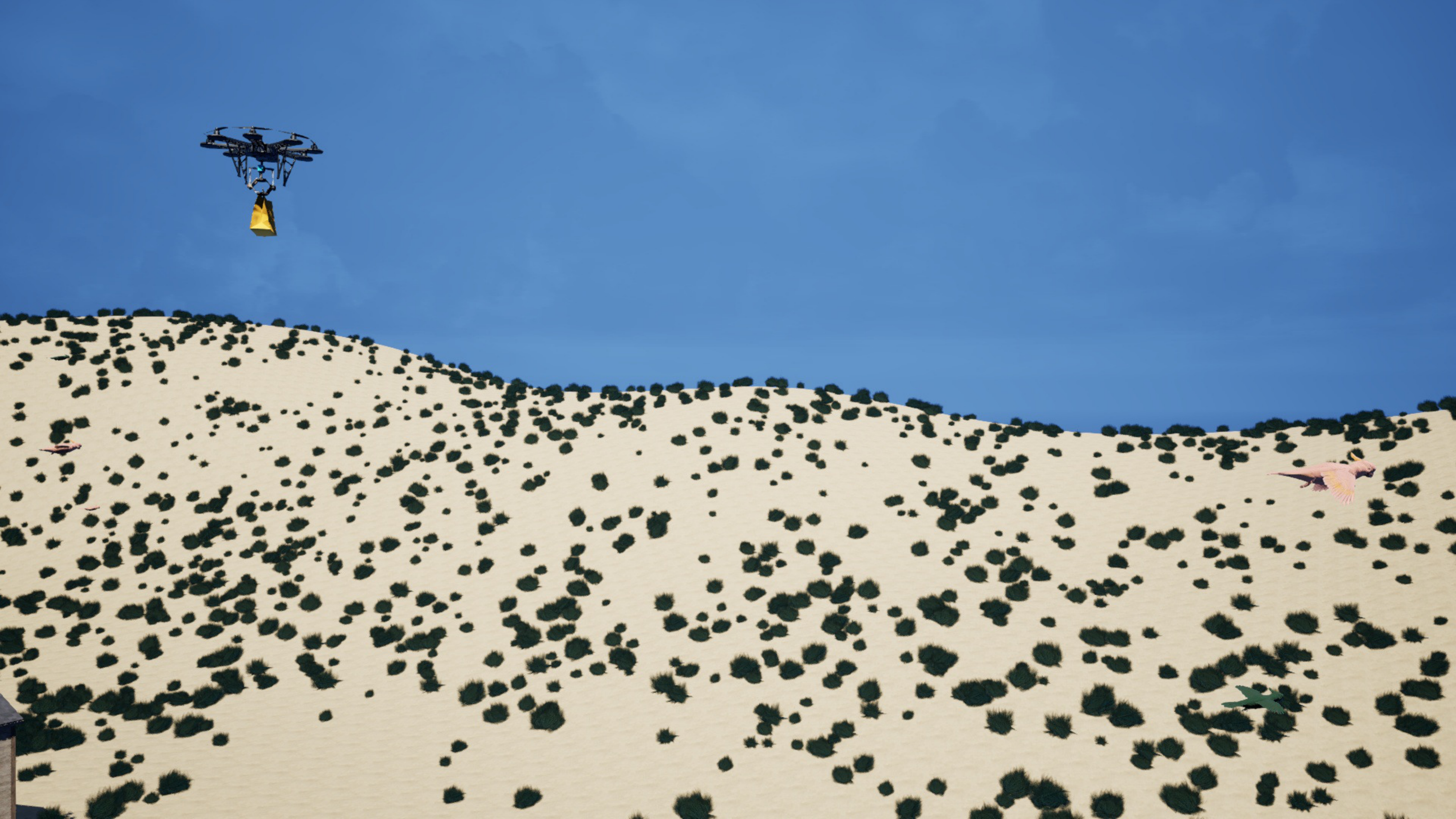} &
\includegraphics[width=0.12\textwidth]{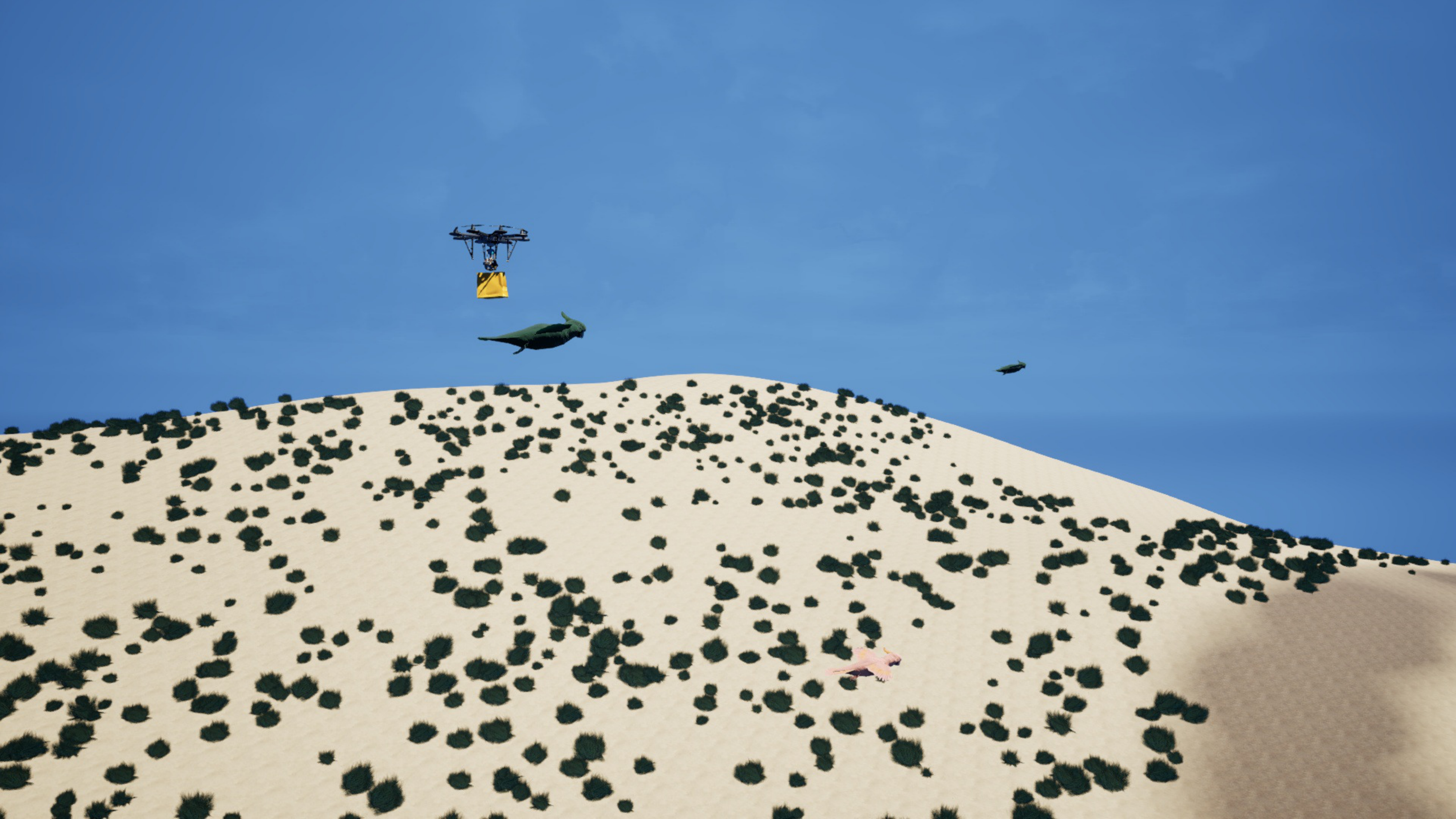} \\

\scriptsize DownTown \cite{UE_Downtown}&
\includegraphics[width=0.12\textwidth]{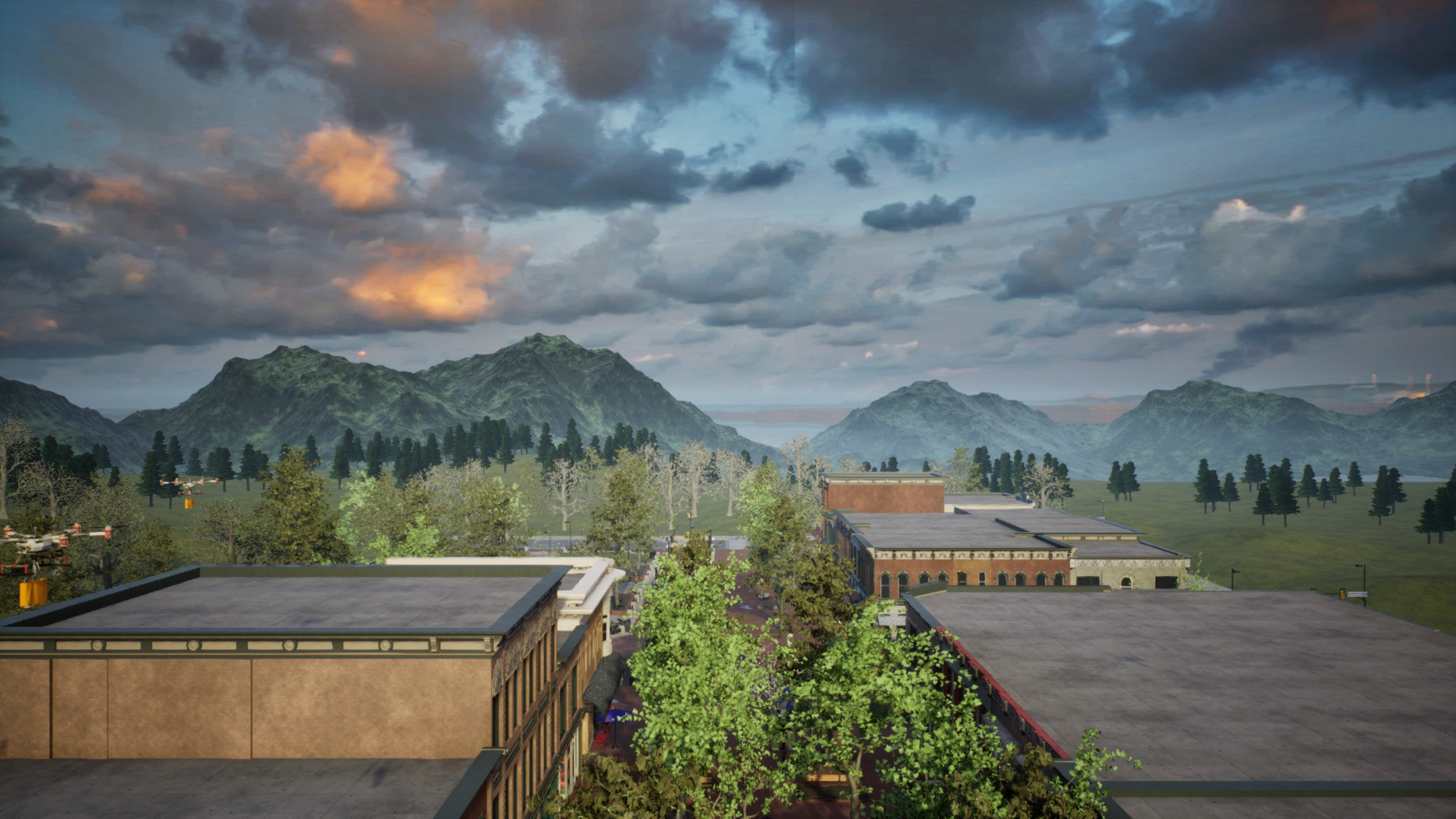} &
\includegraphics[width=0.12\textwidth]{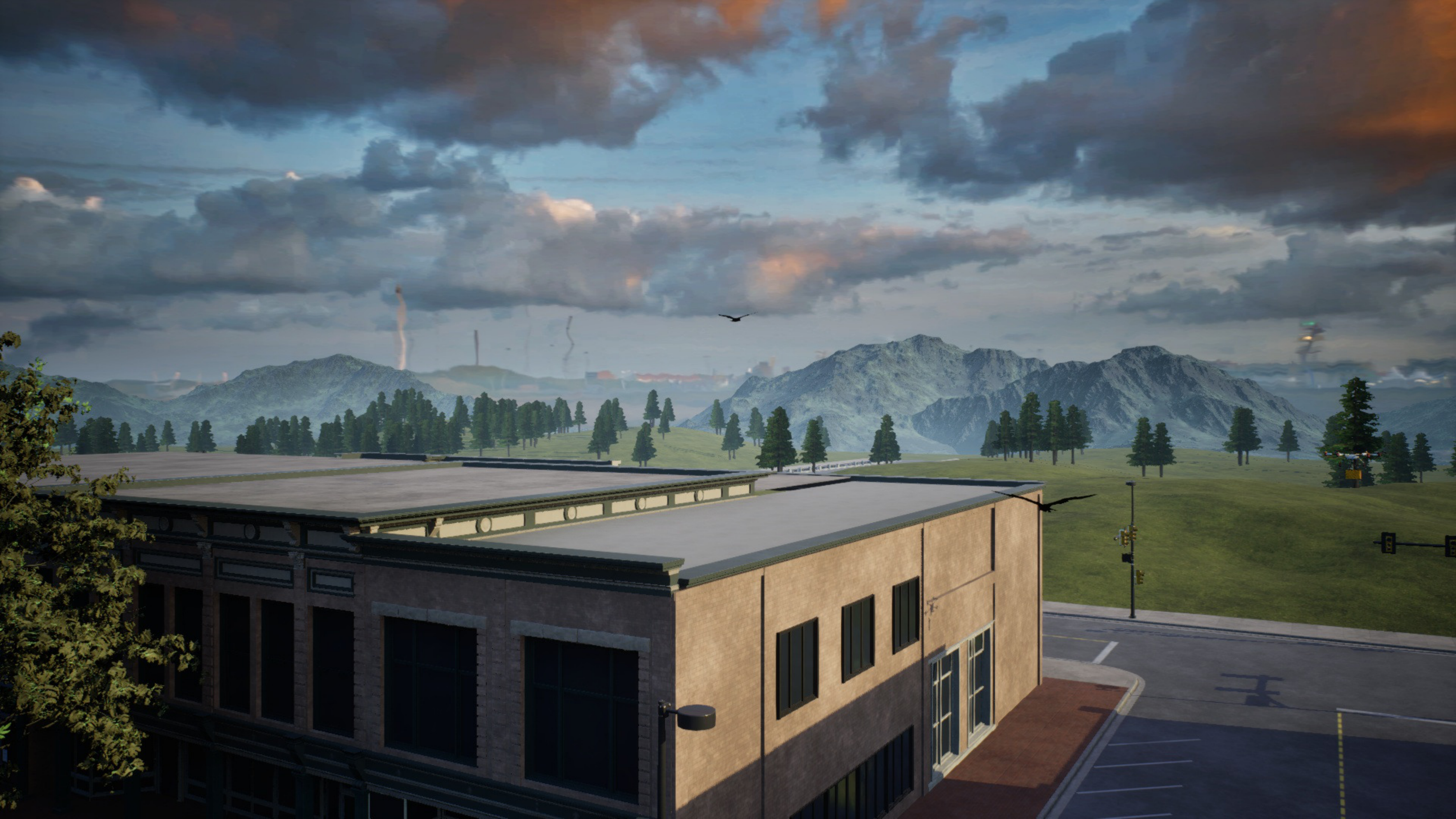} &
\includegraphics[width=0.12\textwidth]{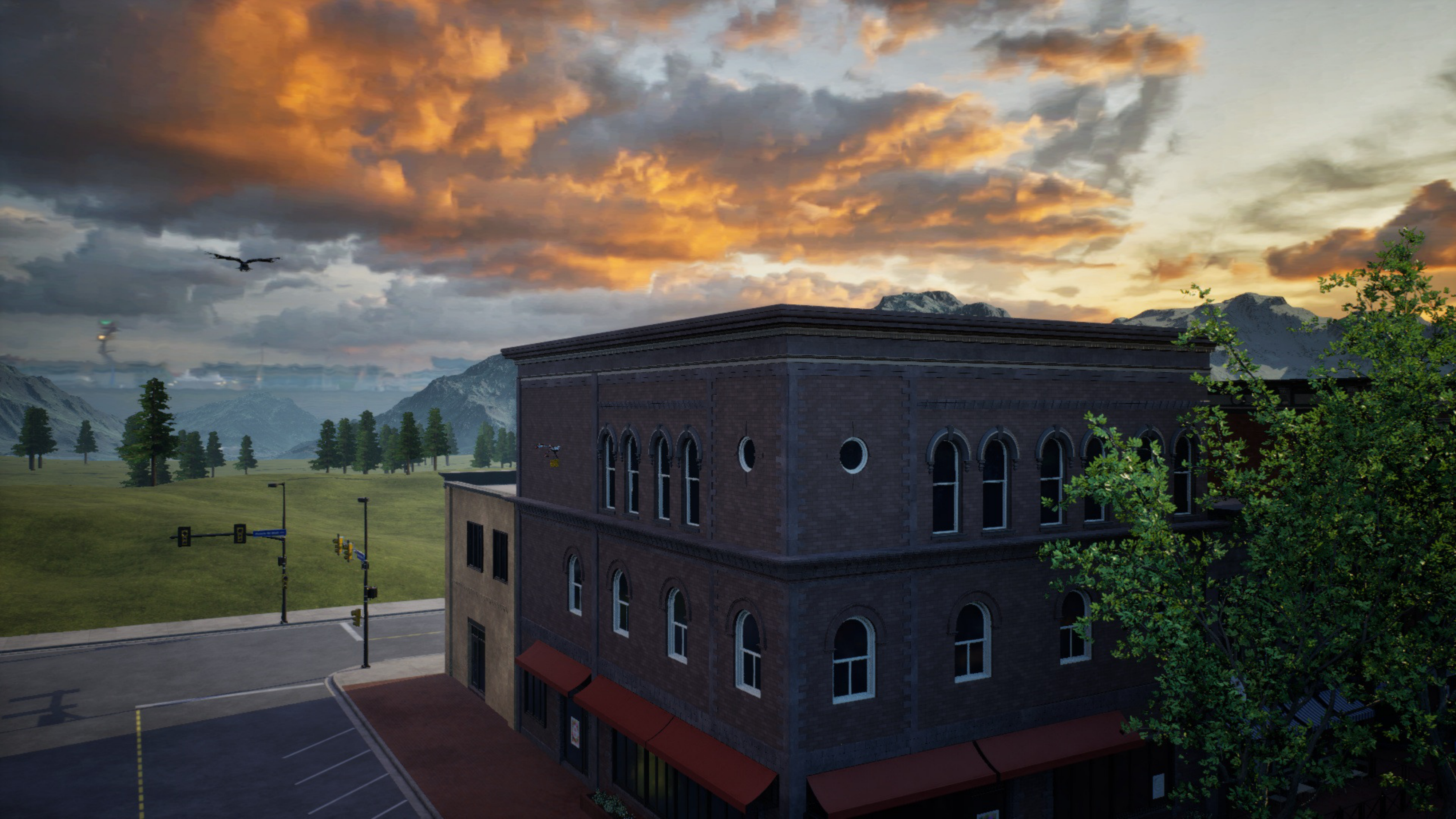} &
\includegraphics[width=0.12\textwidth]{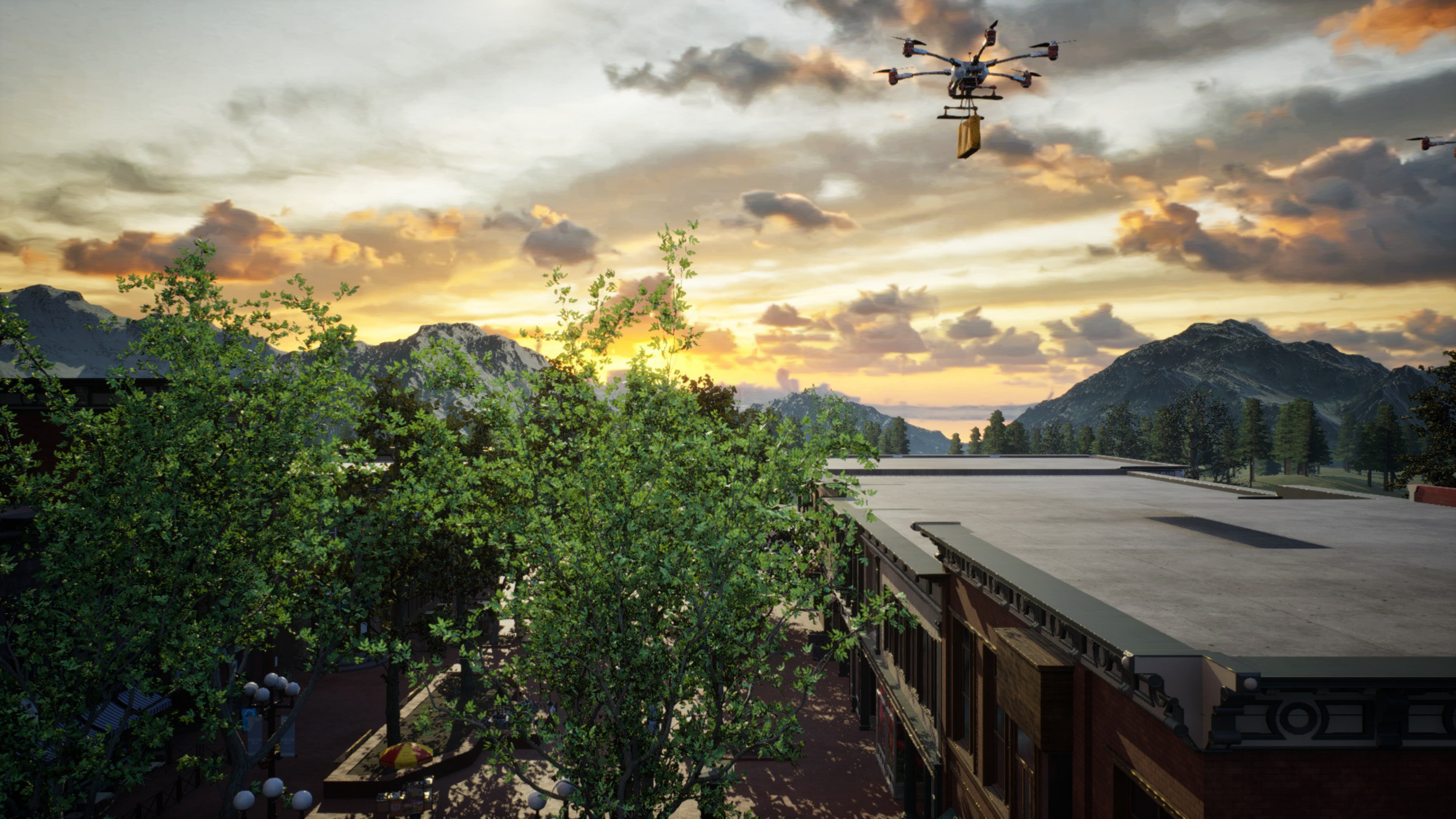} &
\includegraphics[width=0.12\textwidth]{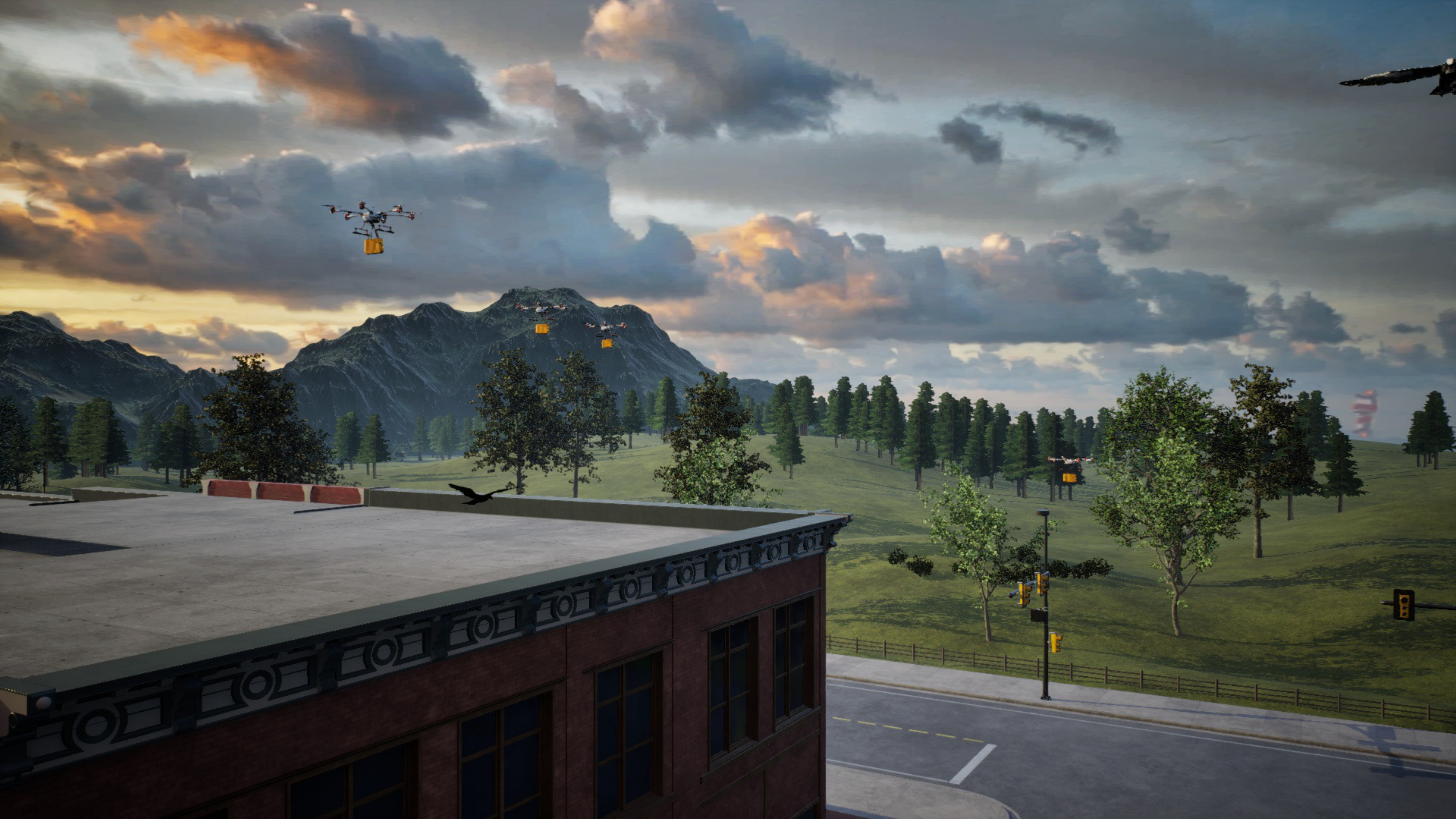} &
\includegraphics[width=0.12\textwidth]{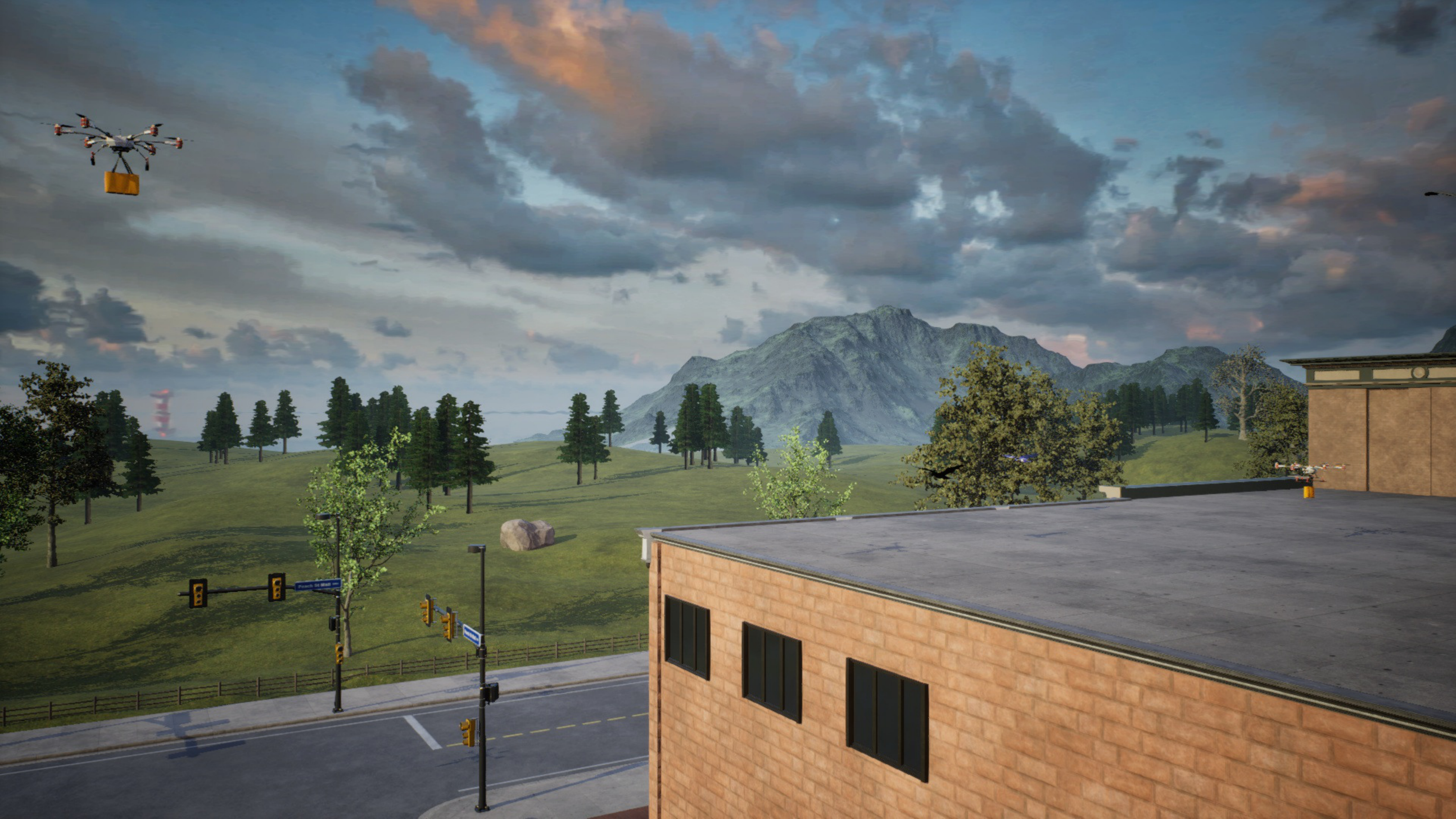} \\

\scriptsize DynamicCity \cite{UE_DynamicCity} &
\includegraphics[width=0.12\textwidth]{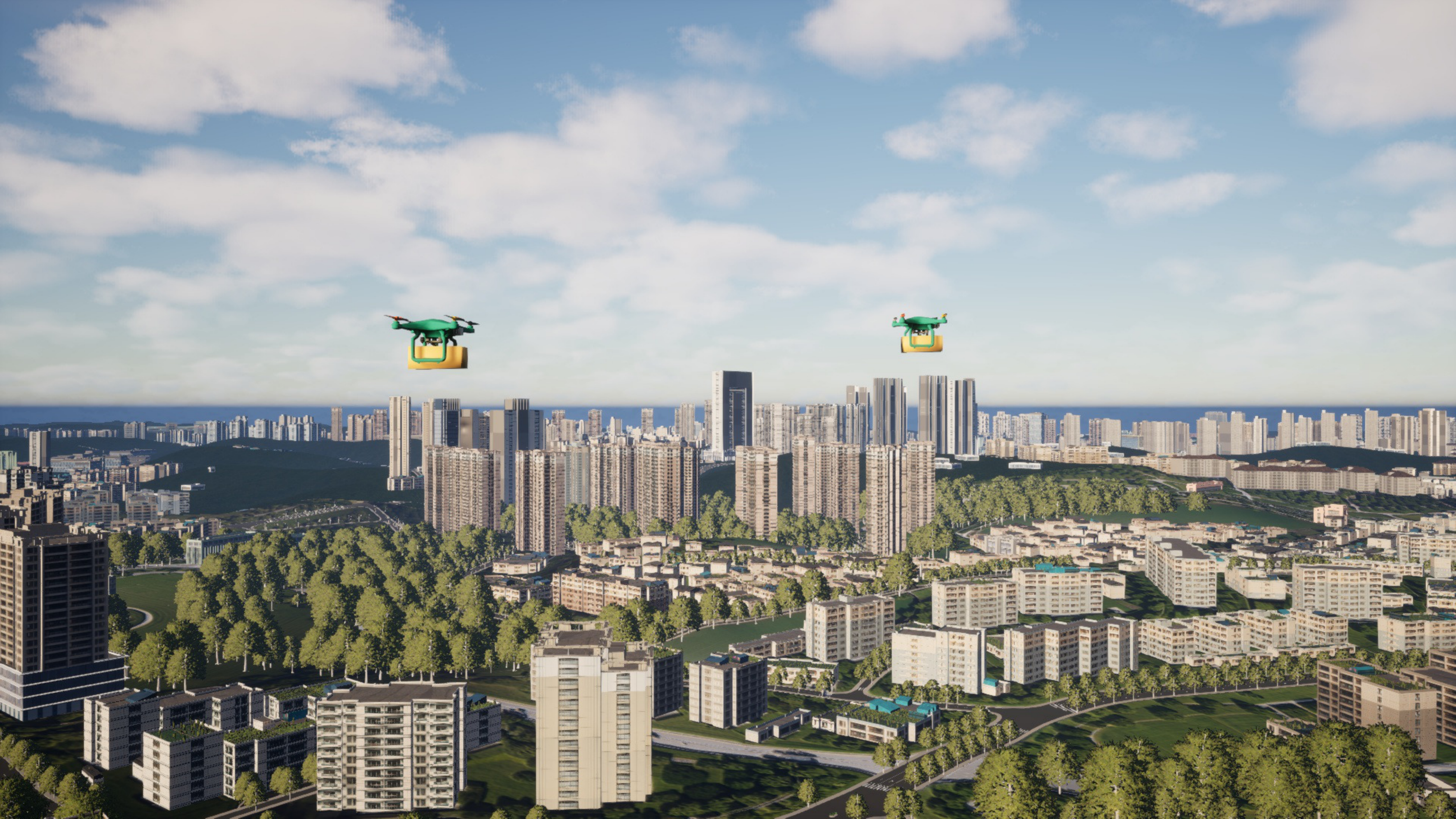} &
\includegraphics[width=0.12\textwidth]{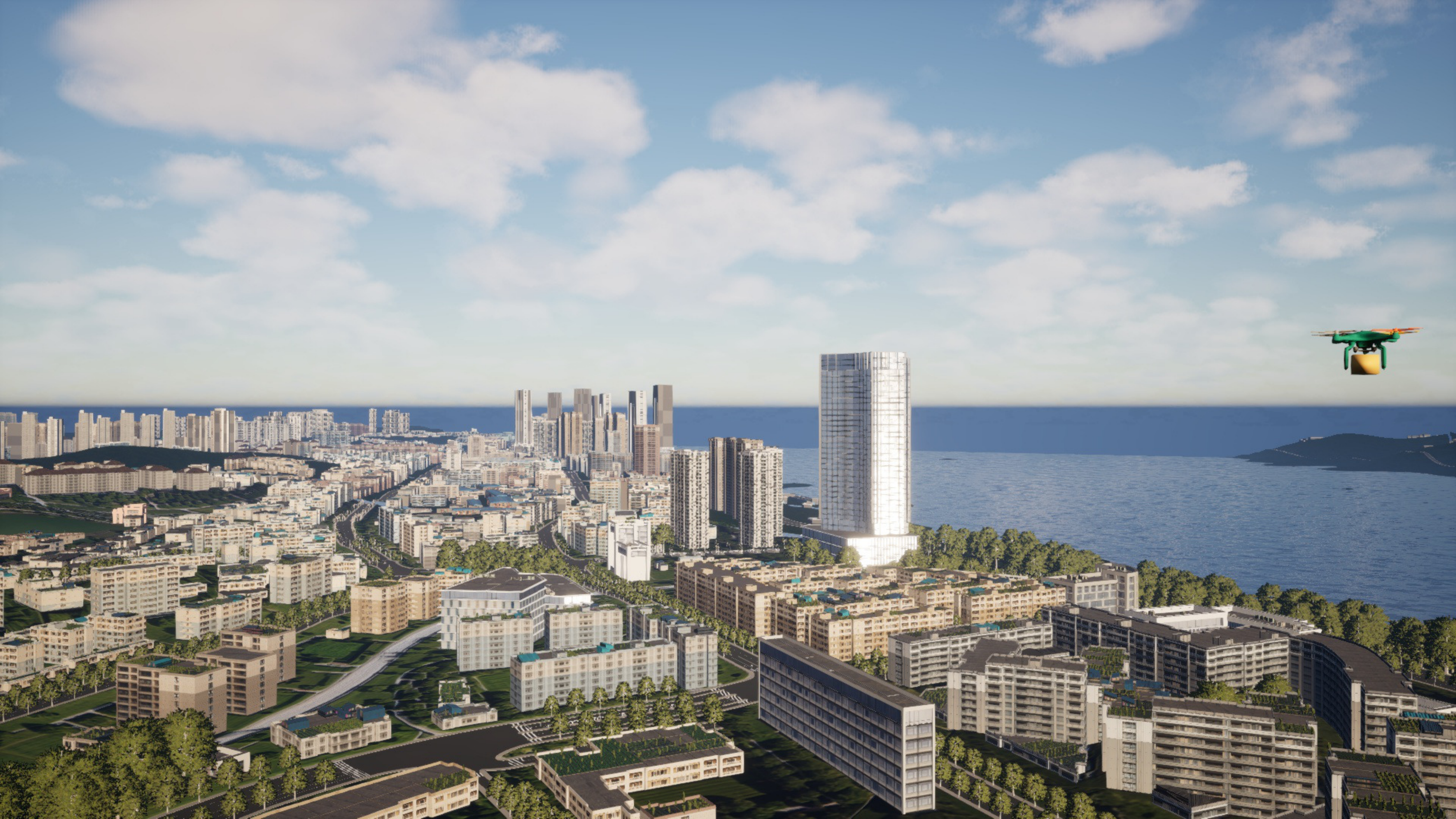} &
\includegraphics[width=0.12\textwidth]{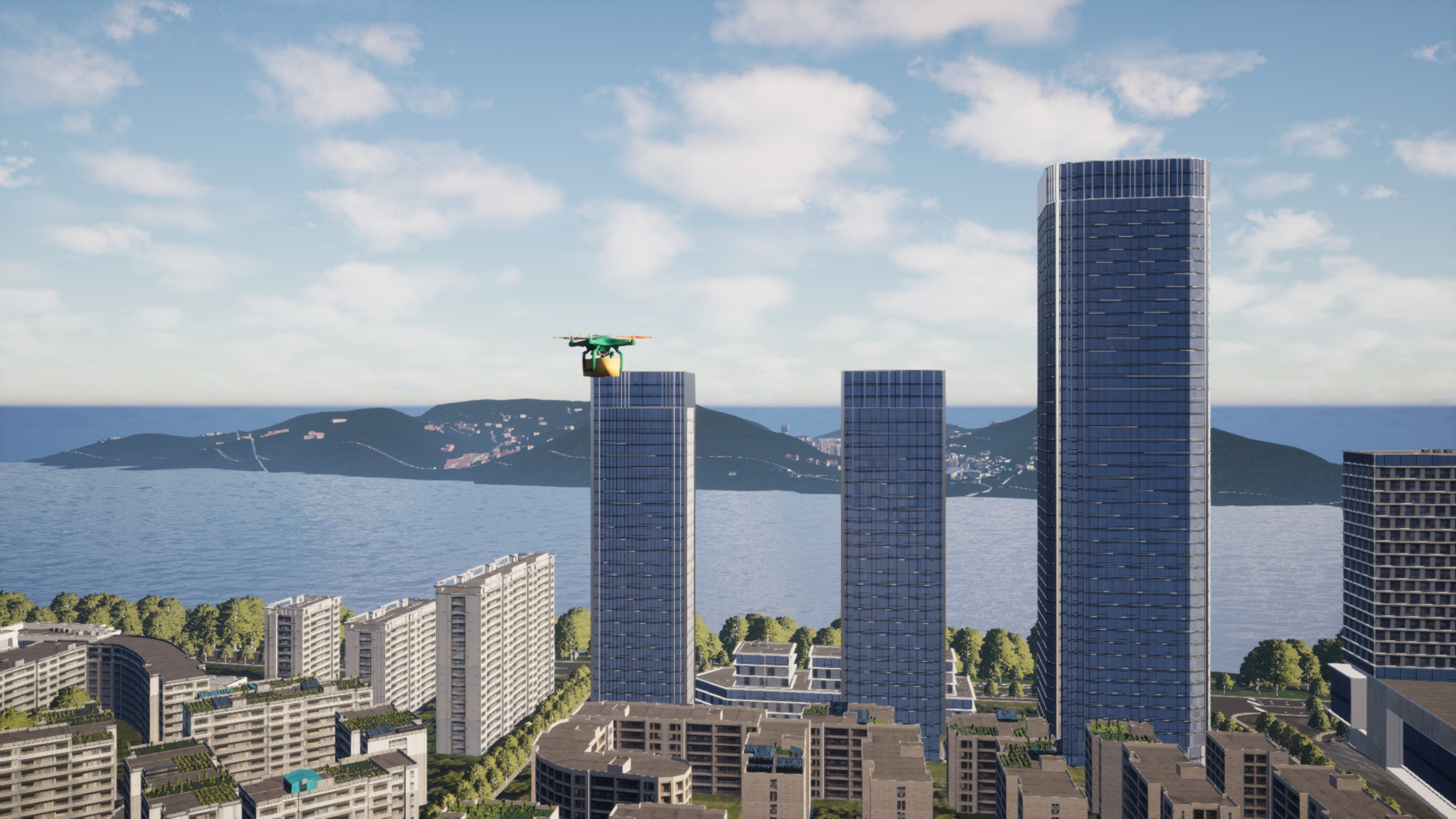} &
\includegraphics[width=0.12\textwidth]{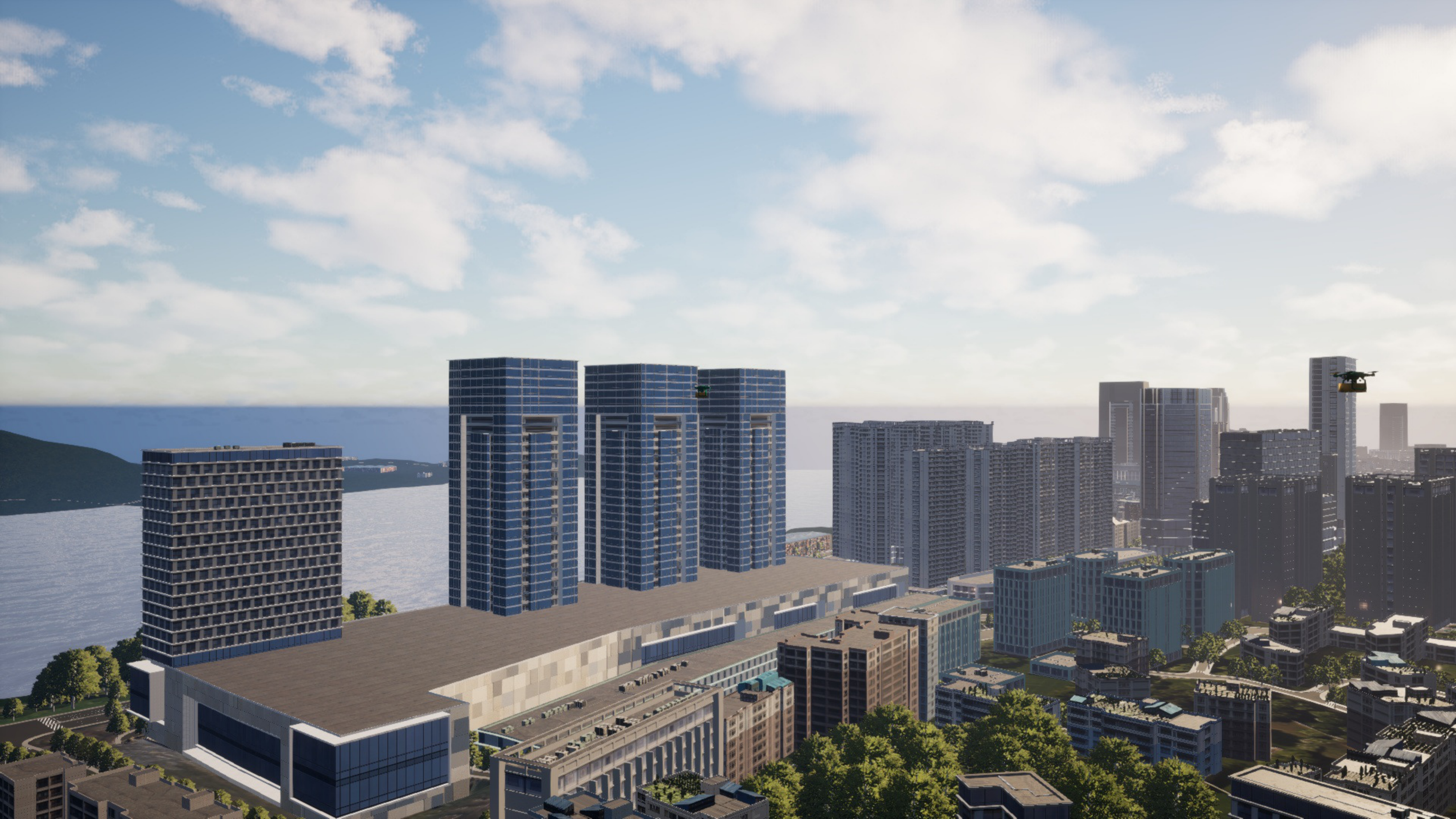} &
\includegraphics[width=0.12\textwidth]{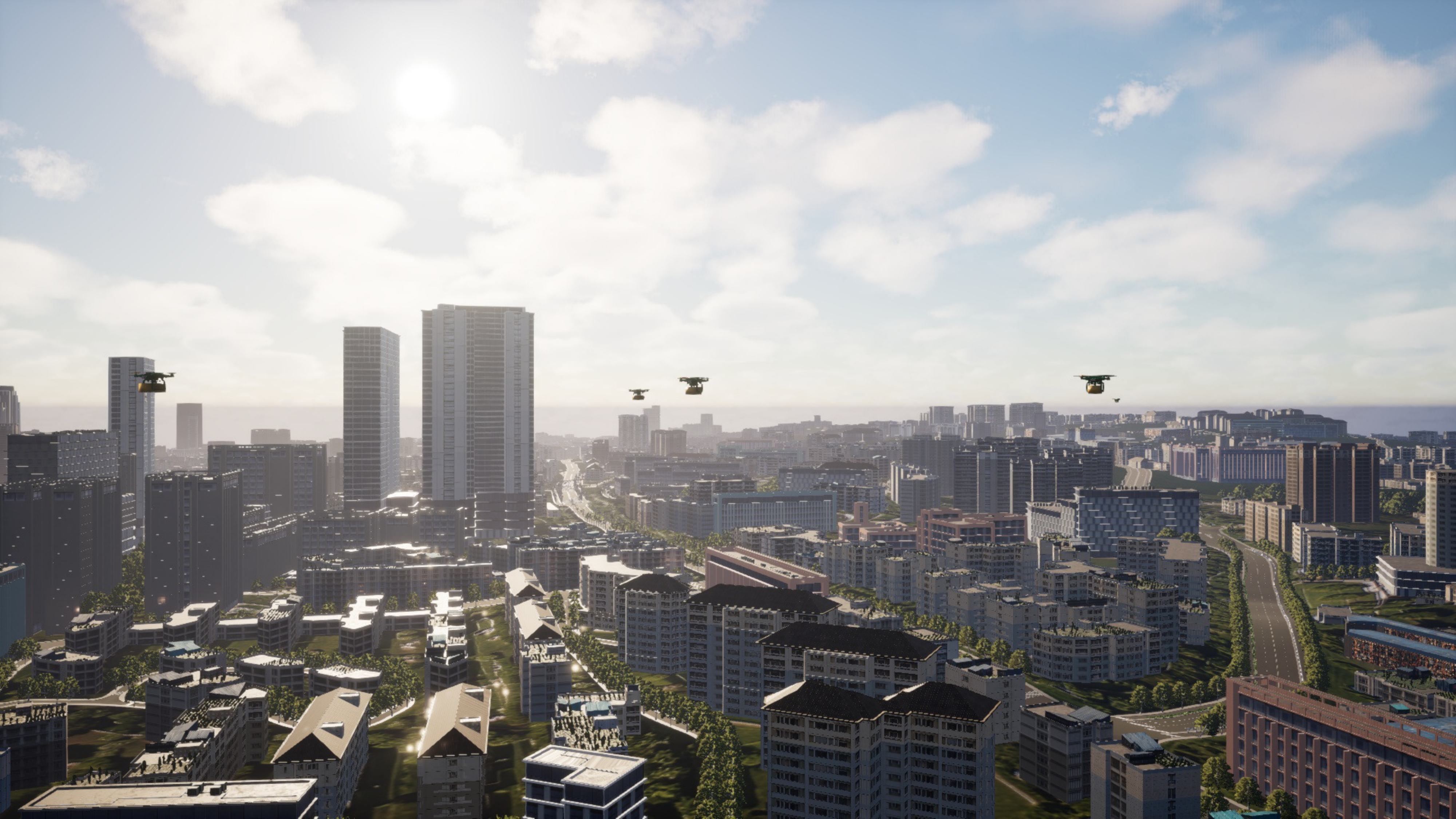} &
\includegraphics[width=0.12\textwidth]{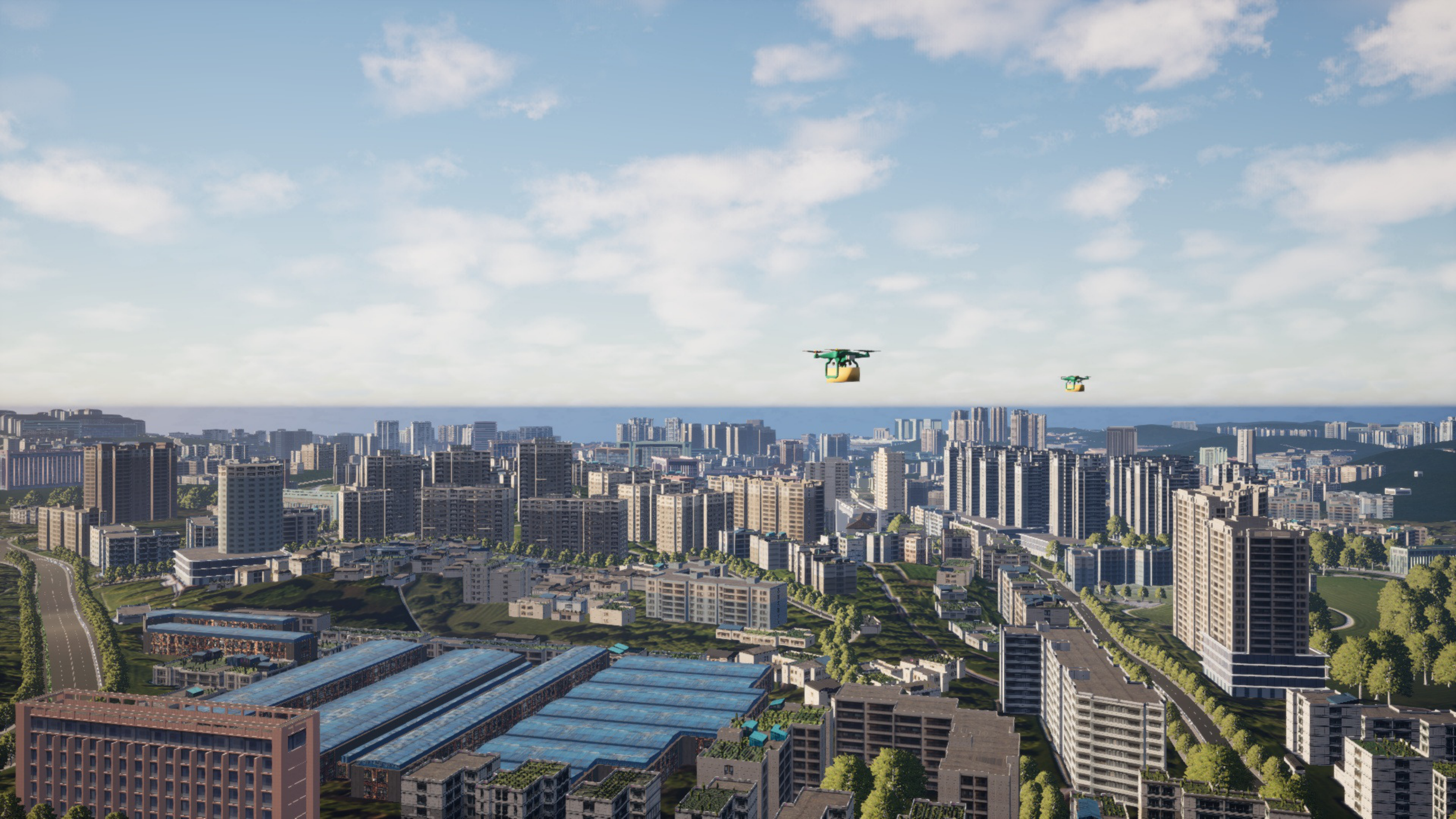} \\

\scriptsize RuralAustralia \cite{UE_RuralAustralia} &
\includegraphics[width=0.12\textwidth]{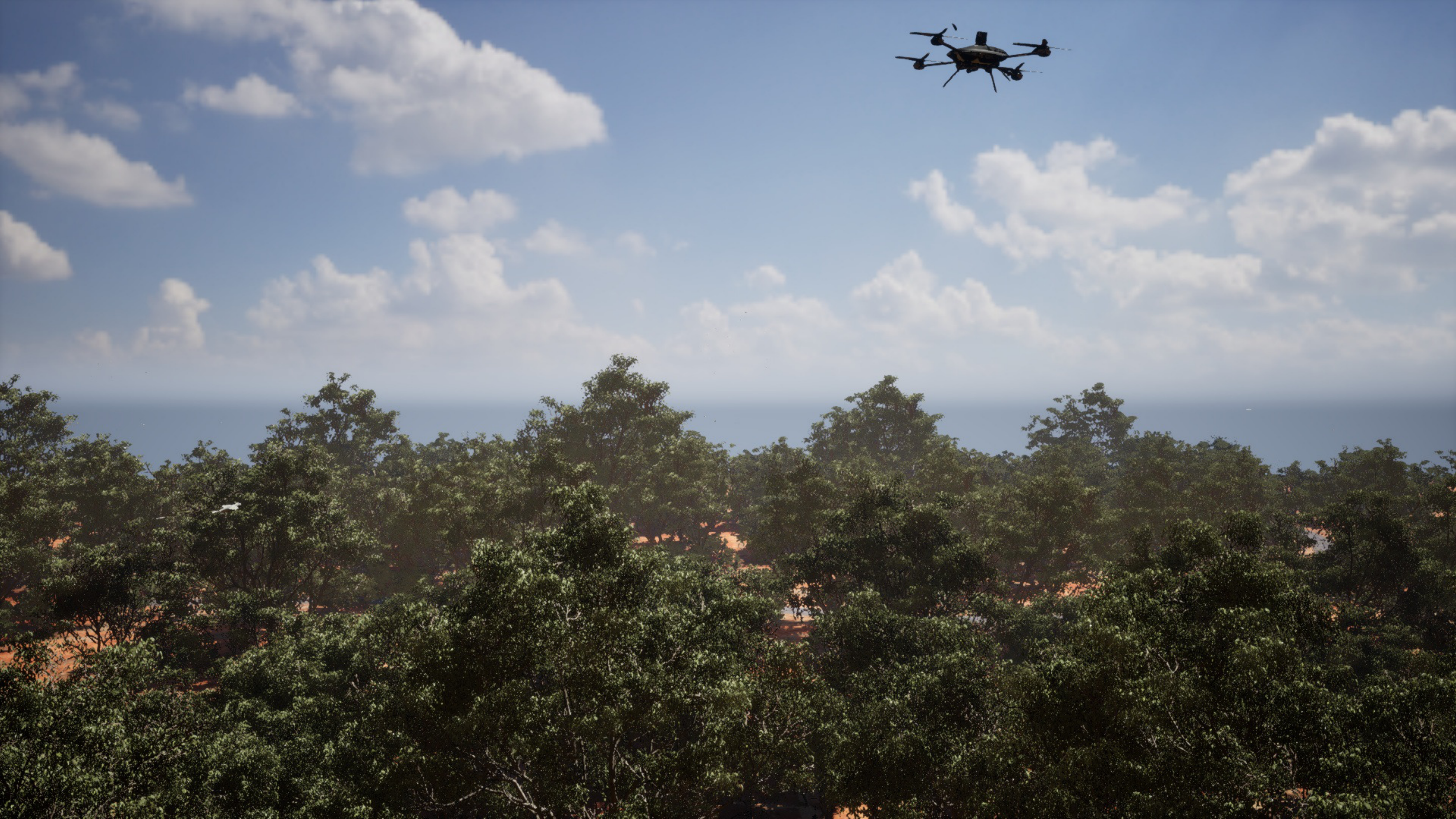} &
\includegraphics[width=0.12\textwidth]{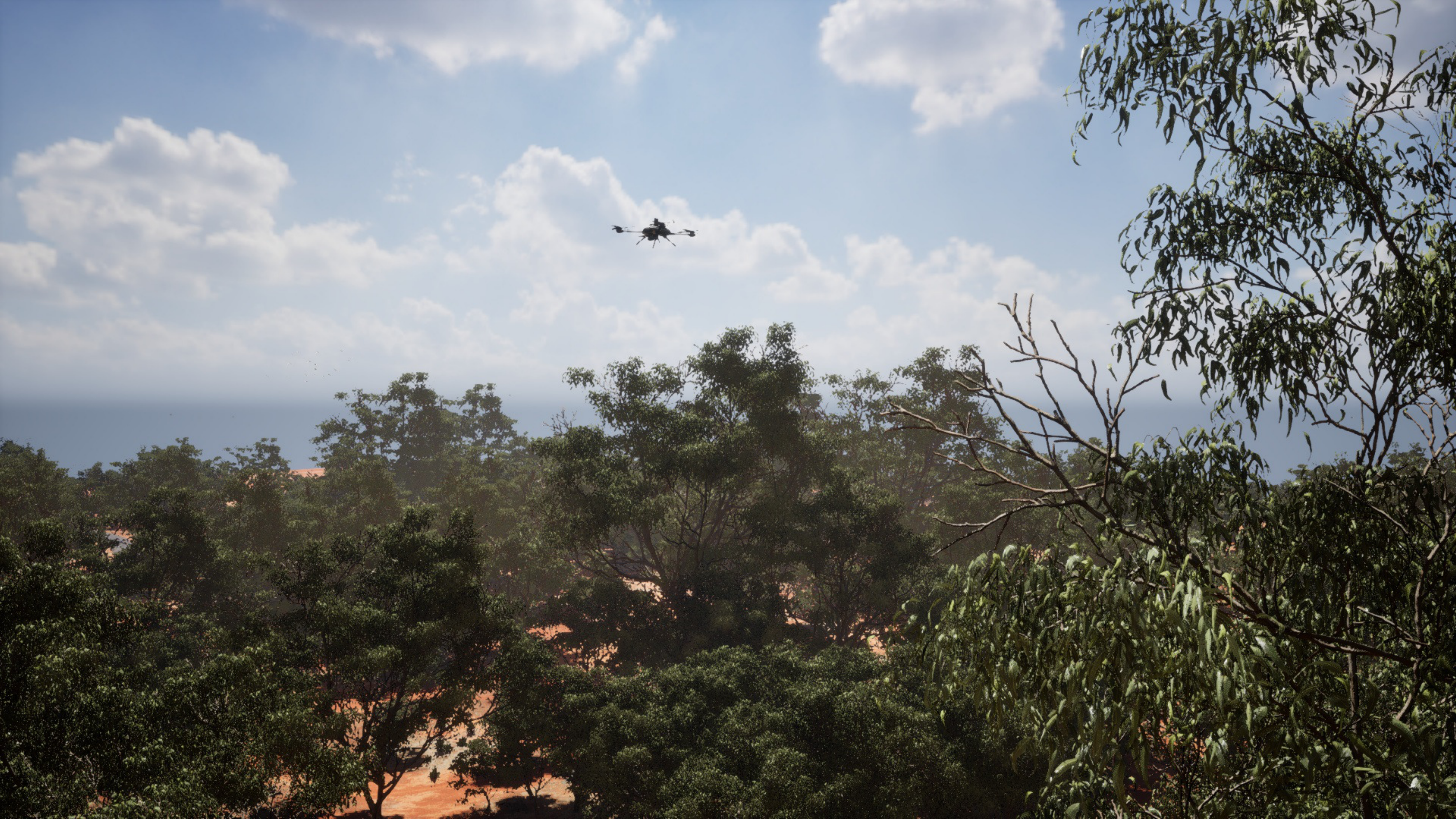} &
\includegraphics[width=0.12\textwidth]{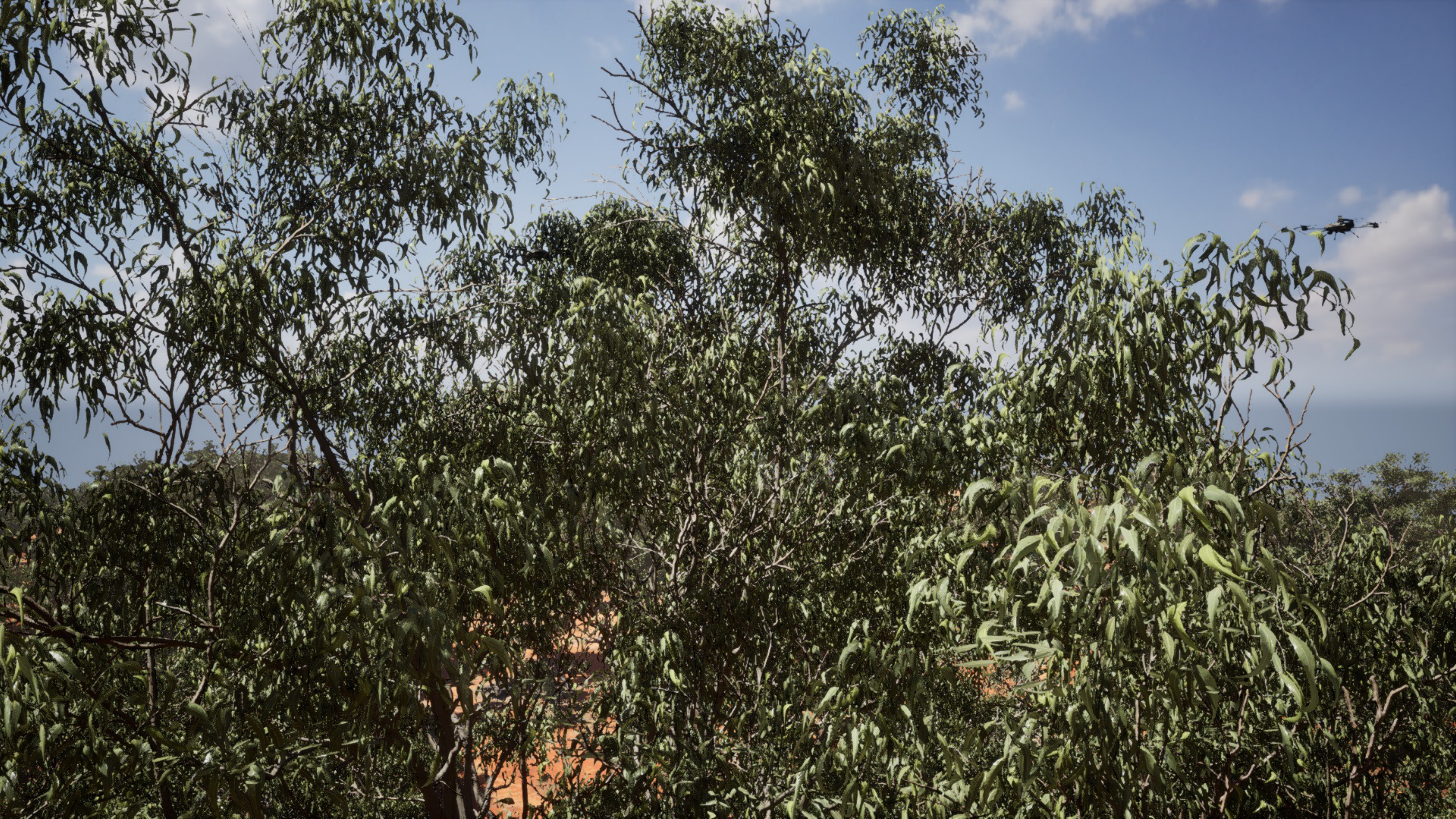} &
\includegraphics[width=0.12\textwidth]{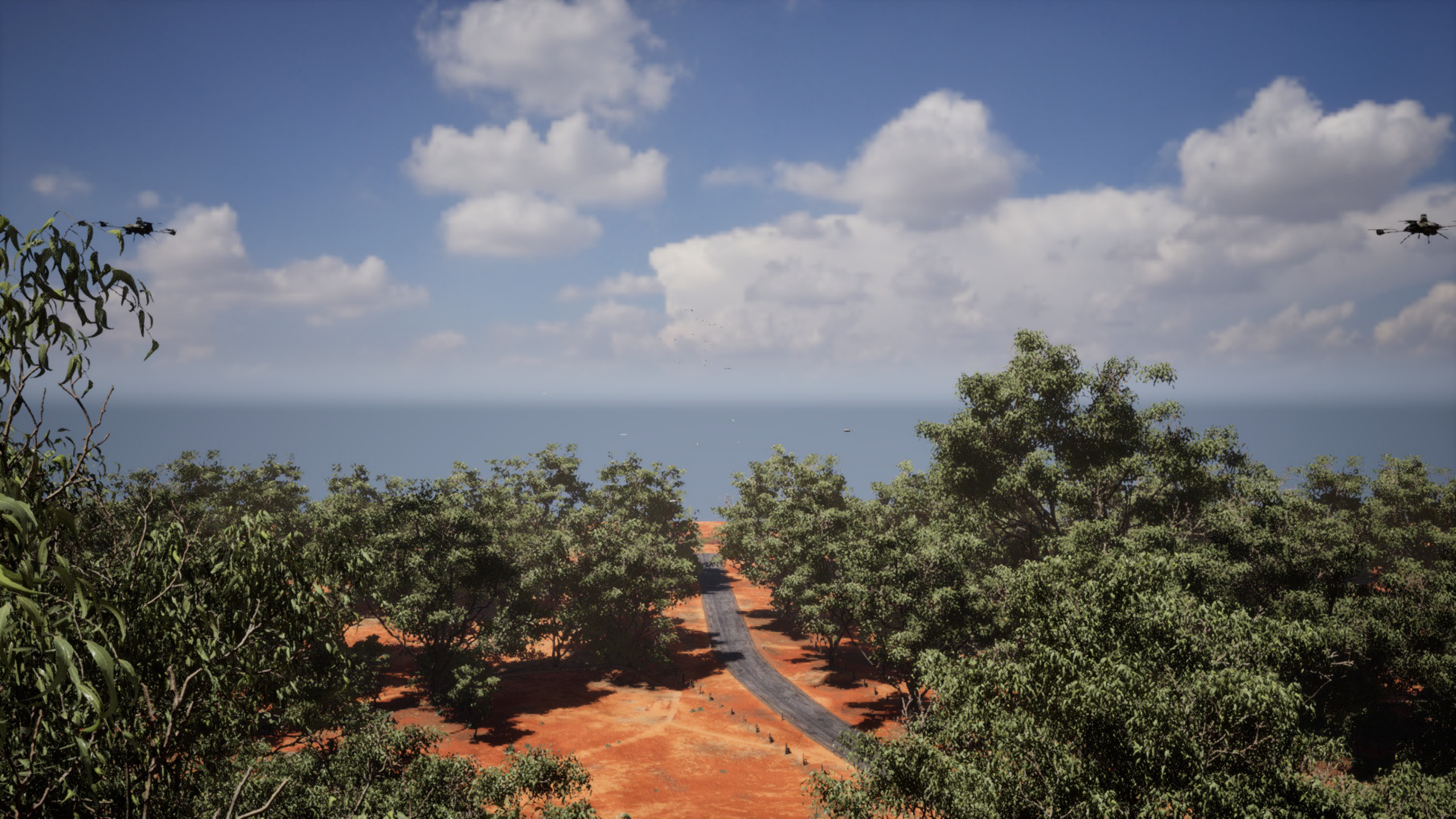} &
\includegraphics[width=0.12\textwidth]{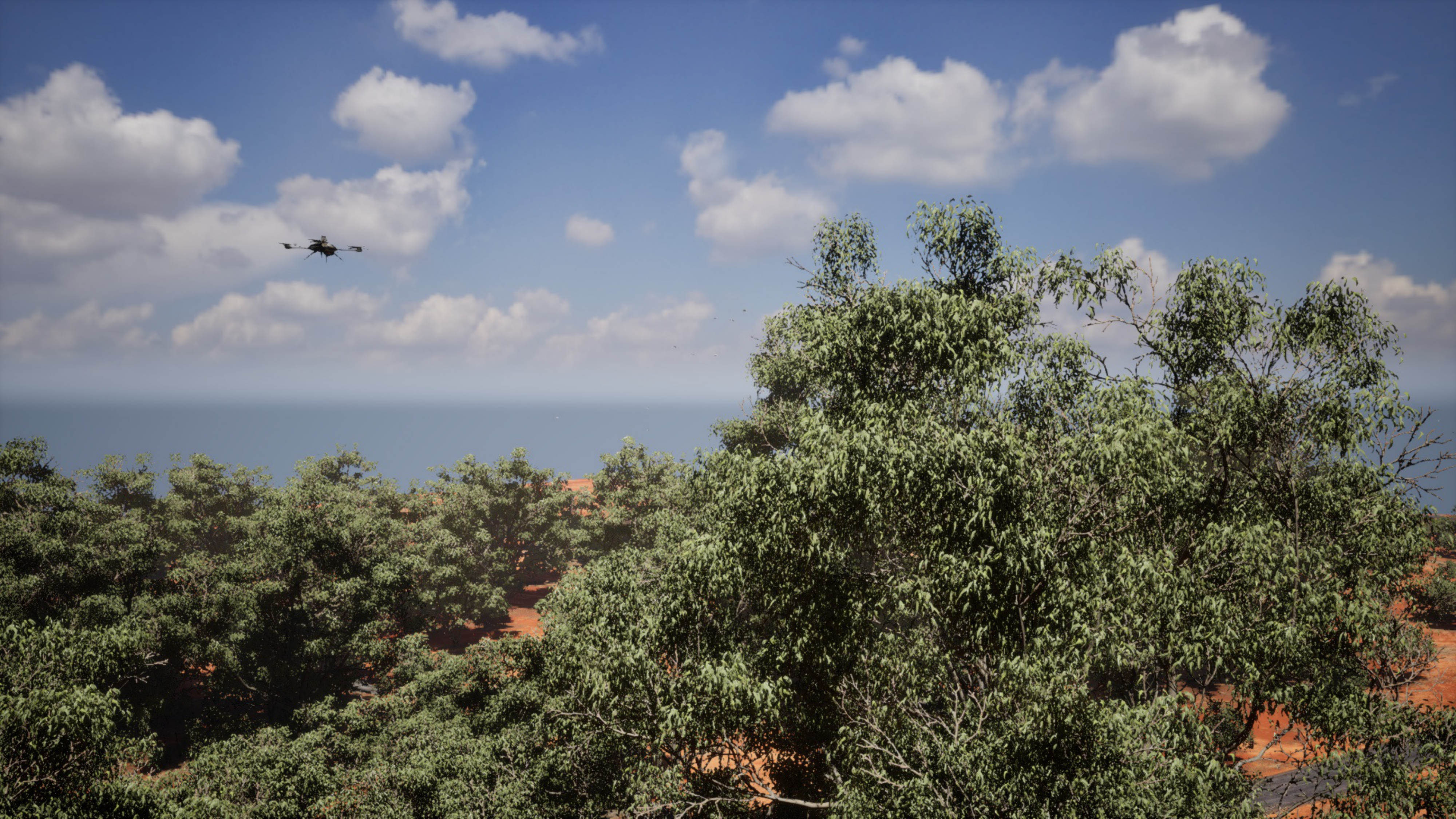} &
\includegraphics[width=0.12\textwidth]{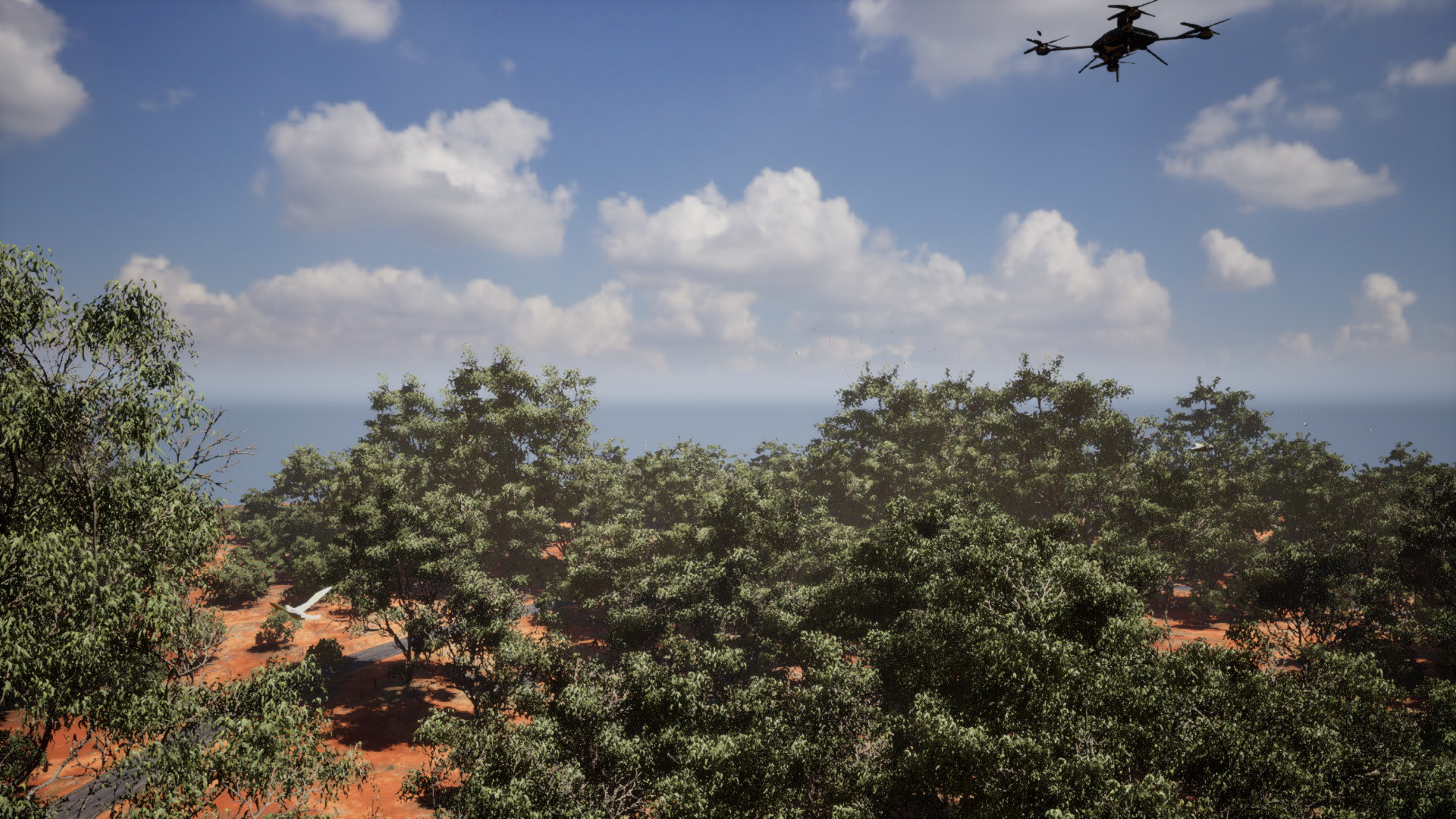} \\

\end{tabular}

\caption{Qualitative multi-camera visualization across diverse simulated environments. The sensing setup comprises six synchronized cameras (cam0--cam5), each with a horizontal field of view of 60°, uniformly arranged in a circular configuration to provide full 360° panoramic coverage of the scene. Columns correspond to camera viewpoints ordered from cam0 to cam5 in a clockwise manner. Each row represents a distinct environment: Row~1-Bridge, Row~2-CityCreator, Row~3-CityPark, Row~4-WWT, Row~5-DownTown, Row~6-DynamicCity, and Row~7-RuralAustralia. The scenes contain dynamic aerial entities, including birds and unmanned aerial vehicles (drones), whose appearance, scale, and visibility vary across camera viewpoints and environments.}
\label{fig:multicam_grid}
\end{figure*}

Within each environment, drone motion and camera placement are varied to capture a wide range of viewpoints and target appearances.
A 360$^\circ$ multi-camera configuration consists of a set of six synchronized virtual cameras cam0, cam1, cam2, cam3, cam4, cam5 each with identical intrinsic parameters and a horizontal
field-of-view of $60^\circ$ mounted on a rigid rig. The cameras are uniformly distributed in azimuth,
providing full $360^\circ$ panoramic coverage of the surrounding environment. Figure \ref{fig:multicam_grid} shows for $360^\circ$ view captured using 6 cameras of each scene.

This setup captures drones and bird objects entering and exiting the field of view from multiple directions and naturally produces challenging cases involving partial observations, strong perspective changes, and rapid motion across adjacent views.  Drones traverse the environments along predefined circular and spiral flight paths with varying radius, altitude, and speed, resulting in image samples that span a wide range of apparent target sizes, from small distant drones to close-up views.

To further enhance realism and domain diversity, environmental conditions are systematically randomized. These variations include changes in time of day, sun position and intensity, cloud coverage, and adverse weather effects such as fog, rain, and low-visibility conditions. Such variations are essential for evaluating detection performance under challenging illumination and weather scenarios commonly encountered in real-world deployments.

The combination of diverse virtual environments, systematic weather variation, structured flight trajectories, and synchronized multi-camera sensing results in a dense and diverse collection of training samples. This design supports the development and evaluation of detection models that are robust to scale variation, viewpoint changes, background complexity, and adverse environmental conditions.

\begin{figure*}[t]
\centering
\setlength{\tabcolsep}{2pt}
\renewcommand{\arraystretch}{1.0}

\begin{tabular}{lccccccc}
 & \scriptsize Clean & \scriptsize 2\% & \scriptsize 4\% & \scriptsize 6\% & \scriptsize 8\% & \scriptsize 10\% & \scriptsize 12\% \\ \hline

\scriptsize Fog &
\includegraphics[width=0.12\textwidth]{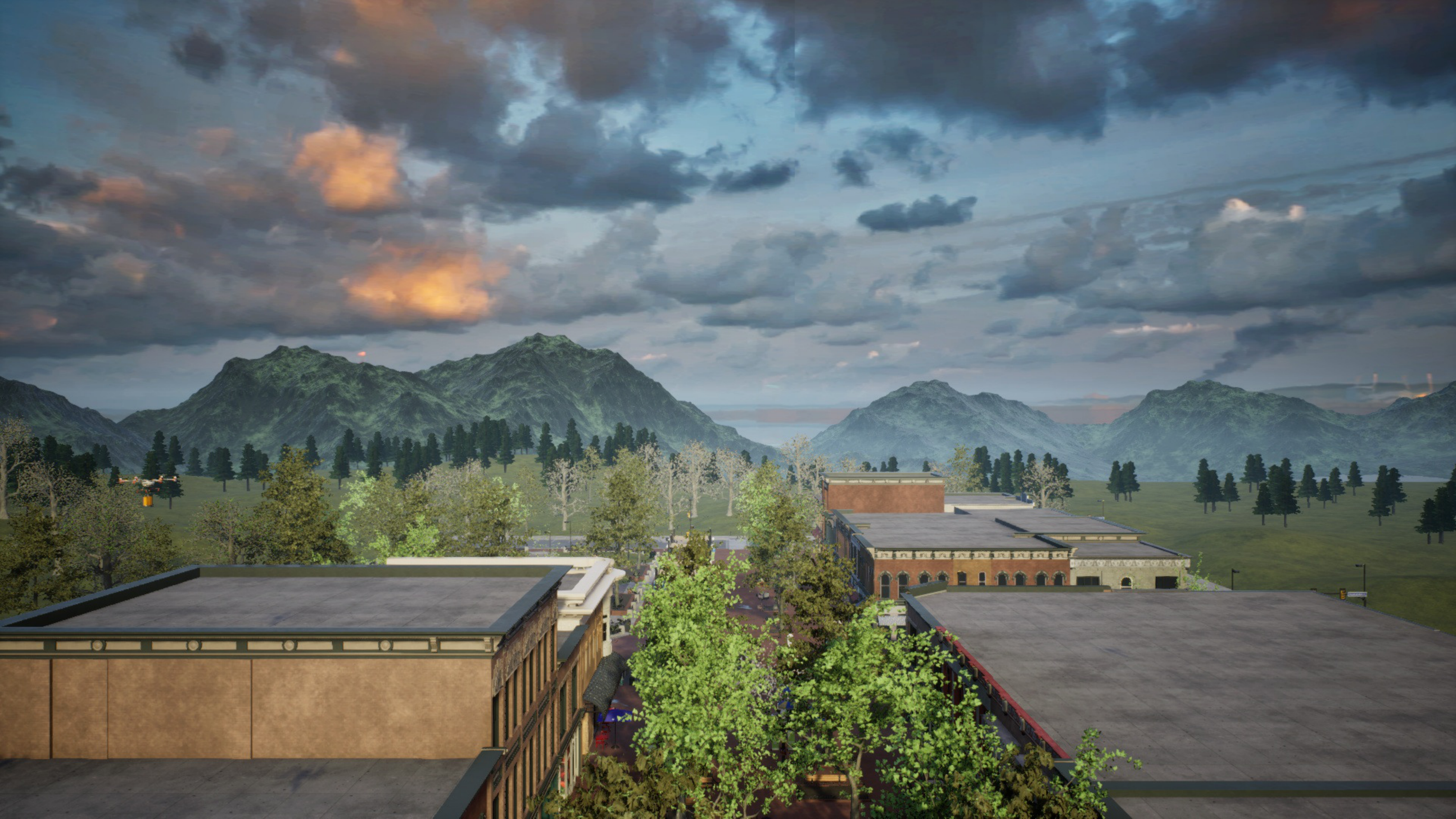} &
\includegraphics[width=0.12\textwidth]{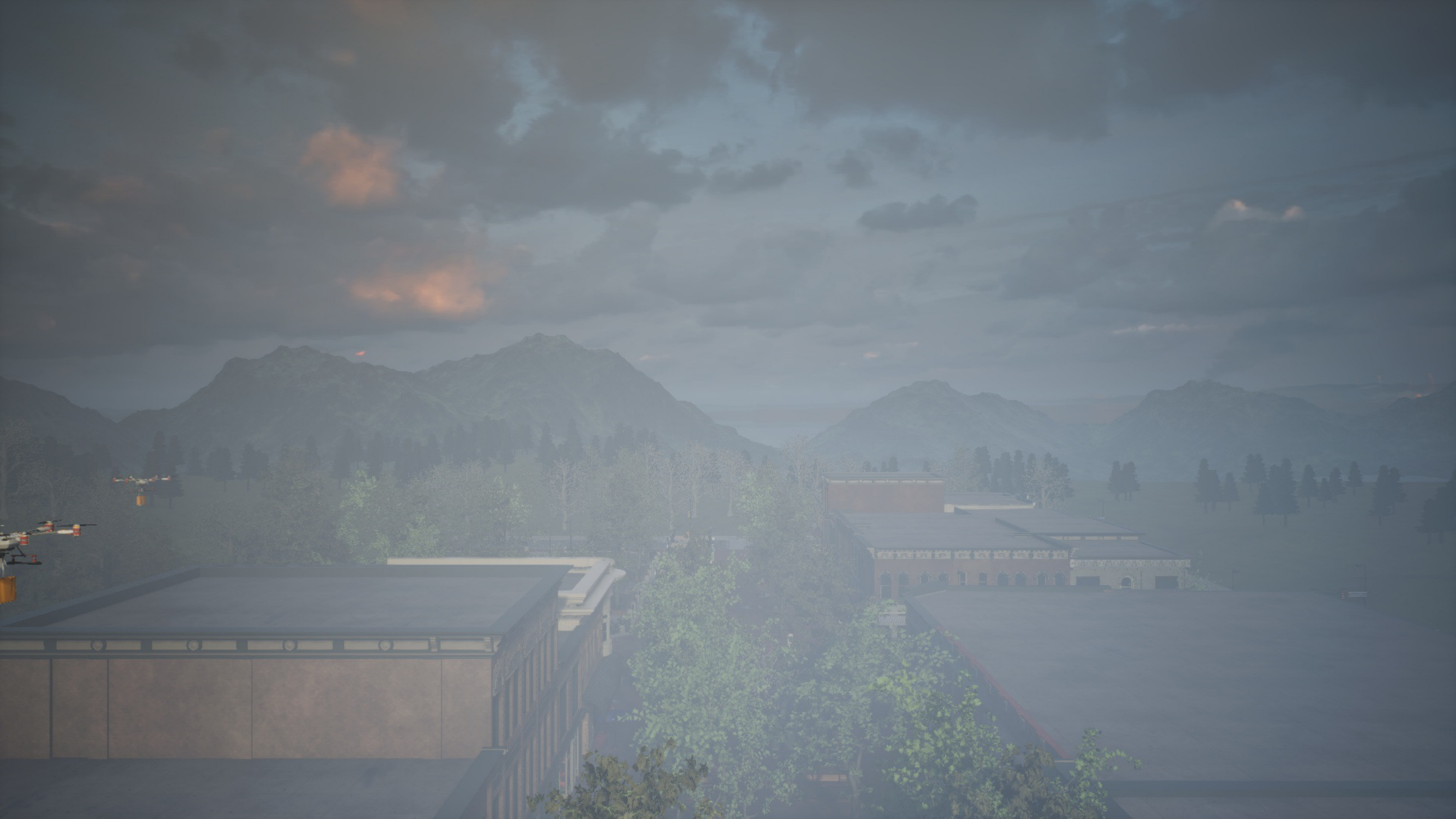} &
\includegraphics[width=0.12\textwidth]{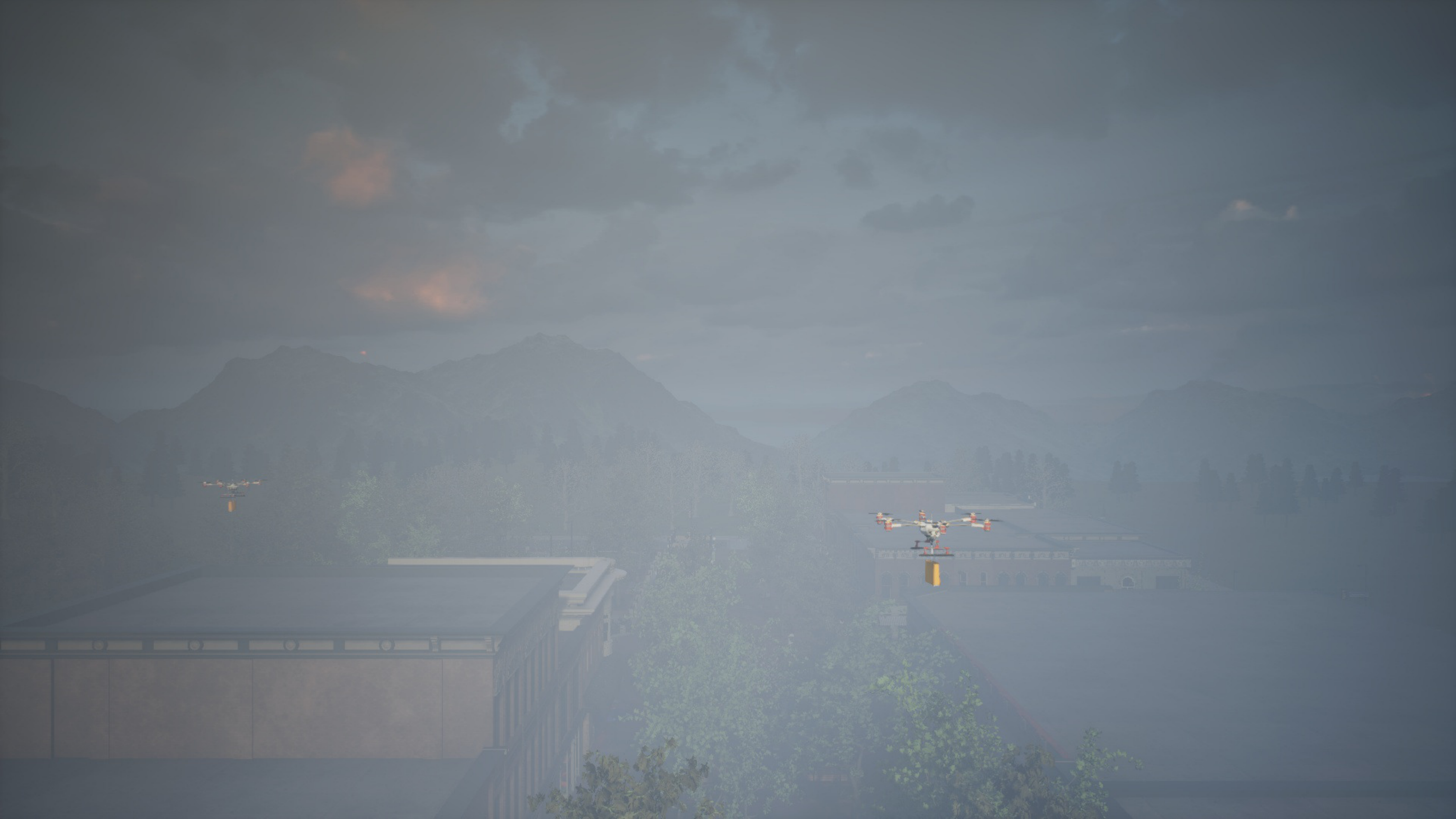} &
\includegraphics[width=0.12\textwidth]{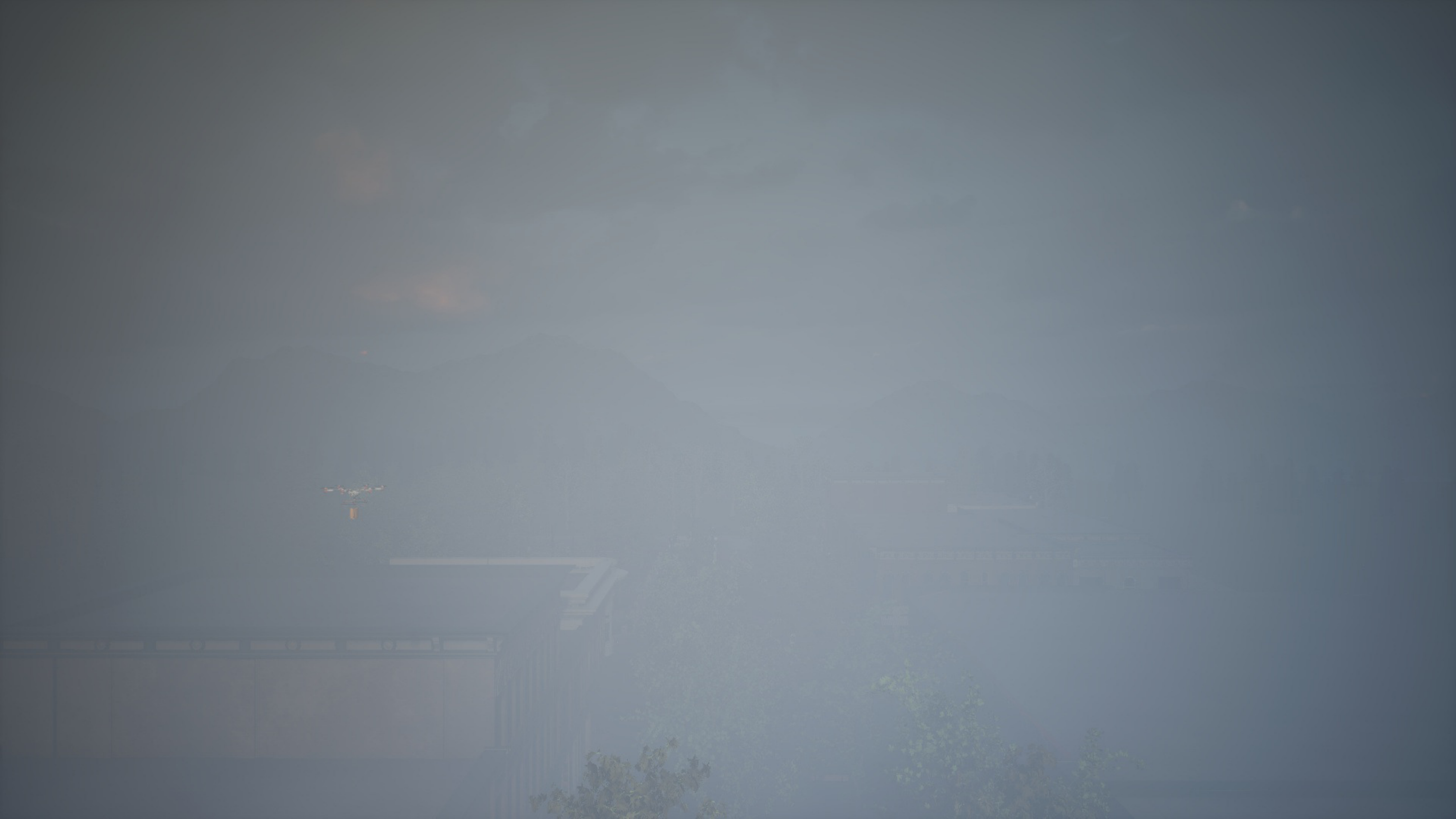} &
\includegraphics[width=0.12\textwidth]{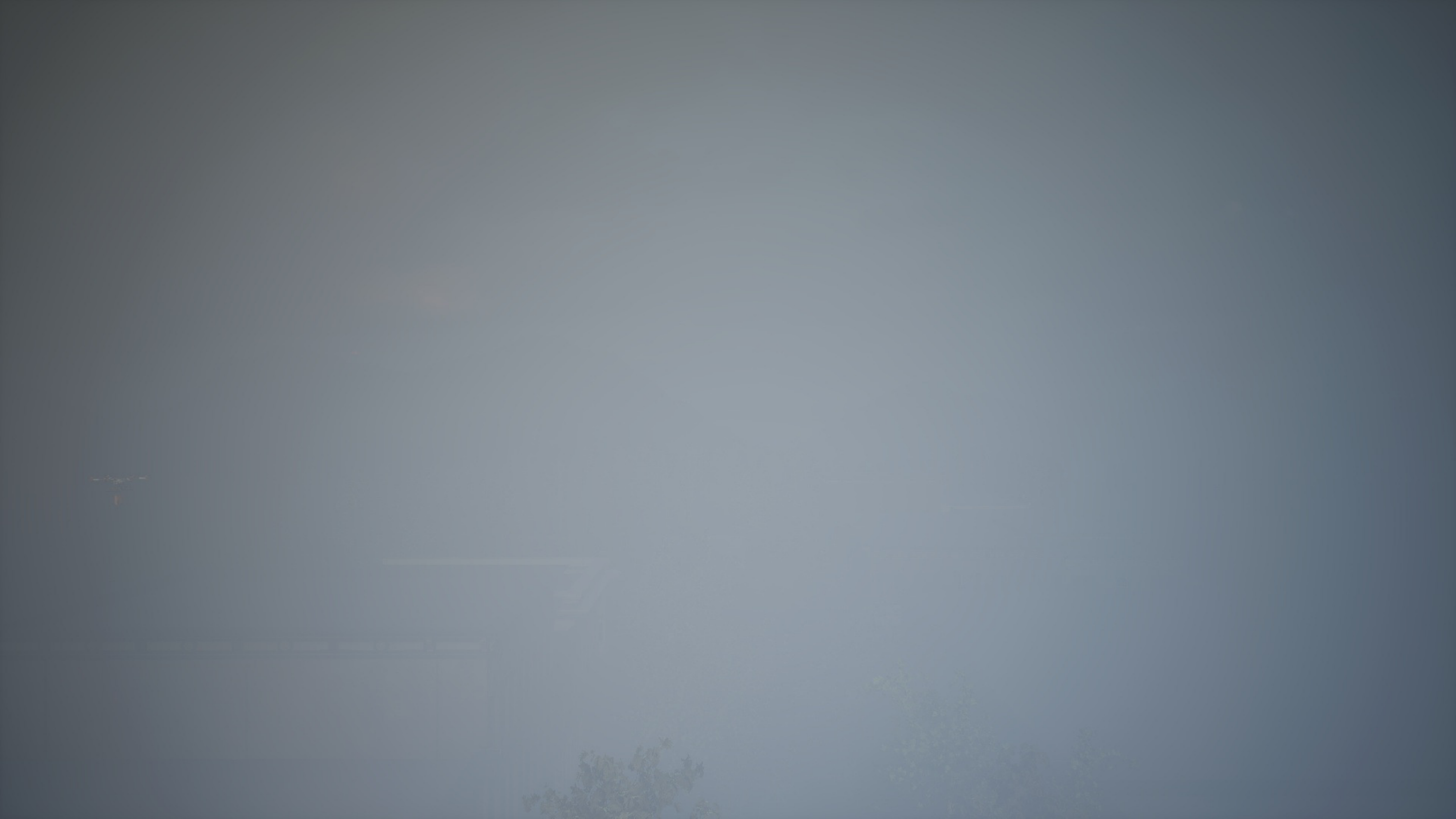} &
\includegraphics[width=0.12\textwidth]{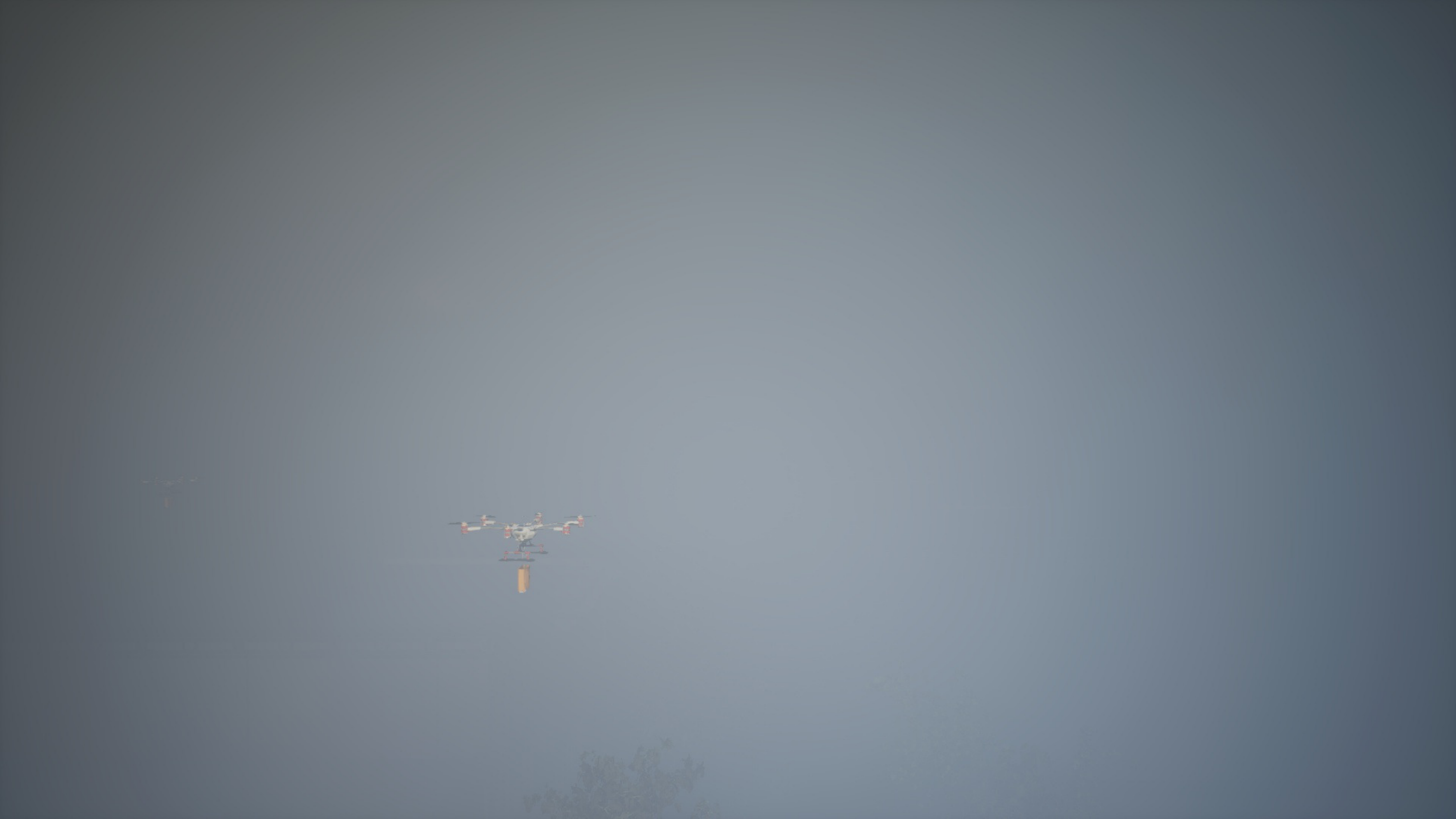} &
\includegraphics[width=0.12\textwidth]{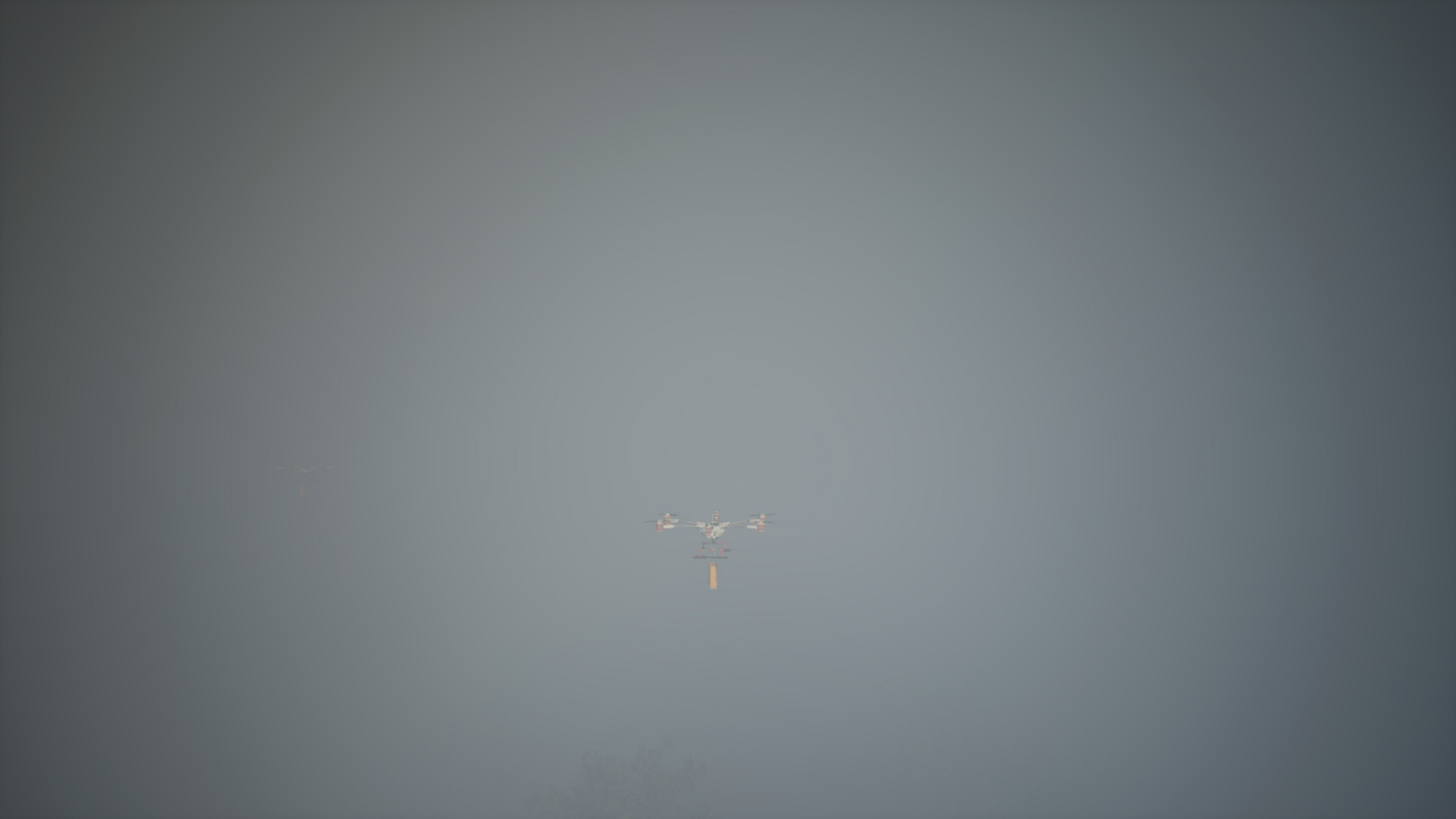} \\

\end{tabular}

\caption{Qualitative visualization of observations under varying environmental conditions. Each column corresponds to increasing severity levels of the environmental condition, ranging from clean observations to progressively stronger corruptions (2\%--12\%). 
The first row illustrates the Fog environment, where visibility degradation increases monotonically from left to right. The scenes contain dynamic aerial objects, including birds and unmanned aerial vehicles (drones).}
\label{fig:fog_grid}
\end{figure*}

\begin{figure*}[t]
\centering
\setlength{\tabcolsep}{2pt}
\renewcommand{\arraystretch}{1.0}

\begin{tabular}{lccccccc}
 & \scriptsize Clean & \scriptsize 5\% & \scriptsize 15\% & \scriptsize 25\% & \scriptsize 35\% & \scriptsize 45\% & \scriptsize 55\% \\ \hline
\scriptsize Snow &
\includegraphics[width=0.12\textwidth]{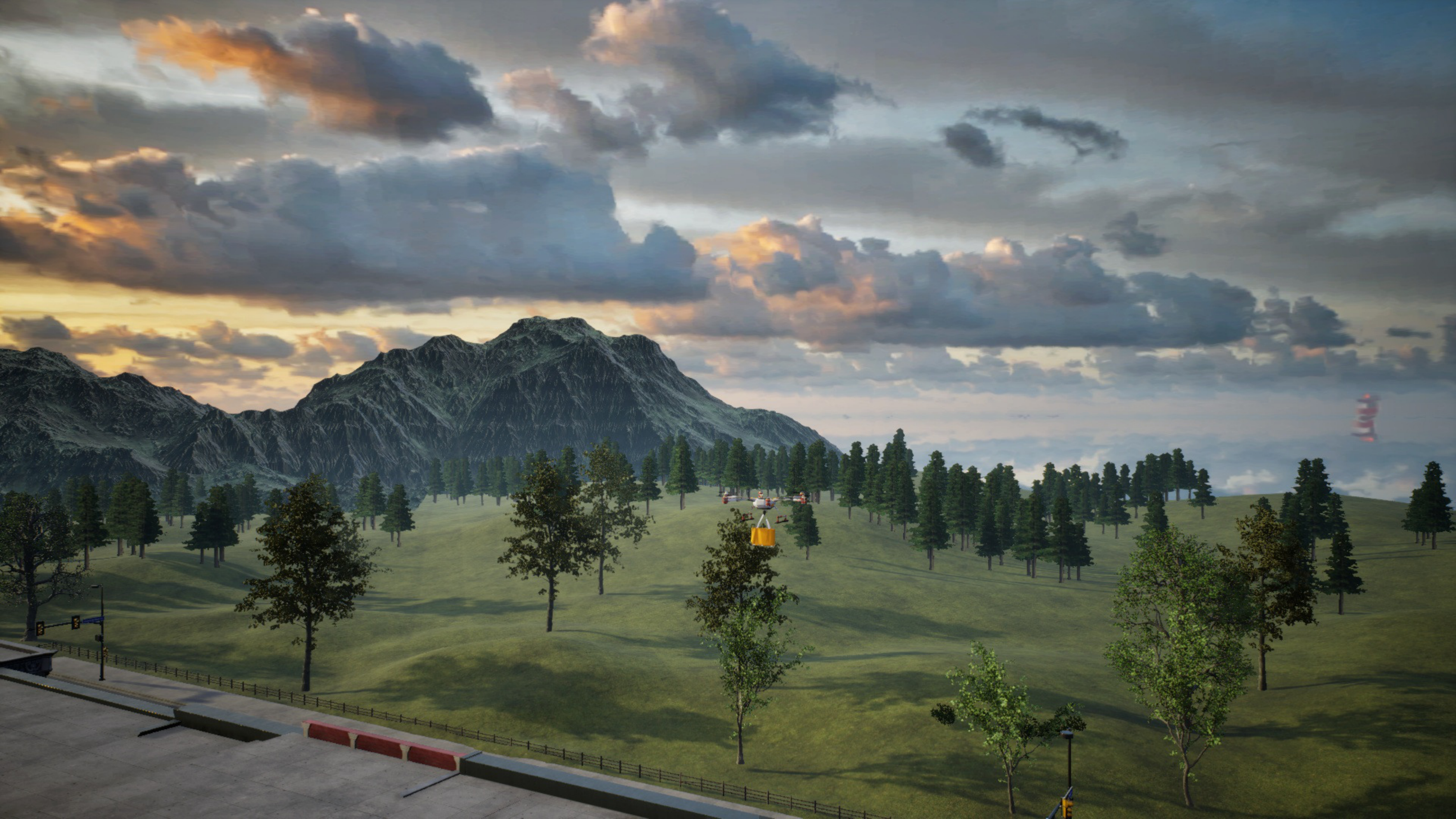} &
\includegraphics[width=0.12\textwidth]{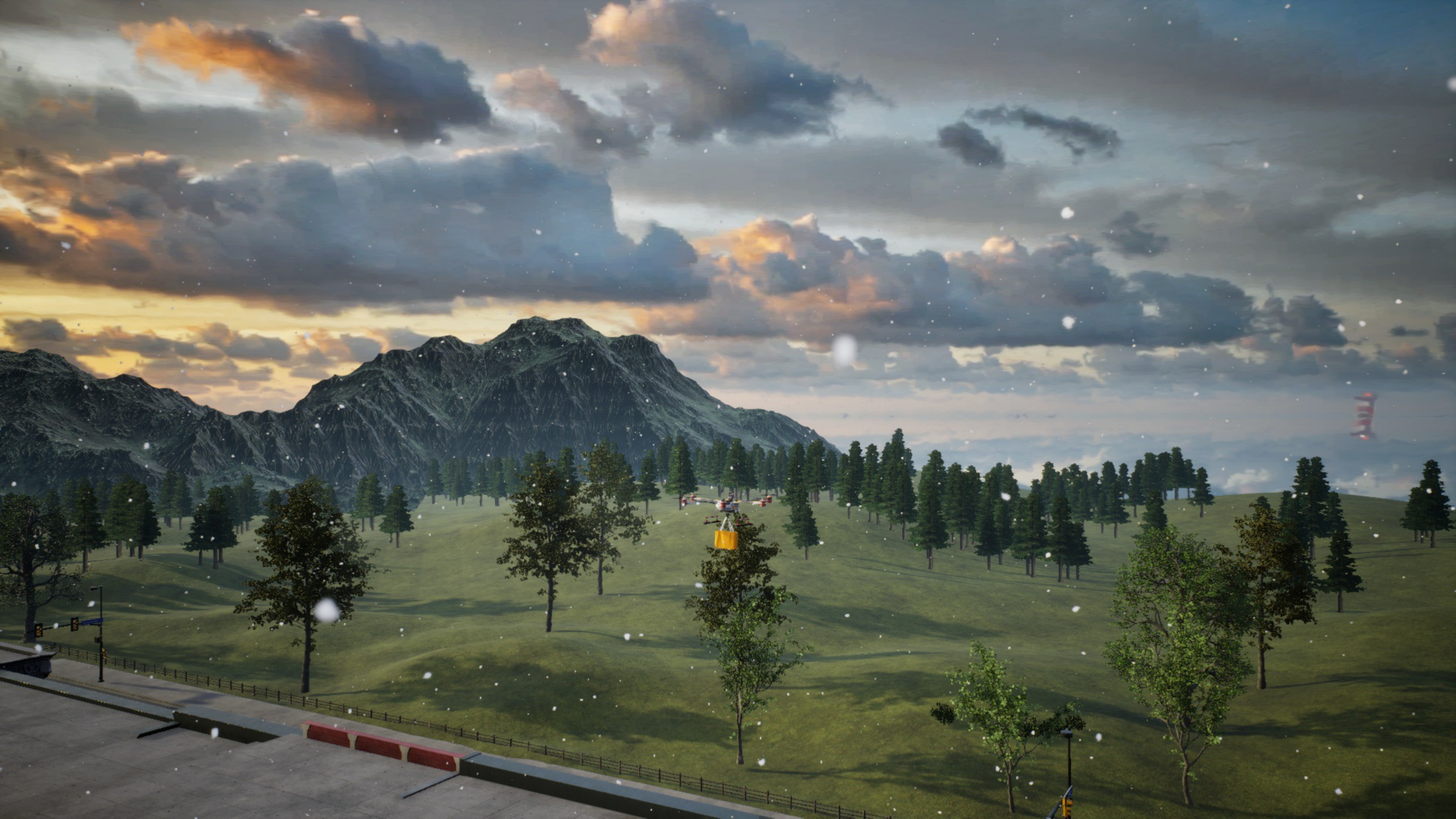} &
\includegraphics[width=0.12\textwidth]{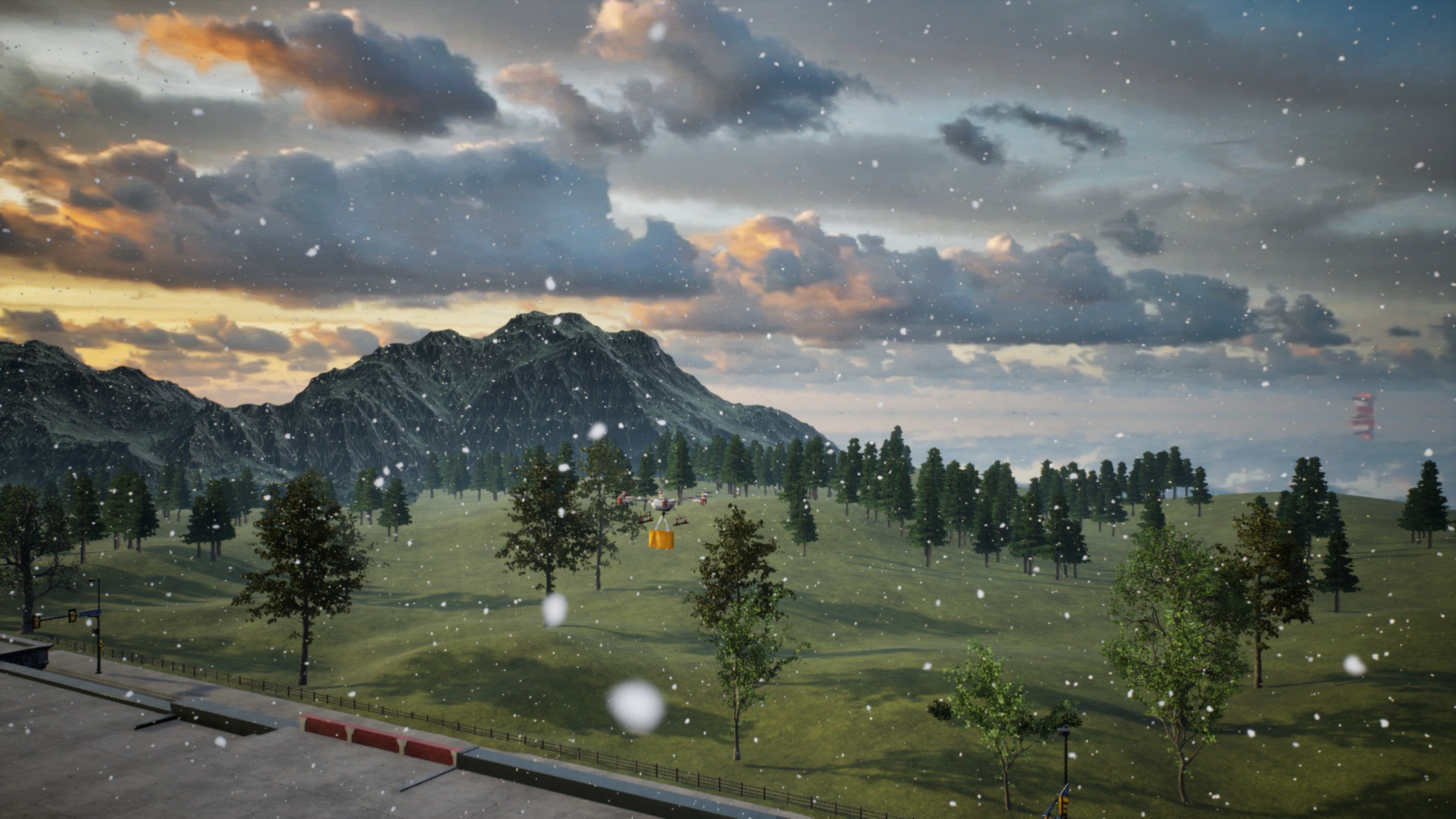} &
\includegraphics[width=0.12\textwidth]{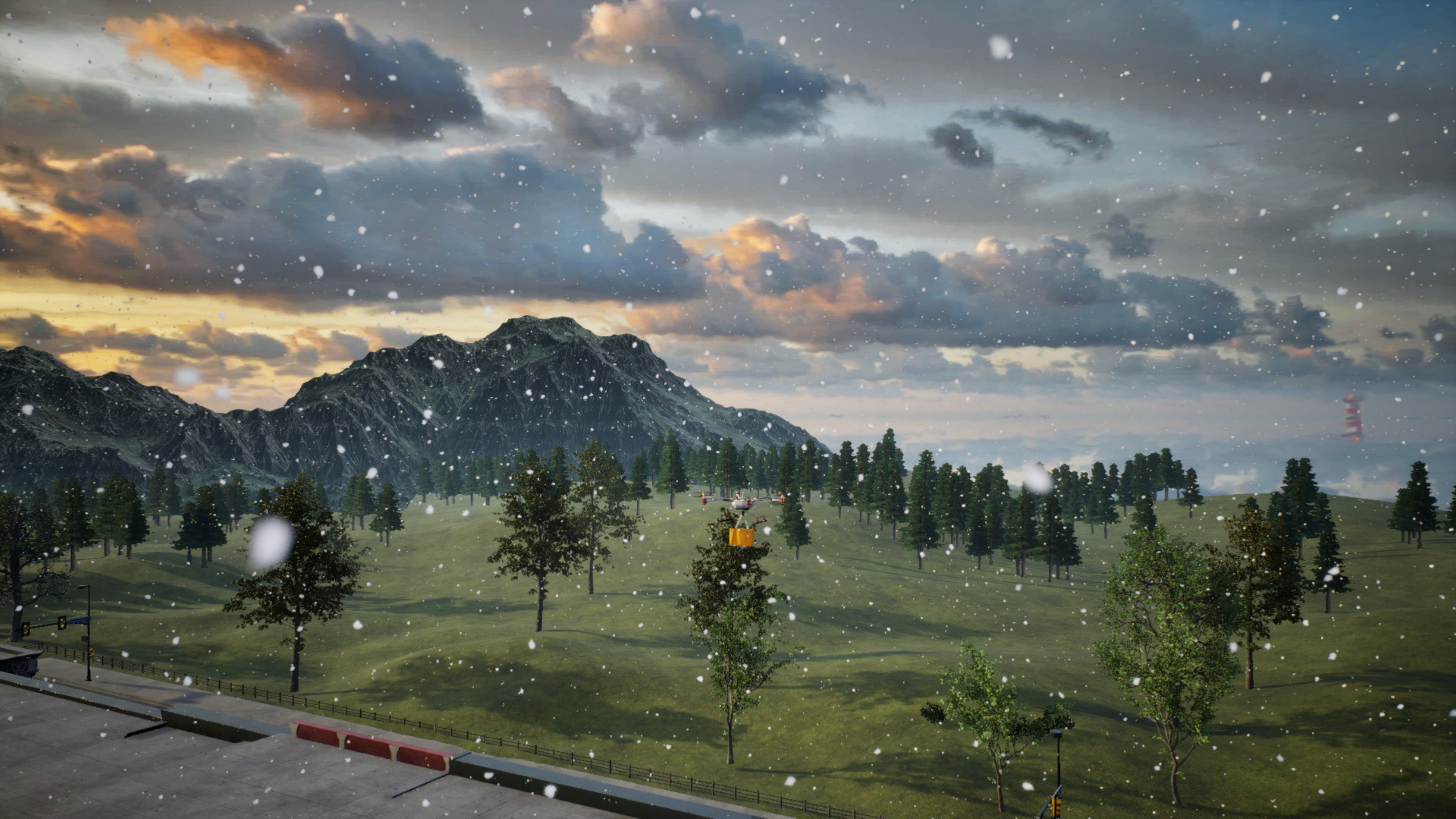} &
\includegraphics[width=0.12\textwidth]{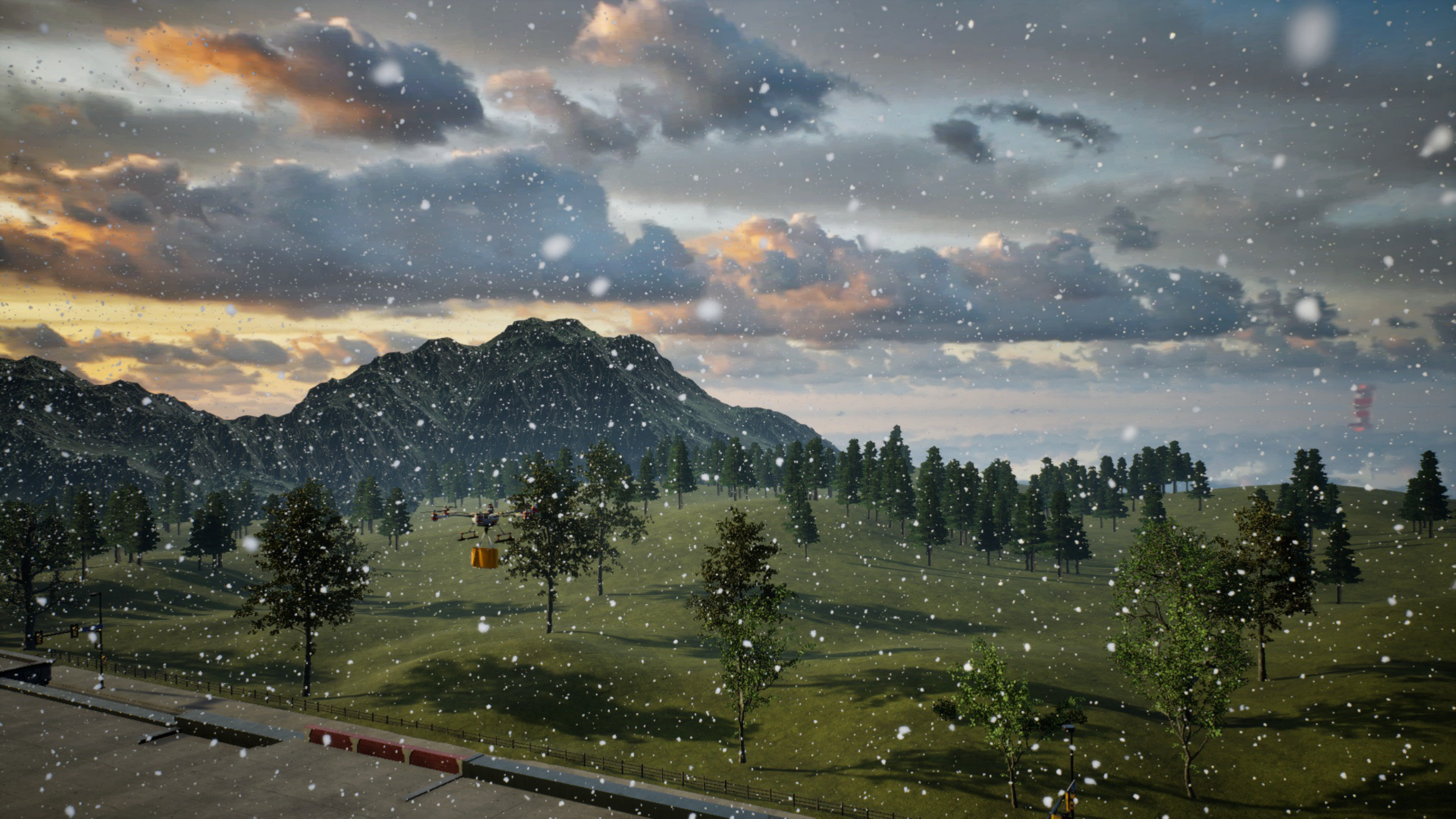} &
\includegraphics[width=0.12\textwidth]{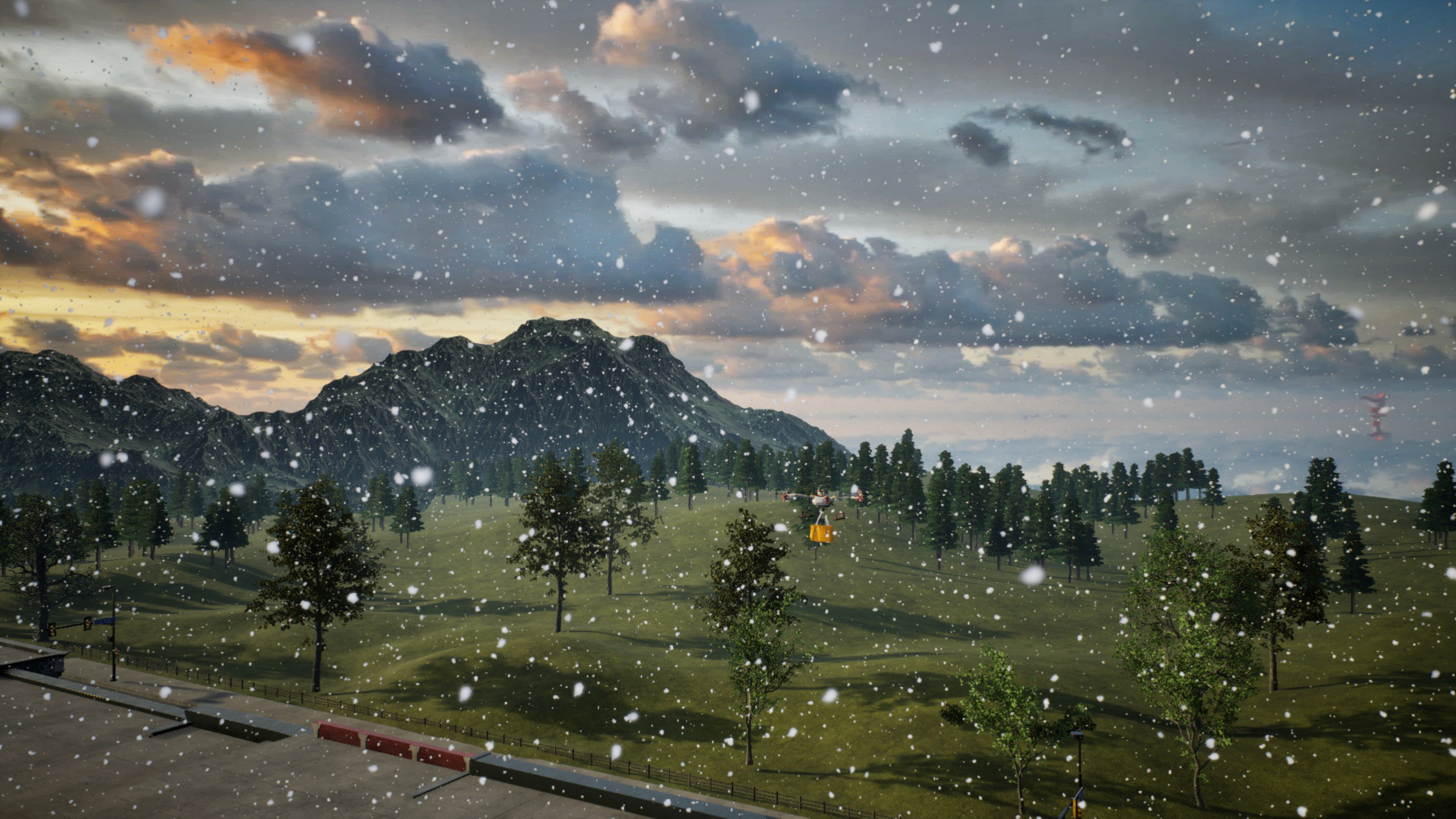} &
\includegraphics[width=0.12\textwidth]{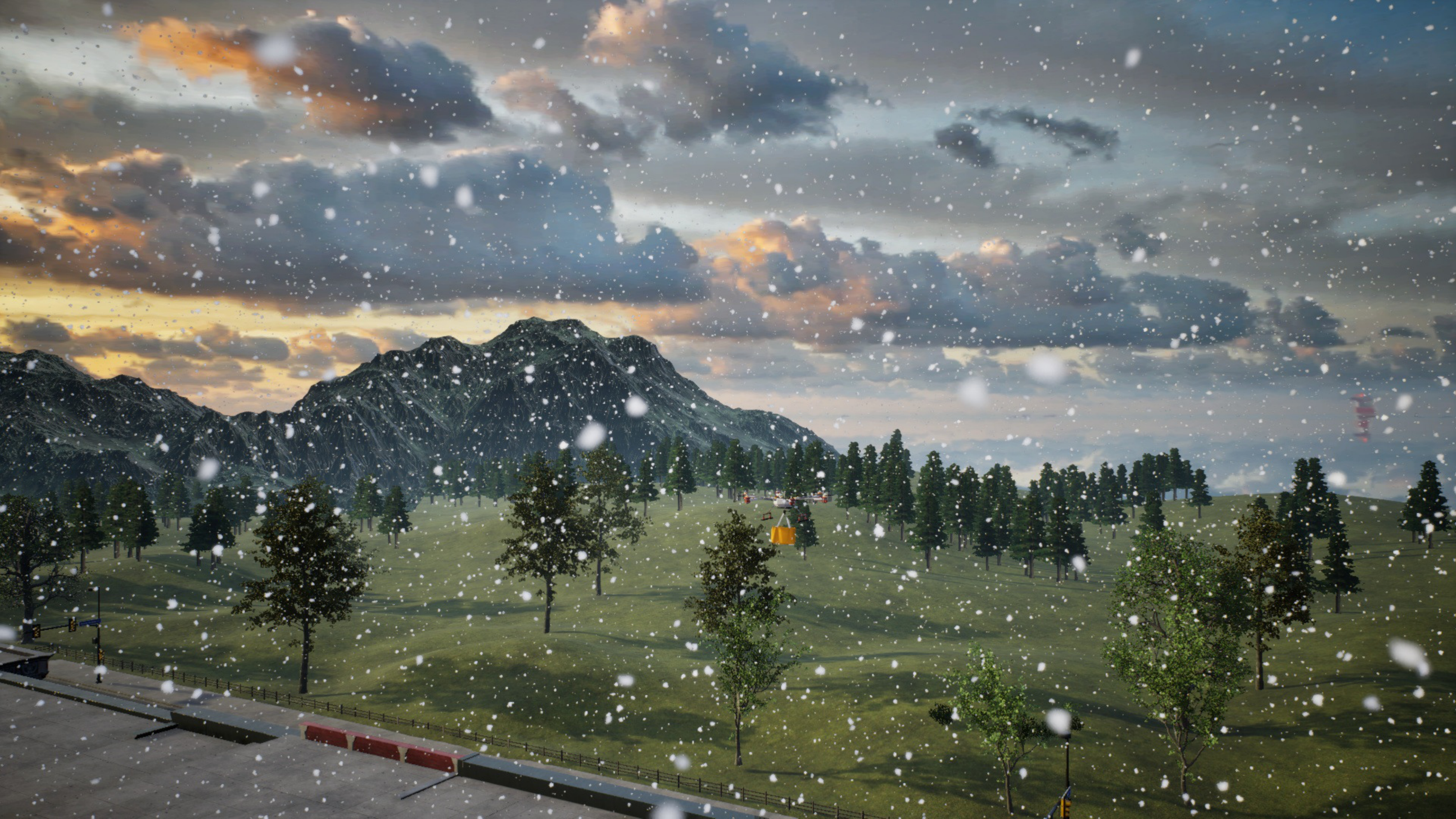}\\

\end{tabular}

\caption{Qualitative visualization of observations under varying environmental conditions. Each column corresponds to increasing severity levels of the environmental condition, ranging from clean observations to progressively stronger corruptions (5\%--55\%). 
The first row illustrates the Snow environment, where visibility degradation increases monotonically from left to right. The scenes contain dynamic aerial objects, including birds and unmanned aerial vehicles (drones).}
\label{fig:snow_grid}
\end{figure*}

\subsection{Adverse Weather Conditions}
\label{sec:adverse_weather}

Most publicly available vision-based drone datasets are dominated by clear-weather imagery with stable illumination and high visibility. In contrast, real-world surveillance systems must operate reliably under adverse conditions such as fog, snow, rain, and low-light environments, where reduced contrast, atmospheric scattering, and visual artifacts significantly degrade image quality. Prior studies have shown that drone detection performance can drop sharply under such conditions when models are trained primarily on clear-weather data \cite{Munir2024UAVAWID,Tahir2024AdverseWeatherOD}. These findings highlight the importance of incorporating adverse weather scenarios directly into dataset design rather than treating them as post hoc augmentation.

To address this challenge, SimD3 includes a dedicated adverse-weather component generated using the weather simulation API provided by Microsoft AirSim (CoSys version) integrated with Unreal Engine~5. AirSim enables physically consistent weather rendering through parameterized controls that allow continuous adjustment of weather intensity. In particular, the API exposes scalar intensity parameters for atmospheric effects such as fog and snow, making it possible to systematically vary weather severity from mild to extreme conditions within the same scene.

Using this capability, we generate weather specific subsets of SimD3 focusing on fog and snow conditions, which are among the most challenging for vision-based drone detection due to visibility loss and texture suppression. For each virtual environment described in previous subsection, weather intensity levels are varied to produce a spectrum of conditions, ranging from light haze or snowfall to dense fog and heavy snow (Figure \ref{fig:fog_grid} and Figure \ref{fig:snow_grid}). These weather effects are applied consistently across environments and combined with the same drone configurations, payload variations, bird distractors, flight trajectories, and multi-camera setup used in clear-weather scenes. As a result, identical scene layouts and object arrangements are observed under different atmospheric conditions, enabling controlled comparisons across weather states.

All weather effects are rendered directly within the simulation engine, ensuring pixel-accurate annotations and consistent ground-truth labels regardless of visibility. Each adverse-weather image contains at least one drone instance, with or without payload, and may include zero or more bird objects acting as structured distractors. This design supports both robustness evaluation and all-weather training, allowing researchers to analyze performance degradation, conduct ablation studies across weather severity, and develop models specifically tailored to handle weather-induced visual degradation.


\subsection{Annotation Pipeline}

All annotations in SimD3 are automatically derived using the segmentation API provided by AirSim. For each rendered frame, pixel-level semantic segmentation maps are generated, with unique labels assigned to each object category. 

Instance-level annotations are obtained by extracting connected components from the semantic segmentation maps, with each connected region treated as an individual object instance.

The dataset includes two primary object classes:
\begin{itemize}
    \item \texttt{drone}: all UAV models, including both payload-equipped and payload-free variants
    \item \texttt{bird}: bird models used as aerial distractors
\end{itemize}

Bounding boxes are tightly fitted to the visible extent of each object, ensuring accurate localization even for small and partially occluded targets.

All annotations are exported in the standard YOLOv5 \cite{yolov5} format, where each object is represented by its class label and normalized bounding box coordinates. This choice ensures immediate compatibility with widely used object detection frameworks and allows the dataset to be readily adopted for training and evaluation without additional preprocessing.
In YOLO format, each object instance is encoded as
\begin{equation}
y = (c,\; x_c,\; y_c,\; w,\; h),
\end{equation}
where $c$ is the class index and $(x_c, y_c)$ denotes the normalized bounding-box center coordinates.
The width $w$ and height $h$ are also normalized by the image dimensions, i.e.,
$x_c = \frac{x_{\min}+x_{\max}}{2W}$,
$y_c = \frac{y_{\min}+y_{\max}}{2H}$,
$w = \frac{x_{\max}-x_{\min}}{W}$, and
$h = \frac{y_{\max}-y_{\min}}{H}$,
with $W$ and $H$ denoting image width and height, respectively.

\subsection{Dataset Statistics}

SimD3 comprises a total of 178{,}639 annotated images spanning three subsets: Non-VFX, VFX, and adverse-weather conditions, as summarized in Table~\ref{tab:simd3_total_counts}. The dataset is structured to include frames containing only drones, only birds, and scenes where drones and birds co-occur, enabling systematic evaluation under varying levels of visual ambiguity and background complexity.

In the \emph{Non-VFX} subset, both drone and bird instances are explicitly annotated. This subset forms the primary source of supervised data for learning drone vs. bird discrimination and includes a balanced distribution across drone only, bird only, and combined drone-bird scenes. In contrast, the \emph{VFX} subset focuses exclusively on drone annotation. In this setting, bird-like motion is generated using Unreal Engine Niagara visual effects, where birds appear as dynamic background clutter but are intentionally not annotated. This design allows the dataset to support scenarios in which birds are treated as part of the background rather than explicit object classes, closely reflecting real-world surveillance conditions where such distinctions are not always available.

Table~\ref{tab:simd3_scene_counts} presents the scene-wise distribution of images across Unreal Engine Marketplace environments. Urban and semi-urban scenes such as \emph{Downtown}, \emph{CityPark}, and \emph{DynamicCity} contribute a substantial portion of the dataset, providing complex structural backgrounds with buildings, roads, and clutter. Additional environments, including \emph{Bridge} and \emph{Rural Australia}, introduce terrain-dominated and open-area contexts, ensuring that the dataset is not biased toward a single scene type and supports robust generalization across heterogeneous environments.

The adverse-weather subset is detailed in Table~\ref{tab:simd3_weather_breakdown}. Fog and snow conditions constitute the majority of this subset, with additional sequences generated under mixed or auxiliary weather configurations to preserve sequence consistency. Each weather condition includes images with drones only, birds only, and combined drone--bird scenarios, enabling controlled analysis of detection performance under degraded visibility and atmospheric effects.

All images in SimD3 are rendered at a fixed resolution of $1920 \times 1080$ pixels. The dataset explicitly targets small-object detection scenarios: the minimum visible extent of annotated objects is approximately 5 pixels, while the largest objects occupy at most 20\% of the image area. This range reflects realistic long-range surveillance conditions, where UAVs often appear as small, low-contrast targets, while still including closer views to support multi-scale feature learning. Overall, these statistics highlight SimD3’s emphasis on scale variability, scene diversity, and challenging detection conditions relevant to real-world UAV monitoring.

\begin{table}[t]
\centering
\caption{SimD3 subset statistics in terms of image-level composition. ``Drone'' and ``Bird'' denote frames containing only that category, while ``Both'' denotes frames containing drones and birds simultaneously. In VFX, birds are rendered using Niagara effects and are treated as background (not annotated).}
\label{tab:simd3_total_counts}
\renewcommand{\arraystretch}{1.5}
\begin{tabular}{lrrrr}
\hline
\textbf{Subset} & \textbf{Drone} & \textbf{Bird} & \textbf{Both} & \textbf{Total} \\
\hline
Non-VFX  & 32{,}366 & 48{,}097 & 32{,}436 & 112{,}899 \\
VFX      & 46{,}086 & 0        & 0        & 46{,}086 \\
Weather  & 7{,}114  & 2{,}939  & 9{,}601  & 19{,}654 \\
\hline
\textbf{Total} & \textbf{85{,}566} & \textbf{51{,}036} & \textbf{42{,}037} & \textbf{178{,}639} \\
\hline
\end{tabular}
\end{table}

\begin{table}[t]
\centering
\caption{Scene-wise distribution of images across Non-VFX and VFX subsets. Scenes are sourced from the Unreal Marketplace. ``--'' indicates the scene is not used in that subset.}
\label{tab:simd3_scene_counts}
\renewcommand{\arraystretch}{1.5}
\begin{tabular}{lrr}
\hline
\textbf{Scene} & \textbf{Non-VFX Images} & \textbf{VFX Images} \\
\hline
WWT            & 5{,}353  & 3{,}331 \\
CityPark       & 21{,}361 & 3{,}364 \\
DynamicCity    & 13{,}221 & 4{,}870 \\
CityCreator    & 4{,}918  & 12{,}741 \\
Downtown       & 35{,}556 & 8{,}531 \\
Bridge         & 32{,}490 & -- \\
RuralAustralia & --       & 13{,}249 \\
\hline
\textbf{Total} & 112{,}899 & 46{,}086 \\
\hline
\end{tabular}
\end{table}

\begin{table}[t]
\centering
\caption{Weather subset breakdown by condition. ``Other'' corresponds to the remaining weather configuration not explicitly tagged as fog/snow in the sequence naming (included to preserve exact totals).}
\label{tab:simd3_weather_breakdown}
\renewcommand{\arraystretch}{1.5}
\begin{tabular}{lrrrr}
\hline
\textbf{Condition} & \textbf{Drone} & \textbf{Bird} & \textbf{Both} & \textbf{Total} \\
\hline
Fog   & 3{,}163 & 1{,}926 & 5{,}121 & 10{,}210 \\
Snow  & 3{,}047 &   865 & 2{,}830 & 6{,}742 \\
Other &   904 &   148 & 1{,}650 & 2{,}702 \\
\hline
\textbf{Total} & 7{,}114 & 2{,}939 & 9{,}601 & 19{,}654 \\
\hline
\end{tabular}
\end{table}

\section{Experiments and Results }

After generating the SimD3 dataset, we conducted extensive experiments for Drone detection. Following our prior study \cite{pandat2025}, YOLOv5 was adopted as the baseline architecture which is suitable for Drone detection application. The SimD3 dataset is trained on medium variant of Yolov5 known as Yolov5m. To further improve the performance on simulated dataset a modified variant termed \textbf{Yolov5m+C3b} proposed.

Detection performance is evaluated using standard object detection metrics,
including Precision, Recall, mean Average Precision at an IoU threshold of 0.5 (mAP@0.5),
and mean Average Precision averaged over IoU thresholds from 0.5 to 0.95 (mAP@0.5:0.95).

\begin{figure*}
    \centering
    \includegraphics[width=\linewidth]{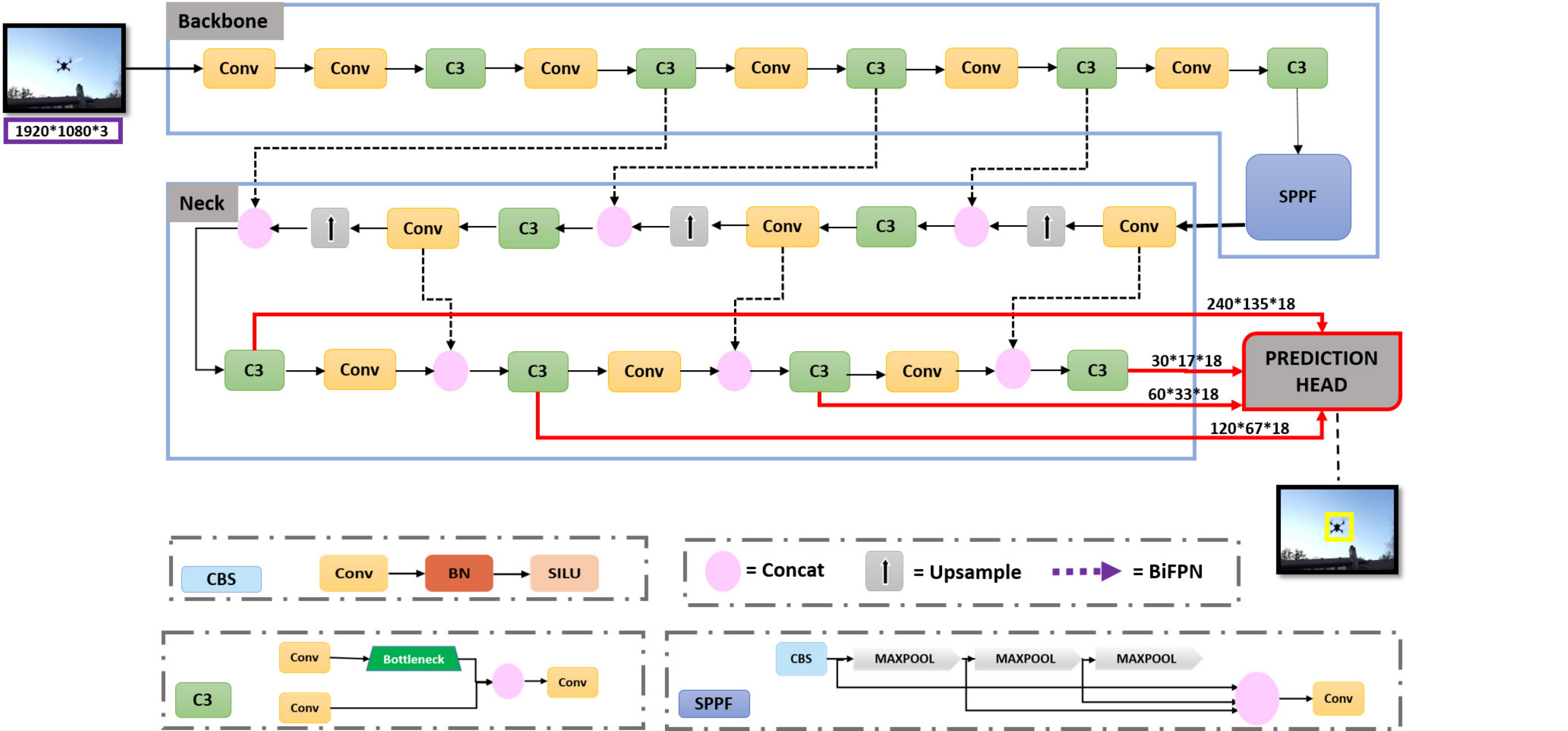}
    \caption{Yolov5 architecture}
    \label{fig:yolov5}
\end{figure*}
\subsection{Yolov5m+C3b Architecture}

\begin{figure}
    \centering
    \includegraphics[width=\linewidth]{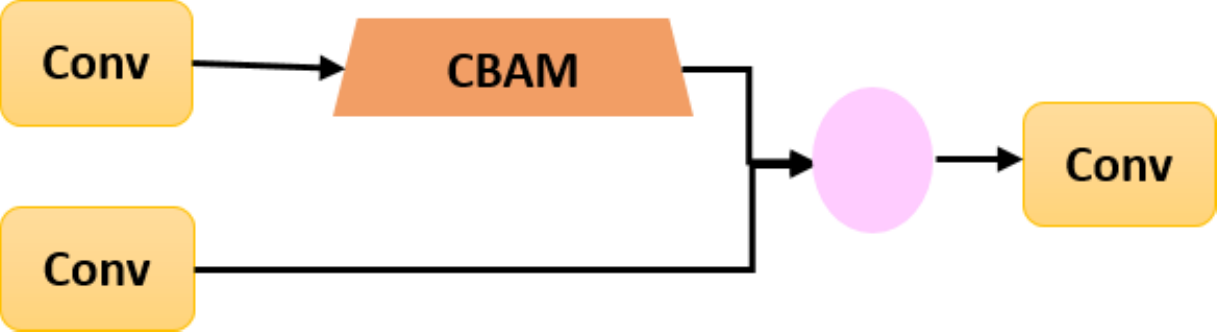}
    \caption{C3b Block}
    \label{fig:c3b_architecture}
\end{figure}

Figure \ref{fig:yolov5} presents the Yolov5 architecture . In the proposed Yolov5m+C3b model, all standard C3 modules of Yolov5 architecture are replaced with C3b blocks \ref{fig:c3b_architecture}. Unlike conventional C3 blocks that rely on bottleneck layers for feature transformation, the C3b module removes these bottlenecks and incorporates attention driven feature refinement using Convolutional Block Attention Module (CBAM) \cite{66} . This design enables the network to:
\begin{itemize}
    \item Emphasize informative spatial and channel-wise features relevant to drone detection,
    \item Improve multi-scale feature aggregation, which is critical for detecting small and distant targets,
    \item Reduce redundant feature representations while preserving contextual information across detection layers.
\end{itemize}

Figure \ref{fig:c3b_architecture} illustrates the internal structure of the C3b block, highlighting how attention-enhanced convolutional pathways replace traditional bottleneck operations.

\subsection{Training Setup}

Both Yolov5m and Yolov5m+C3b models were trained on the SimD3 dataset under identical experimental conditions to ensure a fair comparison. Training was performed for 170 epochs with an early stopping patience of 10 epochs FOR 1920*1080 image resolution. All hyperparameters, including learning rate, batch size, and input resolution, were kept consistent across experiments.

The dataset was split into \textbf{46,622 training images} and \textbf{6,461 validation images}. This split provides sufficient diversity for training while maintaining a representative validation set for unbiased performance evaluation. All images contain annotated drone instances captured under varying backgrounds, scales, and viewpoints, making the dataset well suited for small-object detection studies.

\begin{table*}
\centering
\caption{Performance comparison of Yolov5m and Yolov5m+C3b on the SimD3 dataset}
\label{tab:yolov5m_results}
\renewcommand{\arraystretch}{1.5}
\begin{tabular}{lcccccccc}
\hline
\textbf{Model} & \textbf{Dataset} & \textbf{Precision} & \textbf{Recall} & \textbf{mAP@0.5} & \textbf{mAP@0.5:0.95} & \textbf{Layers} & \textbf{Params (M)} & \textbf{GFLOPs} \\
\hline
Yolov5m & SimD3 & 0.9627 & 0.9306 & 0.9611 & 0.8362 & 378 & 35.25 & 48.9 \\
Yolov5m+C3b & SimD3 & \textbf{0.9820} & \textbf{0.9557} & \textbf{0.9754} & \textbf{0.8774} & 707 & 46.00 & 51.8 \\
\hline
\end{tabular}
\end{table*}

Table~\ref{tab:yolov5m_results} shows that Yolov5m+C3b consistently outperforms the baseline Yolov5m across all evaluation metrics. In particular, improvements of \textbf{1.43\% in mAP@0.5} and \textbf{4.12\% in mAP@0.5:0.95} indicate enhanced localization accuracy and robustness in complex backgrounds.

Although the proposed model introduces a higher number of layers and parameters, the computational overhead remains modest, with an increase of only \textbf{2.9 GFLOPs}. This trade-off is acceptable given the observed gains, especially for small and distant drone targets.

To assess real-world generalization, the SimD3 dataset was augmented with the DUT-AntiUAV benchmark, forming a combined \textbf{SimD3+DUT-AntiUAV} training set. This experiment aims to reduce the synthetic-to-real domain gap and evaluate the transferability of learned representations.

Both models were trained on the combined dataset and evaluated under two settings: (i) testing on the combined dataset and (ii) cross-dataset evaluation on the DUT-AntiUAV test set.

Table~\ref{tab:synthetic_real_results} summarizes the results.

\begin{table*}
\centering
\caption{Performance comparison on synthetic + real (SimD3 + DUT-AntiUAV) datasets}
\label{tab:synthetic_real_results}
\renewcommand{\arraystretch}{1.5}
\begin{tabular}{lcccccc}
\hline
\textbf{Model} & \textbf{Training Dataset} & \textbf{Test Dataset} & \textbf{Precision} & \textbf{Recall} & \textbf{mAP@0.5} & \textbf{mAP@0.5:0.95} \\
\hline
Yolov5m & SimD3+DUT-AntiUAV & DUT-AntiUAV & 0.7361 & 0.5765 & 0.6453 & 0.3863 \\
Yolov5m+C3b & SimD3+DUT-AntiUAV & DUT-AntiUAV & \textbf{0.9009} & \textbf{0.7322} & \textbf{0.8230} & \textbf{0.4989} \\
Yolov5m & SimD3+DUT-AntiUAV & SimD3+DUT-AntiUAV & 0.9041 & 0.8112 & 0.8846 & 0.7107 \\
Yolov5m+C3b & SimD3+DUT-AntiUAV & SimD3+DUT-AntiUAV & \textbf{0.9672} & \textbf{0.9098} & \textbf{0.9475} & \textbf{0.8133} \\
\hline
\end{tabular}
\end{table*}

The results indicate that incorporating real-world data improves robustness for both models. Across all evaluation settings, Yolov5m+C3b consistently achieves higher performance, particularly in cross-dataset evaluation on DUT-AntiUAV, suggesting improved generalization.

\subsection{Cross-Dataset Generalization to Unseen Real-World Benchmarks}

To further assess out-of-distribution performance, both models were evaluated on two unseen real-world datasets, VisioDect and LRDD, without any fine-tuning. Table~\ref{tab:cross_dataset_results} reports the results.

\begin{table*}
\centering
\caption{Cross-dataset evaluation on unseen real-world benchmarks}
\label{tab:cross_dataset_results}
\renewcommand{\arraystretch}{1.5}
\begin{tabular}{lcccccc}
\hline
\textbf{Model} & \textbf{Training Dataset} & \textbf{Test Dataset} & \textbf{Precision} & \textbf{Recall} & \textbf{mAP@0.5} & \textbf{mAP@0.5:0.95} \\
\hline
Yolov5m & SimD3+DUT-AntiUAV & VisioDect & 0.3706 & 0.3759 & 0.1919 & 0.0476 \\
Yolov5m+C3b & SimD3+DUT-AntiUAV & VisioDect & \textbf{0.4242} & 0.3677 & \textbf{0.2183} & \textbf{0.0543} \\
Yolov5m & SimD3+DUT-AntiUAV & LRDD & 0.4062 & 0.2509 & 0.2321 & 0.1073 \\
Yolov5m+C3b & SimD3+DUT-AntiUAV & LRDD & \textbf{0.5538} & \textbf{0.3658} & \textbf{0.3787} & \textbf{0.1586} \\
\hline
\end{tabular}
\end{table*}

Although cross-dataset generalization remains challenging due to significant domain differences, Yolov5m+C3b consistently outperforms the baseline across all unseen benchmarks, indicating improved transferability of learned features.

\subsection{Qualitative Analysis}

Qualitative results further support the quantitative findings. Compared to Yolov5m, the proposed Yolov5m+C3b model produces more stable detections and tighter bounding boxes, particularly for small drones in cluttered backgrounds and low-contrast scenarios. The attention enhanced architecture helps suppress background noise and focus on salient drone features.

Failure cases remain for both models when drones occupy extremely small pixel regions . These observations highlight the inherent challenges of drone detection and motivate future work on domain adaptation and robustness enhancement.

\subsection{Drone Detection under Adverse Weather}

In addition to clear-weather scenarios, we are actively investigating drone detection performance under adverse environmental conditions. To this end, SimD3 includes a dedicated \textit{Weather} subset featuring challenging visual degradations such as fog and snow, which significantly reduce visibility and contrast.

Preliminary experiments indicate that these conditions introduce notable performance degradation compared to clear-weather scenes, primarily due to reduced edge sharpness and increased background ambiguity. This highlights the importance of weather-aware training data and robust feature representations for reliable aerial surveillance.


Overall, the results demonstrate that SimD3 provides a strong foundation for small drone detection, while also enabling systematic study of robustness under challenging weather conditions such as fog and snow.

\section{Conclusion}

This paper introduced \textbf{SimD3}, a large-scale high-fidelity synthetic dataset designed to support robust drone detection in complex aerial environments. By explicitly modeling drones with heterogeneous payloads, incorporating multiple bird species as realistic distractors, and utilizing diverse Unreal Engine~5 environments with controlled weather, lighting, and flight trajectories, SimD3 addresses key limitations of existing synthetic drone datasets. Comprehensive experiments using the YOLOv5 framework demonstrate that SimD3 provides strong supervision for small-object drone detection and improves model robustness when combined with real-world data. Furthermore, the proposed attention-enhanced \textbf{Yolov5m+C3b} architecture consistently outperforms the baseline across synthetic, mixed-domain, and cross-dataset evaluations, indicating improved feature discrimination and generalization. Future work will extend SimD3 to additional drone categories, more extreme environmental conditions, and investigate domain adaptation strategies and multi-modal sensing to further enhance real-world deployment performance.
\bibliographystyle{IEEEtran}
\bibliography{ref}

\end{document}